%% file: final.tex
\documentclass[10pt,journal,compsoc]{IEEEtran}

\ifCLASSOPTIONcompsoc
\usepackage[nocompress]{cite}
\usepackage{times}
\usepackage{epsfig}
\usepackage{graphicx}
\usepackage{amsmath}
\usepackage{amssymb}
\usepackage{xcolor}
\usepackage{multirow}
\usepackage{makecell}
\usepackage{algorithm}
\usepackage{algorithmic}
\usepackage{bbm}
\usepackage{url}
\usepackage{adjustbox}
\usepackage{booktabs}
\usepackage{threeparttable}
\usepackage[colorlinks]{hyperref}
\usepackage{setspace}
\hypersetup{
	colorlinks=red,
	linkcolor=red,
	filecolor=red,
	citecolor=green,      
	urlcolor=cyan,
}
\else
\usepackage{cite}
\usepackage{times}
\usepackage{epsfig}
\usepackage{graphicx}
\usepackage{amsmath}
\usepackage{amssymb}
\usepackage{xcolor}
\usepackage{multirow}
\usepackage{makecell}
\usepackage{algorithm}
\usepackage{algorithmic}
\usepackage{bbm}
\usepackage{url}
\usepackage{adjustbox}
\usepackage{threeparttable}
\usepackage{caption}
\usepackage[colorlinks]{hyperref}
\hypersetup{
	colorlinks=red,
	linkcolor=red,
	filecolor=red,
	citecolor=red,      
	urlcolor=cyan,
}

\fi

\begin{document}
	\title{Visual Object Tracking across Diverse Data Modalities: A Review}
	
	\author{Mengmeng~Wang,~\IEEEmembership{Member,~IEEE},
		Teli~Ma, Shuo~Xin, Xiaojun~Hou, Jiazheng~Xing, Guang Dai, Jingdong~Wang,~\IEEEmembership{Fellow,~IEEE}, and~Yong~Liu,~\IEEEmembership{Member,~IEEE}
		\IEEEcompsocitemizethanks{\IEEEcompsocthanksitem Mengmeng Wang, Teli~Ma, Shuo~Xin, Xiaojun~Hou, Jiazheng~Xing, and Yong Liu are with the Laboratory of Advanced Perception on Robotics and Intelligent Learning, College of Control Science and Engineering, Zhejiang University, Hangzhou 310027, Zhejiang, China; E-mail: mengmengwang@zju.edu.cn, telima9868@gmail.com, 22232036@zju.edu.cn, 22232009@zju.edu.cn, jzxing83@gmail.com and yongliu@iipc.zju.edu.cn. Yong Liu is the corresponding author. Mengmeng~Wang and Teli~Ma are equal contributors.
  \IEEEcompsocthanksitem Guang Dai is with State Grid Shanxi Electric Power Company Limited, E-mail: guang.gdai@gmail.com.
  \IEEEcompsocthanksitem Jingdong Wang is with Baidu, E-mail: wangjingdong@outlook.com}
		\thanks{Manuscript received December 11, 2022; revised March 1, 2024.}}

	\markboth{Journal of \LaTeX\ Class Files,~Vol.~14, No.~8, August~2015}%
	{Shell \MakeLowercase{\textit{et al.}}: Bare Demo of IEEEtran.cls for Computer Society Journals}

	\IEEEtitleabstractindextext{%
		\begin{abstract}
			Visual Object Tracking (VOT) is an attractive and significant research area in computer vision, which aims to recognize and track specific targets in video sequences where the target objects are arbitrary and class-agnostic. The VOT technology could be applied in various scenarios, processing data of diverse modalities such as RGB, thermal infrared and point cloud. Besides, since no one sensor could handle all the dynamic and varying environments, multi-modal VOT is also investigated. This paper presents a comprehensive survey of the recent progress of both single-modal and multi-modal VOT, especially the deep learning methods. Specifically, we first review three types of mainstream single-modal VOT, including RGB, thermal infrared and point cloud tracking. In particular, we conclude four widely-used single-modal frameworks, abstracting their schemas and categorizing the existing inheritors. Then we summarize four kinds of multi-modal VOT, including RGB-Depth, RGB-Thermal, RGB-LiDAR and RGB-Language. Moreover, the comparison results in plenty of VOT benchmarks of the discussed modalities are presented. Finally, we provide recommendations and insightful observations, inspiring the future development of this fast-growing literature.

		\end{abstract}

		}

	\maketitle
	\IEEEdisplaynontitleabstractindextext
	\IEEEpeerreviewmaketitle

    \input{intro}
    \input{related_works}
	\input{single_modal}	
    \input{multi_modal}

    \input{datasets}
    \input{discussion}

	\section{Conclusion}
	In this paper, we comprehensively review the literature on visual object tracking from various data modalities. First, we outline three types of unimodality, including RGB images, thermal infrared images, and 3D LiDAR-based point clouds. In particular, we recognize four paradigms of unimodal VOT, including the popular discriminative correlation filter and Siamese-like matching, the rarely summarized instance classification/detection, and the latest unified Transformer. Next, we review four kinds of multi-modalities, including RGB-Depth, RGB-Thermal, RGB-LiDAR and RGB-Language. We present the widely used benchmarks for all the modalities and list the comparison results of vast methods. Finally, we discuss our recommendations for the future development directions of VOT. We hope that this survey provides the reader with a comprehensive overview and an easy-to-follow guide to getting started with VOT.

	\ifCLASSOPTIONcaptionsoff
	\newpage
	\fi
	
	\bibliographystyle{IEEEtran}
	\bibliography{IEEEabrv,egbib.bib}
       
\end{document}

%% file: intro.tex
\IEEEraisesectionheading{\section{Introduction}\label{sec:introduction}}

		\begin{figure}[tbp]
		\begin{center}
			\includegraphics[width=1\columnwidth]{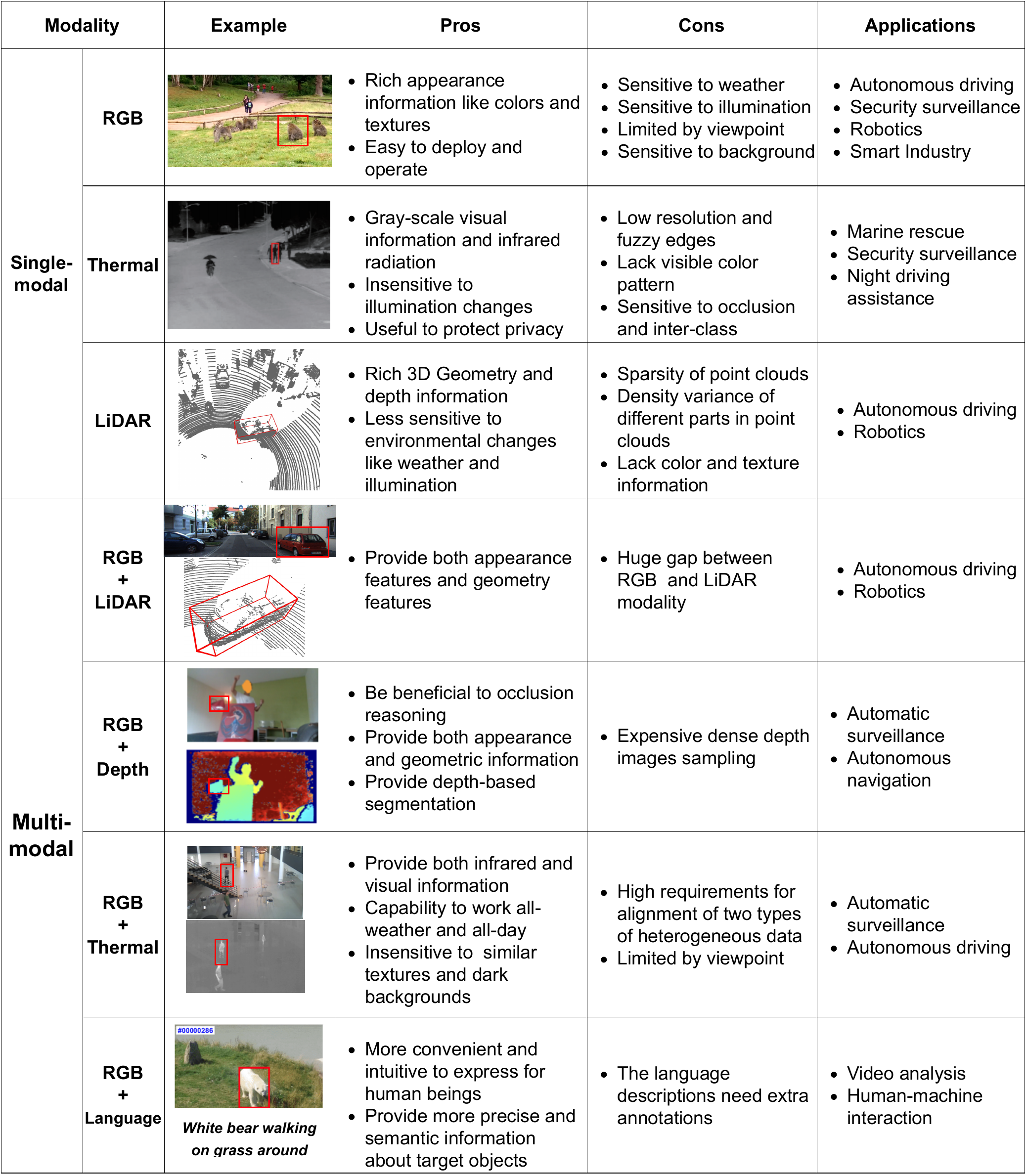}
		\end{center}
  \vspace{-10.5pt}
		\caption{\textcolor{black}{The advantages, disadvantages and applications of the discussed three single-modal VOT and four multi-modal VOT.}}
  		\vspace{-10.5pt}
		\label{figure:proscons}
	\end{figure}

	\IEEEPARstart{V}{isual} object tracking (VOT) has been a highly active topic in the past decades due to its crucial applications in broad scenarios like video surveillance~\cite{simtrack2022,henriques2014high,siamrpn2018}, autonomous vehicles~\cite{3dsiameserpn,v2b}, mobile robots~\cite{wang2017,wang2018}, human-machine interaction~\cite{gti,snlt}, etc. The VOT task is defined as: given a target bounding box location in the first frame, a tracker needs to recognize and locate this target in all the following sequences consistently and robustly, \textcolor{black}{where the targets could be arbitrary instances and class-agnostic.} The task is challenging since 1) the target will undergo lots of complex appearance changes like deformation, rotation, scale variation, motion blur, and out of view; 2) the background will raise various uncontrollable influences like illumination change, similar object distractors, occlusions, and clutter and 3) the video capture sensors may be shaking and moving. 
	
	As an essential task of computer vision, there are various data modalities for VOT. The most common one is RGB videos, which attract lots of researchers to this task due to their popularity and easy accessibility. \textcolor{black}{The RGB-modal VOT provides approximate target locations in the image coordinate with 2D bounding boxes, laying the foundation of many higher-level image analysis tasks like pose estimation, gait/activity recognition, fine-grained classification, and so on. The evolution of RGB-based VOT~\cite{kalal2011tracking,henriques2014high,hare2015struck,dsst2016} is enduring and has a long history, which is further accelerated by the advent of deep learning~\cite{ostrack2022,chen2020siamese,fear2022,siamfc,DiMP} and large-scale datasets~\cite{lasot,got10k,trackingnet}.}
 	We mainly focus on the methods of the past decade, especially the methods based on Deep Neural Network (DNN). 
	According to their pipelines, we classify the mainstream RGB trackers into four categories, namely Discriminative Correlation Filters (DCF)~\cite{DiMP,mayer2022transforming}, Siamese Trackers~\cite{siamrpn++2019,siamcar2020,siamban2022}, Instance Classification/Detection (ICD)~\cite{MDNet,uct2017,han2017branchout} and One-stream Transformers (OST)~\cite{simtrack2022,sbt2022,mixformer2022}. We present the four deep-learning-based paradigms with the most compact components in Fig.~\ref{figure:pipeline} for illustration. The first two paradigms have been very popular in the last decade, while the latter two are rarely or even not mentioned in previous surveys because ICD is not as common as DCF and siamese, and OST is a new paradigm \textcolor{black}{just born since 2022}.

	On the other hand, drawbacks of RGB modality are also obvious and the application scenarios are limited. First, it works unsatisfactorily during the night and bad weather like rainy and snowy days. Under these harsh visual conditions, the Thermal Infrared (TIR) based VOT~\cite{ECO-MM,GFSNet,ECO-LS} recorded by the TIR cameras can be used, capturing the heat radiated from living beings and track the target in the absence of light.
	Second, the lack of depth prior hinders the single RGB-modal VOT from sensing 3D geometric and localization information, which is quite significant for application scenarios like autonomous vehicles and mobile robots.
	Recently, the LiDAR-based VOT~\cite{p2b,v2b,bat,m2track} has emerged to solve this problem via exploring the intrinsic geometric structure of 3D point clouds. 
    The geometric structure of LiDAR points is beneficial to perceiving the depth of the target object, thereby providing accurate 3D shapes and locations. 
	Therefore, we also outline the methods of the two unimodal VOT (TIR-based and LiDAR-based) in this paper. Besides, it is easy to seek common paradigms between these modalities for better understanding. For example, the TIR-based trackers always build the framework following the DCF and siamese, as TIR data format is quite similar to RGB images. Also, the LiDAR-based VOT borrows the idea of siamese schema from RGB modality, and evolves the paradigm to dominate the 3D tracking community.
	
	Moreover, since different unimodal VOT has their pros and cons, trackers fused with multi-modal cues have been proposed with the potential improvement of both accuracy and robustness. More specifically, fusion means combining the information of two or more modalities to track targets. For instance, the TIR sensors are insensitive to illumination changes, camouflage, and pose variations of the object, but it is difficult to distinguish the TIR profile of different people in the crowd. On the other hand, RGB sensors have the opposite characteristics. Hence, it is intuitive to fuse the two modalities to complement each other~\cite{SOWP,SGT,ADRNet}. Also, the fusion choices may vary substantially for different applications. For example, the RGB-LiDAR~\cite{F-siamese,alireza} could be a good candidate for robot following, which requires accurate 3D information. The RGB-Language VOT~\cite{guo2022divert,gti,snlt} could be applied for human-machine interaction. With a large number of practical requirements, some researchers in the VOT community are motivated to shift their work toward integrating multiple modalities for constructing robust tracking systems.
	
	Existing review papers about VOT mainly focus on different aspects and taxonomy of single RGB-modal methods \cite{fiaz2019handcrafted,javed2022visual,ondravsovivc2021siamese,zhang2021recent,chen2022visual,marvasti2021deep,li2018deep,soleimanitaleb2022single,han2022single}. For instance, the most recent surveys \cite{chen2022visual} classify the existing RGB trackers into generative trackers and discriminative trackers. Javed \textit{et al.} \cite{javed2022visual} introduces two well-known paradigms of the RGB-based VOT, namely, DCF and siamese. However, these previous works do not contain the latest popular transformer-based methods, which not only set up new state-of-the-art performance but also bring many insightful research directions. Besides,  the ICD paradigm is not well demonstrated. Moreover, there are very few reviews on multi-modal VOT, either presenting only two modalities (RGB-Depth and RGB-TIR) \cite{zhang2020multi} or emphasizing the fusion of multi-cue features (color, gradient, contour, spatial energy, thermal profile, etc.) \cite{walia2016recent,kumar2020recent}. Last five years, we have witnessed significant progress in multi-modal VOT. At the same time, several new research directions have emerged, such as LiDAR-based VOT, RGB-LiDAR VOT, and RGB-language VOT. However, none of these studies have been well summarized in previous VOT surveys.
	
	In this paper, we systematically review the VOT methods from the perspective of the data modalities, considering the recent development of both the single-modal VOT and the multi-modal VOT. We summarize the reviewed modalities in Fig.~\ref{figure:proscons} with their representative examples, advantages, disadvantages, and applications. Specifically, we first outline the methods of three prevalent unimodal VOT: RGB-based, TIR-based and LiDAR-based. 
	Next, we introduce trackers of four types of multiple fused modalities, including RGB-Depth, RGB-TIR, RGB-LiDAR and RGB-Language. Except for the algorithms, we also report and discuss the benchmark datasets and results. The main contributions of this survey are summarized as follows.
	\begin{enumerate}
		\item We comprehensively review the VOT methods from the perspective of data modalities, including both prevalent single modalities (three types) and multiple modalities (four types). 
		To the best of our knowledge, this is the first review work that demonstrates the newly emerged LiDAR-based, RGB-LiDAR, and RGB-Language VOT methods.
		\item We summarize four widely-used frameworks for single-modal DNN-based trackers, abstracting their schemas and presenting their corresponding customized inheritors.
		\item We provide a comprehensive overview of over 300 papers from the VOT community on more recent and advanced approaches, thus offering the readers with state-of-the-art methods and pipelines.
		\item We present extensive comparisons of existing methods on widely-used benchmarks for various modalities, and finally give insightful discussions and promising future research directions. 
	\end{enumerate}
	
    The rest of this paper is organized as follows: Section~\ref{sec:related} introduces the existing surveys of VOT and elaborates on the different aspects of this paper. Section~\ref{sec:singlemodal} reviews VOT methods using different single data modalities and their comparison results, respectively. Section~\ref{sec:multimodal} summarize multi-modality VOT approaches. Section~\ref{sec:datasets} introduces the VOT datasets of different modalities. Finally, Section~\ref{sec:discussion} discusses the potential future development of VOT. Note that due to the page limitation, we move several result tables, including part of the single-modal and all the multi-modal, into Appendix A, and introduce the VOT datasets of different modalities in Appendix B. 
	

%% file: related_works.tex
\section{Related Works}\label{sec:related}
	VOT is a fundamental and valuable technology for practical applications, and it has been a prevalent research area for many years, with plenty of methods proposed every year. There are many VOT surveys that have been published in the past decades which are devoted to sort out the development of this field from various perspective. We will present these works from the perspective of the modality and illustrate the difference in this paper.

	\subsection{Single-modality VOT Surveys} 
	The prior art of single-modality reviews are mainly about RGB-based trackers~\cite{yang2011recent,li2013survey,smeulders2013visual,wang2019comparison,zhang2013sparse,fiaz2019handcrafted,pflugfelder2017depth,ondravsovivc2021siamese,zhang2021recent,chen2022visual,marvasti2021deep,li2018deep,yilmaz2006object,soleimanitaleb2022single,han2022single}. In 2006, Yilmaz at al.~\cite{yilmaz2006object} proposed the first survey about generic VOT rather than specialized target tracking like the human-specific ones. This survey categorized the tracking methods based on the object and motion representations, like points, primitive geometric shapes, object silhouette, and contour, etc.
 
    Numerous VOT surveys were then proposed. For instance, Yang \textit{et al.}~\cite{yang2011recent} reviewed VOT methods on the basis of the tracking procedure, including feature descriptors, online learning, integration of context information, and Monte Carlo sampling. Li \textit{et al.}~\cite{li2013survey} focused on the appearance modeling of VOT, listing several kinds of representations like raw pixel representation, optical flow representation, histogram representation, and so on. In the past decade, we have witnessed the rapid development of deep-learning-based trackers 
    The most famous DCF and siamese pipelines are also demonstrated in recent surveys.
    Ondra{\v{s}}ovi{\v{c}} \textit{et al.}~\cite{ondravsovivc2021siamese} summarized the main challenges and the core principles of  siamese-based trackers. Zhang \textit{et al.}~\cite{zhang2021recent} reviewed the tracking algorithms from the DCF-based and deep-learning-based viewpoints. Marvasti \textit{et al.}~\cite{marvasti2021deep} provided a systematical survey about deep-learning-based trackers from nine key network aspects like network architecture, network exploitation, network training, and so on. More recently, Soleimanitaleb \textit{et al.}~\cite{soleimanitaleb2022single} classified the tracking algorithms into four branches, namely feature-based, segmentation-based, estimation-based and learning-based ones. Chen \textit{et al.}~\cite{chen2022visual} divided the trackers into generative and discriminative trackers, introducing both deep and non-deep methods. As for the VOT methods based on single TIR modality or LiDAR modality, we have found no general object tracking surveys that merely introduce these two kinds of single-modality VOT in literature up to now.
	
	Our work differs from previous single-modal surveys mainly in three folds: (1) Unlike previous unimodal reviews, our survey is more comprehensive that includes three mainstream single-modal VOT (the RGB modality, the TIR modality, and the LiDAR modality). To the best of our knowledge, we are the first to extensively introduce LiDAR-based trackers, which is a newly emerged direction.
    (2) For the RGB modality, we divide the existing trackers into four more complete paradigms, namely, discriminative correlation filters, Siamese-like matching, instance classification/detection, and unified transformers, where the last two classes are rarely or even not mentioned in previous surveys. (3) We systematically review existing VOT approaches, including the most state-of-the-art trackers of not only the three kinds of single modalities but also eight multiple modalities.
 
	\subsection{Multi-modality Surveys} 
We have observed very few previous surveys\cite{zhang2020multi,walia2016recent,kumar2020recent,yang2022rgbd,zhang2020object} that focus on multi-modal trackers. Walia \textit{et al.}~\cite{walia2016recent} is the first multi-modality survey, which emphasized two single-modal VOT of RGB and TIR and five kinds of multi-cue trackers including RGB-TIR, RGB-Audio, RGB-Laser, RGB-RF and stereo vision. Zhang \textit{et al.}~\cite{zhang2020multi} reviewed two kinds of multi-modality trackers, namely RGB-Depth and RGB-TIR, based on the auxiliary modality and tracking framework. Kumar \textit{et al.}~\cite{kumar2020recent} also presented multi-modal trackers (RGB-TIR, RGB-Depth, RGB-Audio and stereo vision) along with single-modal ones, and they classified papers into traditional approaches and deep-learning-based approaches. 
Zhang \textit{et al.}~\cite{zhang2020object} classified RGB-TIR trackers into pixel-level, feature-level and decision-level fusion ways. Yang \textit{et al.}~\cite{yang2022rgbd} focused on RGB-Depth trackers and introduced their fusion strategies, depth usages, and tracking frameworks.

The differences between previous surveys and this paper are twofold: (1) We review three types of single modalities as well as four multi-sensor combinations, providing a much more comprehensive overview than prior works and covering the most state-of-the-art methods. (2) To the best of our knowledge, we are the first to summarize both the RGB-language and RGB-LiDAR modalities, which are emerging yet highly promising directions.

%% file: single_modal.tex
\section{Single-modal Object Tracking}\label{sec:singlemodal}
	\begin{figure*}[tbp]
   \vspace{-1.5pt}
		\begin{center}
			\includegraphics[width=2\columnwidth]{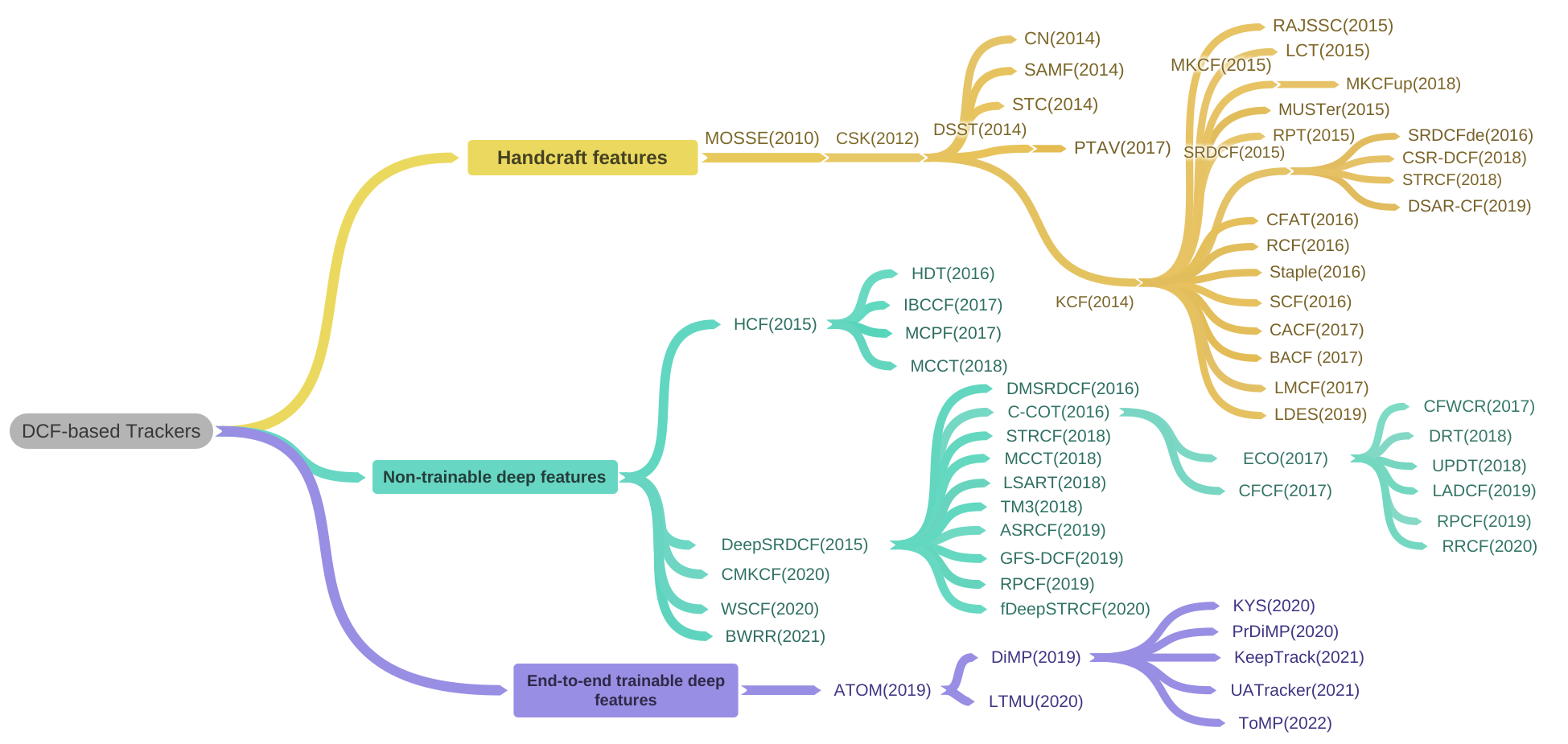}
		\end{center}
   \vspace{-1.5pt}
		\caption{Develop lineage of representative DCF-based RGB trackers. The methods are divided according to their employed features, including handcraft features, non-trainable deep features and end-to-end trainable deep features.}
   \vspace{-1.5pt}
		\label{figure:dcf}
	\end{figure*}
\begin{table*}[!ht]
    \centering
    \caption{The performance of different RGB-based trackers from 2022 to 2024. SR: success rate; PR: precision rate; NPR: normalized precision rate; AO: average overlap; S0.5/0.75: success rate at overlap threshold 0.5/0.75; EAO: expected average overlap; A: accuracy; R: robustness; FPS: mean speed (Take the maximum speed that can be achieved in the corresponding paper).}
 \label{tab:RGB2024}
    \footnotesize
    \scalebox{0.68}{\begin{tabular}{c c|c c c|c c|c c c|c c c|c c|c c|c c|c}
    \hline
        \multirow{2}*{Trackers} & \multirow{2}*{Publication} 
        & \multicolumn{3}{c|}{LaSOT~\cite{lasot} }
        & \multicolumn{2}{c|}{TrackingNet~\cite{trackingnet}} 
        & \multicolumn{3}{c|}{GOT-10k~\cite{got10k}} 
        & \multicolumn{3}{c|}{VOT2018~\cite{vot2018}} 
        & \multicolumn{2}{c|}{UAV123~\cite{uav123}} 
        & \multicolumn{2}{c|}{NFS30~\cite{nfs}} 
        & \multicolumn{2}{c|}{OTB100~\cite{otb100}} 
        & \multirow{2}*{FPS}   
        \\
        \cline{3-19}
        ~ & ~ & SR(\%) & PR(\%) & NPR(\%) &SR(\%) & PR(\%) & AO(\%) & S0.5(\%) & S0.75(\%) & EAO & A & R & SR(\%) & PR(\%) & SR(\%) & PR(\%) & SR(\%) & PR(\%) & ~ \\ \hline
A3DCF\cite{zhu2021robust}                  & TMM2022     & 37.6 & 38.2 & -    & 62.5  & 58.0  & 42.7 & 46.7 & 13.8 & 0.406  & 0.548  & 0.162 & 53.5 & 77.8 & -    & -    & 69.6  & 92.5  & 42.6  \\
BWRR\cite{BWRR2021}                        & TMM2022     & -    & -    & -    & -     & -     & -    & -    & -    & -      & -      & -     & 47.6 & 65.0 & -    & -    & 67.1  & 87.4  & 3.7  \\
SiamCorners\cite{yang2021siamcorners}      & TMM2022     & 48.0 & 55.5 & -    & 69.5  & 64.7  & -    & -    & -    & -      & -      & -     & 61.4 & 81.9 & 53.7 & -    & 67.1  & 88.7  & 42    \\
STGL\cite{jiang2020stgl}                   & TMM2022     & -    & -    & -    & -     & -     & -    & -    & -    & -      & -      & -     & -    & -    & -    & -    & 62.3  & 86.8  & -     \\
Fan \textit{et al.}\cite{fan2021discriminative}     & TMM2022     & 50.5 & 48.7 & -    & 75.1  & 71.0  & 62.1 & 72.4 & 50.6 & \textbf{0.516}  & \textbf{0.621}  & 0.147 & -    & -    & -    & -    & 71.7  & \textbf{94.1}  & -     \\
SiamBAN\cite{siamban2022}                  & TPAMI2022   & 53.1 & 54.1 & -    & 71.6  & 68.5  & -    & -    & -    & 0.473  & 0.598  & 0.155 & 64.4 & 84.6 & -    & -    & 70.2  & 92.3  & -     \\
Chan \textit{et al.}\cite{chan2022siamese}          & TIP2022     & -    & -    & 73.9 & -     & -     & -    & -    & -    & 0.344  & -      & -     & 67.1 & 86.5 & -    & -    & 68.2  & 88.4  & -     \\
Li \textit{et al.}\cite{li2022}                     & TIP2022     & -    & -    & -    & 56.8  & 49.6  & -    & -    & -    & -      & -      & -     & -    & -    & 40.6 & -    & 65.3  & -     & 40    \\
MixFormer\cite{mixformer2022}              & CVPR2022    & 70.1 & 76.3 & 79.9 & 83.9  & 83.1  & 71.2 & 80.0 & 67.8 & -      & -      & -     & 70.4 & 91.8 & -    & -    & -     & -     & 25    \\
GTELT\cite{zhou2022global}                 & CVPR2022    & 67.7 & 73.2 & 75.9 & 82.5  & 81.6  & -    & -    & -    & -      & -      & -     & -    & -    & -    & -    & -     & -     & 26    \\
UTT\cite{utt2022}                          & CVPR2022    & -    & 67.2 & -    & 79.7  & 77.0  & 67.2 & 76.3 & 60.5 & -      & -      & -     & -    & -    & -    & -    & -     & -     & 25    \\
ToMP\cite{mayer2022transforming}           & CVPR2022    & 68.5 & 73.5 & 79.2 & 81.5  & 78.9  & -    & -    & -    & -      & -      & -     & 69.0 & -    & 66.9 & -    & 70.1  & -     & 24.8  \\
RBO\cite{RBO2022}                          & CVPR2022    & 55.8 & 57.0 & -    & -     & -     & 64.4 & 76.7 & 50.9 & -      & -      & -     & 64.5 & -    & 61.3 & -    & 70.1  & -     & -     \\
CSWinTT\cite{CSWinTT2022}                  & CVPR2022    & 66.2 & 70.9 & 75.2 & 81.9  & 79.5  & 69.4 & 78.9 & 65.4 & -      & -      & -     & 70.5 & 90.3 & -    & -    & -     & -     & 12    \\
SBT\cite{sbt2022}                          & CVPR2022    & 66.7 & 71.1 & -    & -     & -     & 70.4 & 80.8 & 64.7 & -      & -      & -     & -    & -    & -    & -    & 71.9  & 92.4  & 62    \\
InBN\cite{InBN2022}                        & IJCAI2022   & 65.7 & 70.7 & 76.0 & 81.7  & -     & -    & -    & -    & -      & -      & -     & 69.0 & 89.0 & 66.8 & 80.2 & 71.1  & -     & 67    \\
SparseTT\cite{sparsett2022}                & IJCAI2022   & -    & 70.1 & 74.8 & 81.7  & 79.5  & 69.3 & 79.1 & 63.8 & -      & -      & -     & 70.4 & -    & -    & -    & 70.4  & -     & 39    \\
HybTransT \cite{HybTransT2022}             & IJCAI2022   & -    & -    & 73.9 & -     & -     & -    & -    & -    & 0.285  & -      & -     & 67.1 & 86.5 & -    & -    & 68.2  & 88.4  & -     \\
SLTtrack\cite{SLTtrack2022}                & ECCV2022    & 66.8 & -    & 75.5 & 82.8  & 81.4  & 67.5 & 76.5 & 60.3 & -      & -      & -     & -    & -    & -    & -    & -     & -     & -     \\
Unicorn\cite{unicorn}                      & ECCV2022    & 68.5 & 74.1 & 76.6 & 83.0  & 82.2  & -    & -    & -    & -      & -      & -     & -    & -    & -    & -    & -     & -     & 23    \\
RTS\cite{rts2022}                          & ECCV2022    & 69.7 & 73.7 & 76.2 & 81.6  & 79.4  & \textbf{85.2} & \textbf{94.5} & \textbf{82.6} & -      & -      & -     & 67.6 & 89.4 & 65.4 & \textbf{82.8} & -     & -     & 30    \\
OSTrack\cite{ostrack2022}                  & ECCV2022    & 71.1 & 77.6 & 81.1 & 83.9  & 83.2  & 73.7 & 83.2 & 70.8 & -      & -      & -     & 70.7 & -    & -    & -    & -     & -     & 105.4 \\
FEAR\cite{fear2022}                        & ECCV2022    & 60.9 & 60.9 & -    & -     & -     & 64.5 & 74.6 & -    & -      & -      & -     & -    & -    & -    & -    & -     & -     & \textbf{205}   \\
AiATrack\cite{aiatrack2022}                & ECCV2022    & 69.0 & 73.8 & 79.4 & 82.7  & 80.4  & 69.6 & 80.0 & 63.2 & -      & -      & -     & 70.6 & -    & 67.9 & -    & 69.6  & -     & 38    \\
SimTrack\cite{simtrack2022}                & ECCV2022    & 70.5 & 76.2 & 79.7 & 83.4  & \textbf{87.4}  & 69.8 & 78.8 & 66.0 & -      & -      & -     & \textbf{71.2} & 91.6 & -    & -    & -     & -     & 40   \\
 \textcolor{black}{F-BDMTrack}~\cite{yang2023foreground} &ICCV2023 &72.0 &77.7 &81.5  & 84.5& 84.0 &75.4 &84.3 &72.9&-&-&- & 70.9 &- &67.3 &- & 69.9&- &-\\
  \textcolor{black}{ROMTrack}~\cite{romtrack}  &ICCV2023 &69.3 &75.6 &78.8  & 83.6 &82.7 &72.9 &82.9 &70.2 &-&-&- &- &-& 68.0 &- &71.4&- &62  \\
 \textcolor{black}{MixFormer V2}~\cite{mixformerv2} & NeurIPS2023 &70.6 &76.2& 80.8 &83.4&81.6 &73.9&-&- &-&-&-  &69.9& \textbf{92.1} &- &-&-&- & 165\\
 \textcolor{black}{RFGM}~\cite{rfgm} &NeurIPS2023 &70.3 &76.4 &82.0  &84.7 &83.6 &74.1 &84.6 &71.8 &-&-&-  &68.5 &-& - &- &71.5&- &71  \\
  \textcolor{black}{DropMAE}~\cite{dropmae} &CVPR2023 &71.8&78.1 &81.8 & 84.1 &- &75.9 &86.8 &72.0 &-&-&- &-&-&-&- &69.6&- &- \\
   \textcolor{black}{GRM}~\cite{grm} &CVPR2023 & 69.9 &75.8 &79.3 &84.0 &83.3 &73.4 &82.9 &70.4 &-&-&- &70.2 &- &65.6& &-&- &45\\
 \textcolor{black}{ARTrack}~\cite{artrack} &CVPR2023 &73.1 &80.3 &82.2 & 85.6 &86.0 &78.5 &87.4 &77.8 &-&-&-  &\textbf{71.2} &-& 67.9 &- &-&- &26   \\
 \textcolor{black}{SeqTrack}~\cite{seqtrack} &CVPR2023 &72.5 &79.3 &81.5  &85.5 &85.8 &74.8 &81.9 &72.2 &-&-&- &68.5 &-& 66.2 &- &-&- &5 \\
 \textcolor{black}{VideoTrack}~\cite{videotrack} &CVPR2023 & 70.2 &76.2 &- &83.8 & 83.1 & 72.9&81.9&69.8&-&-&- &69.7 &89.9 & - &- &-&- &- \\
  \textcolor{black}{HIPTrack}~\cite{hiptrack} &None2023 & 72.7 &79.5 &82.9 &84.5 &83.8 &77.4 &88.0 &74.5 &-&-&-  &70.5 &- &\textbf{68.1 }&- &71.0 &- &-\\
\textcolor{black}{ ODTrack}~\cite{odtrack} &AAAI2024 &\textbf{74.0} &\textbf{82.3} &\textbf{84.2} &\textbf{86.1} &86.7 &78.2 &87.2 &77.3 &-&-&- &- &- & - &- &72.4&- &32  \\
 \textcolor{black}{EVPTrack}~\cite{evptrack} &AAAI2024 &72.7 &80.3 &82.9 &84.4&- &76.6 &86.7 &73.9  &-&-&-  &70.9 &- & - &- &-&- &28 \\

\hline
\end{tabular}}
\end{table*}
\begin{table*}[!ht]
    \centering
    \caption{The performance of different RGB-based trackers from 2019 to 2021. SR: success rate; PR: precision rate; NPR: normalized precision rate; AO: average overlap; S0.5/0.75: success rate at overlap threshold 0.5/0.75; EAO: expected average overlap; A: accuracy; R: robustness; FPS: mean speed (Take the maximum speed that can be achieved in the corresponding paper).}
 \label{tab:RGB2021}
    \footnotesize
    \scalebox{0.65}{\begin{tabular}{c c|c c c|c c|c c c|c c c|c c|c c|c c|c}
    \hline
        \multirow{2}*{Trackers} & \multirow{2}*{Publication} 
        & \multicolumn{3}{c|}{LaSOT} 
        & \multicolumn{2}{c|}{TrackingNet} 
        & \multicolumn{3}{c|}{GOT-10k} 
        & \multicolumn{3}{c|}{VOT2018} 
        & \multicolumn{2}{c|}{UAV123} 
        & \multicolumn{2}{c|}{NFS30} 
        & \multicolumn{2}{c|}{OTB100} 
        & \multirow{2}*{FPS}   
        \\
        \cline{3-19}
        ~ & ~ & SR(\%) & PR(\%) & NPR(\%) &SR(\%) & PR(\%) & AO(\%) & S0.5(\%) & S0.75(\%) & EAO & A & R & SR(\%) & PR(\%) & SR(\%) & PR(\%) & SR(\%) & PR(\%) & ~ \\ \hline
DSAR-CF\cite{DSARCF2019}                   & TIP2019     & -    & -    & -    & -     & -     & -    & -    & -    & -      & -      & -     & -    & -    & -    & -    & 63.9  & 83.2  & 6     \\
LADCF\cite{LADCF2019}                      & TIP2019     & -    & -    & -    & -     & -     & -    & -    & -    & 0.389  & 0.512  & 0.159 & 51.8 & -    & -    & -    & 69.6  & 90.6  & 10.8  \\
Dong \textit{et al.}\cite{dong2019quadruplet}       & TIP2019     & -    & -    & -    & -     & -     & -    & -    & -    & -      & -      & -     & -    & -    & -    & -    & 73.9  & 78.2  & -     \\
Li \textit{et al.}\cite{li2018visual}               & TPAMI2019   & -    & -    & -    & -     & -     & -    & -    & -    & -      & -      & -     & -    & -    & -    & -    & 63.2  & 89.4  & -     \\
LDES\cite{LDES2019}                        & AAAI2019    & -    & -    & -    & -     & -     & -    & -    & -    & -      & -      & -     & -    & -    & -    & -    & 63.4  & 76.0  & 20    \\
SPM-Tracker\cite{wang2019spm}              & CVPR2019    & -    & -    & -    & -     & -     & -    & -    & -    & -      & -      & -     & -    & -    & -    & -    & 68.7  & 89.9  & 120   \\
SiamRPN++\cite{siamrpn++2019}              & CVPR2019    & 49.6 & 56.9 & -    & 73.3  & 69.4  & -    & -    & -    & -      & -      & -     & 61.3 & 80.7 & -    & -    & 69.6  & 91.4  & 35    \\
RPCF\cite{sun2019roi}                      & CVPR2019    & -    & -    & -    & -     & -     & -    & -    & -    & -      & -      & -     & -    & -    & -    & -    & 69.6  & 93.2  & 5     \\
SiamMask\cite{siammask2019}                & CVPR2019    & -    & -    & -    & -     & -     & -    & -    & -    & 0.387  & 0.642  & 0.276 & -    & -    & -    & -    & -     & -     & 55    \\
TADT\cite{TADT}                            & CVPR2019    & -    & -    & -    & -     & -     & -    & -    & -    & -      & -      & -     & -    & -    & -    & -    & 66.0  & 86.6  & 33.7  \\
C-RPN\cite{C-RPN2019}                      & CVPR2019    & 45.9 & -    & -    & 66.9  & 61.9  & -    & -    & -    & -      & -      & -     & -    & -    & -    & -    & 66.3  & -     & 36    \\
ASRCF\cite{ASRCF2019}                      & CVPR2019    & 35.9 & 33.7 & -    & -     & -     & -    & -    & -    & -      & -      & -     & -    & -    & -    & -    & 69.2  & 92.2  & 28    \\
SiamDW\cite{siamdw2019}                    & CVPR2019    & -    & -    & -    & -     & -     & -    & -    & -    & -      & -      & -     & -    & -    & -    & -    & 67.0  & 90.0  & -     \\
GCT\cite{gao2019graph}                     & CVPR2019    & -    & -    & -    & -     & -     & -    & -    & -    & -      & -      & -     & 50.8 & 73.2 & -    & -    & 64.8  & 85.4  & 50    \\
ATOM\cite{danelljan2019atom}               & CVPR2019    & 51.5 & -    & 57.6 & 70.3  & 64.8  & -    & -    & -    & 0.401  & 0.590  & 0.204 & 65.0 & -    & 59.0 & -    & -     & -     & 30    \\
Huang \textit{et al.}\cite{huang2019bridging}       & ICCV2019    & -    & -    & -    & -     & -     & -    & -    & -    & -      & -      & -     & 58.6 & -    & 51.5 & -    & 64.7  & -     & 10    \\
MLT\cite{mlt2019}                          & ICCV2019    & 36.8 & -    & -    & -     & -     & -    & -    & -    & -      & -      & -     & -    & -    & -    & -    & 61.1  & -     & 48    \\
GradNet\cite{li2019gradnet}                & ICCV2019    & 36.5 & 35.1 & -    & -     & -     & -    & -    & -    & -      & -      & -     & -    & -    & -    & -    & 63.9  & 86.1  & 80    \\
DiMP\cite{DiMP}                            & ICCV2019    & 56.9 & -    & -    & 74.0  & 68.7  & 61.1 & 71.7 & 49.2 & 0.440  & 0.597  & 0.153 & 65.4 & -    & 62.0 & -    & 68.4  & -     & 43    \\
GFS-DCF\cite{xu2019joint}                  & ICCV2019    & -    & -    & -    & 60.90 & 56.57 & -    & -    & -    & 0.397  & 0.511  & 0.143 & -    & -    & -    & -    & 69.3  & 93.2  & 8     \\
UpdateNet\cite{updatenet2019}              & ICCV2019    & 47.5 & 56.0 & -    & 67.7  & 62.5  & -    & -    & -    & 0.481  & 0.610  & 0.206 & -    & -    & -    & -    & -     & -     & -     \\
SPLT\cite{splt2019}                        & ICCV2019    & -    & -    & -    & -     & -     & -    & -    & -    & -      & -      & -     & -    & -    & -    & -    & -     & -     & 25.7  \\
Ma \textit{et al.}\cite{ma2020situp}                & TIP2020     & -    & -    & -    & -     & -     & -    & -    & -    & -      & -      & -     & -    & -    & -    & -    & 57.6  & 78.2  & -     \\
CMKCF\cite{CMKCF2020}                      & TMM2020     & -    & -    & -    & -     & -     & -    & -    & -    & 0.2462 & 0.5015 & -     & -    & -    & -    & -    & 61.5  & 82.2  & 18.13 \\
fDeepSTRCF\cite{fDeepSTRCF2020}            & TIP2020     & 31.1 & 29.1 & -    & -     & -     & -    & -    & -    & -      & -      & -     & 51.1 & 73.7 & -    & -    & 68.6  & -     & 20    \\
MSCF\cite{zhang2020mining}                 & TIP2020     & -    & -    & -    & -     & -     & -    & -    & -    & -      & -      & -     & -    & -    & -    & -    & 67.7  & -     & 20    \\
WSCF\cite{WSCF2020}                        & TIP2020     & 26.5 & 24.5 & -    & -     & -     & -    & -    & -    & 0.1841 & 0.5404 & 2.17  & -    & -    & -    & -    & 62.4  & 82.6  & 30    \\
Han \textit{et al.}\cite{han2019ensemble}           & TMM2020     & -    & -    & -    & -     & -     & -    & -    & -    & 0.291  & 0.552  & 0.294 & 57.3 & 80.9 & -    & -    & 70.6  & 93.6  & 0.99  \\
Lu \textit{et al.}\cite{lu2020deep}                 & TPAMI2020   & 34.0 & 34.2 & -    & -     & -     & -    & -    & -    & 0.274  & 0.500  & 0.279 & 53.5 & 76.8 & -    & -    & 66.2  & 91.8  & -     \\
DROL\cite{drol2020}                        & AAAI2020    & 53.7 & 62.4 & -    & 74.6  & 70.8  & -    & -    & -    & 0.481  & 0.616  & -     & 65.2 & 85.5 & -    & -    & 71.5  & 93.7  & 60    \\
SiamFC++\cite{siamfc++2020}                & AAAI2020    & 54.4 & -    & -    & 75.4  & 70.5  & 59.5 & 69.5 & 47.9 & 0.426  & 0.587  & 0.183 & -    & -    & -    & -    & 68.3  & -     & 160   \\
GlobalTrack\cite{globaltrack2020}          & AAAI2020    & 52.1 & 52.7 & 59.9 & 70.4  & 65.6  & -    & -    & -    & -      & -      & -     & -    & -    & -    & -    & -     & -     & 6     \\
DTNet\cite{zhang2020online}                & NeurIPS2020 & 51.6 & 57.9 & -    & 73.7  & 69.8  & -    & -    & -    & 0.418  & 0.604  & 0.197 & 64.9 & 83.1 & -    & -    & 66.0  & 89.1  & 36    \\
TLPG-Tracker\cite{li2021tlpg}              & IJCAI2020   & 58.1 & -    & -    & 75.8  & 70.6  & 62.9 & 73.5 & -    & 0.459  & 0.606  & 0.149 & -    & -    & -    & -    & -     & 69.8  & 36    \\
KYS\cite{bhat2020know}                     & ECCV2020    & 55.4 & -    & 63.3 & 74.0  & -     & -    & -    & -    & 0.462  & 0.609  & 0.143 & -    & -    & 63.4 & -    & 69.5  & -     & 20    \\
PG-Net\cite{liao2020pg}                    & ECCV2020    & 53.1 & 60.5 & -    & -     & -     & -    & -    & -    & 0.447  & 0.618  & 0.192 & -    & -    & -    & -    & 69.1  & 89.2  & 42    \\
Liu \textit{et al.}\cite{liu2020object}             & ECCV2020    & -    & -    & -    & -     & -     & -    & -    & -    & 0.436  & 0.616  & 0.201 & 64.9 & -    & -    & -    & 69.3  & -     & 13    \\
Ocean\cite{ocean2020}                      & ECCV2020    & 56.0 & 56.6 & -    & -     & -     & 61.1 & 72.1 & -    & 0.489  & 0.598  & 0.117 & -    & -    & -    & -    & 68.4  & 92.0  & 58    \\
MAML\cite{wang2020tracking}                & CVPR2020    & 52.3 & -    & -    & 75.7  & -     & -    & -    & -    & 0.452  & 0.635  & 0.159 & -    & -    & -    & -    & 71.2  & 92.6  & 40    \\
SiamCAR\cite{siamcar2020}                  & CVPR2020    & 51.4 & -    & 59.8 & -     & -     & -    & -    & -    & 0.452  & 0.597  & 0.178 & 63.1 & 83.3 & 59.4 & -    & 69.6  & 91.0  & 52.27 \\
Siam R-CNN\cite{SiamRCNN2020}              & CVPR2020    & 64.8 & -    & 72.2 & 81.2  & 80.0  & 64.9 & -    & -    & 0.408  & 0.624  & 0.220 & 64.9 & 83.4 & 63.9 & -    & 70.1  & 89.1  & 15.2  \\
SiamBAN\cite{chen2020siamese}              & CVPR2020    & 51.4 & -    & 59.8 & -     & -     & -    & -    & -    & 0.452  & 0.597  & 0.178 & 63.1 & 83.3 & 59.4 & -    & 69.6  & 91.0  & -     \\
ROAM\cite{roam2020}                        & CVPR2020    & 44.7 & 44.5 & -    & 67.0  & 62.3  & 46.5 & 53.2 & 23.6 & -      & -      & -     & -    & -    & -    & -    & 68.1  & 90.8  & 20    \\
RLS-RTMDNet\cite{gao2020recursive}         & CVPR2020    & -    & -    & -    & -     & -     & -    & -    & -    & -      & -      & -     & 51.6 & -    & -    & -    & -     & -     & -     \\
PrDiMP\cite{danelljan2020probabilistic}    & CVPR2020    & 59.8 & -    & -    & 75.8  & 70.4  & 63.4 & 73.8 & 54.3 & 0.442  & 0.618  & 0.165 & 68.0 & -    & 63.5 & -    & 69.6  & -     & 30    \\
LTMU\cite{ltmu2020}                        & CVPR2020    & 57.2 & 57.2 & -    & -     & -     & -    & -    & -    & -      & -      & -     & -    & -    & -    & -    & -     & -     & 13    \\
SiamAttn\cite{SiamAttn2020}                & CVPR2020    & 56.0 & -    & 64.8 & 75.2  & -     & -    & -    & -    & 0.47   & 0.63   & 0.16  & 65   & 84.5 & -    & -    & 71.2  & 92.6  & 45    \\
CGACD\cite{CGACD2020}                      & CVPR2020    & 51.8 & -    & 62.6 & 71.1  & 69.3  & -    & -    & -    & 0.449  & 0.615  & 0.173 & 63.3 & 83.3 & -    & -    & 71.3  & 92.2  & 70    \\
KeepTrack\cite{mayer2021learning}          & ICCV2021    & 67.1 & 70.2 & 77.2 & -     & -     & -    & -    & -    & -      & -      & -     & 69.7 & -    & 66.4 & -    & 71.2  & -     & 29.6  \\
DTT\cite{yu2021high}                       & ICCV2021    & 60.1 & -    & -    & 79.6  & 78.9  & 68.9 & 79.8 & 62.2 & -      & -      & -     & -    & -    & 65.9 & -    & -     & -     & 54.5  \\
AutoMatch\cite{automatch2021}              & ICCV2021    & -    & 59.9 & -    & 76.0  & 72.6  & 65.2 & 76.6 & 54.3 & -      & -      & -     & -    & -    & -    & -    & 71.4  & 92.6  & 50    \\
SAOT\cite{saot2021}                        & ICCV2021    & 61.6 & -    & 70.8 & -     & -     & 64.0 & 74.9 & -    & 0.501  & -      & -     & -    & -    & 65.6 & 77.8 & 71.4  & 92.6  & 29    \\
Stark\cite{stark2021}                      & ICCV2021    & -    & -    & 77.0 & 82.0  & -     & 68.8 & 78.1 & 64.1 & -      & -      & -     & -    & -    & -    & -    & -     & -     & 30    \\
RRCF\cite{RRCF2020}                        & TMM2021     & 37.2 & -    & 41.0 & -     & -     & -    & -    & -    & 0.335  & 0.538  & 0.890 & -    & -    & -    & -    & 69.9  & 92.0  & 9.8   \\
Tian \textit{et al.}\cite{tian2020siamese}          & TMM2021     & 39.9 & 39.3 & -    & -     & -     & 39.0 & 43.2 & 13.9 & -      & -      & -     & 53.5 & 72.6 & -    & -    & 68.2 & 88.3 & -     \\
CAT\cite{zhang2020cat}                     & TMM2021     & 40.9 & 50.4 & -    & -     & -     & -    & -    & -    & 0.374  & 0.577  & 0.252 & 56.9 & 78.5 & -    & -    & 69.0  & 93.4  & 13.2  \\
STGL\cite{jiang2020stgl}                   & TMM2021     & -    & -    & -    & -     & -     & -    & -    & -    & -      & -      & -     & -    & -    & -    & -    & 62.3  & 86.8  & -     \\
Pu \textit{et al.}\cite{pu2020learning}             & TIP2021     & -    & -    & -    & -     & -     & -    & -    & -    & -      & -      & -     & 52.5 & -    & -    & -    & 65.0  & 88.7  & -     \\
DTC\cite{yang2019visual}                   & TPAMI2021   & -    & -    & -    & -     & -     & -    & -    & -    & -      & -      & -     & -    & -    & -    & -    & 63.8  & 84.8  & 50    \\
Nocal-Siam\cite{tan2021nocal}              & TIP2021     & 53.3 & -    & -    & -     & -     & 60.1 & 68.8 & -    & 0.474  & 0.592  & 0.159 & -    & -    & -    & -    & 67.8  & 90.7  & -     \\
SiamCAN\cite{zhou2021siamcan}              & TIP2021     & 53.8 & 53.3 & -    & -     & -     & -    & -    & -    & 0.462  & 0.605  & 0.183 & 64.8 & 85.7 & -    & -    & 70.5  & 91.9  & -     \\
Dong \textit{et al.}\cite{dong2019dynamical}        & TPAMI2021   & -    & -    & -    & -     & -     & -    & -    & -    & 0.201  & -      & -     & -    & -    & -    & -    & 63.4  & 83.3  & 25    \\
Zhang \textit{et al.}\cite{zhang2021toward}         & TIP2021     & -    & -    & -    & -     & -     & -    & -    & -    & -      & -      & -     & -    & -    & -    & -    & -     & -     & -     \\
TransformerTrack\cite{wang2021transformer} & CVPR2021    & 63.9 & 61.4 & -    & 78.4  & 73.1  & 67.1 & 77.7 & 58.3 & 0.462  & 0.600  & 0.141 & 67.5 & -    & 66.5 & -    & 71.1  & -     & 35    \\
TransT\cite{transT2021}                    & CVPR2021    & 64.9 & 69.0 & 73.8 & 81.4  & 80.3  & 72.3 & 82.4 & 68.2 & -      & -      & -     & 69.1 & -    & 65.7 & -    & 69.4  & -     & 50    \\
DMTrack\cite{dmtrack2021}                  & CVPR2021    & 57.4 & 58.0 & -    & -     & -     & -    & -    & -    & -      & -      & -     & -    & -    & -    & -    & -     & -     & 31    \\
STMTrack\cite{RESiamNets2021}              & CVPR2021    & 60.6 & 63.3 & 69.3 & 80.3  & 76.7  & 64.2 & 73.7 & 57.9 & 0.447  & 0.590  & 0.159 & 64.7 & -    & -    & -    & 71.9  & -     & 37    \\
SiamBAN-ACM\cite{ACM2021}                  & CVPR2021    & 57.2 & -    & -    & 75.3  & 71.2  & -    & -    & -    & -      & -      & -     & 64.8 & -    & -    & -    & 72.0  & -     & 172   \\
SiamRN\cite{siamrn2021}                    & CVPR2021    & 52.7 & 53.1 & -    & -     & -     & -    & -    & -    & 0.470  & 0.595  & \textbf{0.131} & 64.3 & 86.1 & -    & -    & 70.1  & 93.1  & -     \\
SiamGAT\cite{guo2021graph}                 & CVPR2021    & 53.9 & 53.0 & 63.3 & -     & -     & 62.7 & 74.3 & 48.8 & -      & -      & -     & 64.6 & 84.3 & -    & -    & 71.0  & -     & 165   \\
CapsuleRRT\cite{ma2021capsulerrt}          & CVPR2021    & 61.7 & -    & -    & 77.4  & 72.6  & 65.6 & 74.8 & 52.4 & -      & -      & -     & 68.7 & -    & 64.5 & -    & 71.3  & -     & 27    \\
AlphaRefine\cite{yan2021alpha}             & CVPR2021    & 65.3 & 68.0 & 73.2 & 80.5  & 78.3  & 70.1 & 80.0 & 64.2 & -      & -      & -     & -    & -    & -    & -    & -     & -     & 50    \\
CRACT\cite{fan2021cract}                   & IROS2021    & 54.9 & 62.8 & -    & 75.4  & 72.4  & 62.0 & 72.8 & 49.6 & 0.455  & 0.611  & 0.175 & 66.4 & 86.0 & 62.5 & -    & \textbf{72.6}  & 93.6  & 28    \\
UATracker\cite{zhou2021model}              & AAAI2021    & 59.4 & -    & 68.0 & -     & -     & -    & -    & -    & 0.458  & 0.614  & 0.159 & 67.6 & 87.9 & -    & -    & 70.5  & -     & 45    \\
HDRL\cite{zhang2021visual}                 & AAAI2021    & 55.3 & 54.6 & 62.8 & -     & -     & 58.2 & 68.5 & 44.3 & -      & -      & -     & 62.0 & 82.7 & -    & -    & 67.0  & 87.6  & 40    \\
\hline
\end{tabular}}
\end{table*}
\begin{table*}[!ht]
    \centering
    \caption{The performance of different RGB-based trackers from 2015 to 2018. SR: success rate; PR: precision rate; EAO: expected average overlap; A: accuracy; R: robustness; FPS: mean speed (Take the maximum speed that can be achieved in the corresponding paper).}
\label{tab:RGBbefore2018}
    \footnotesize
    \scalebox{0.8}{\begin{tabular}{c c|c c|c c|c c c|c c c|c c c|c}
    \hline
        \multirow{2}*{Trackers} & \multirow{2}*{Publication} 
        & \multicolumn{2}{c|}{OTB50} 
        & \multicolumn{2}{c|}{OTB100} 
        & \multicolumn{3}{c|}{VOT2015} 
        & \multicolumn{3}{c|}{VOT2016} 
        & \multicolumn{3}{c|}{VOT2017} 
        & \multirow{2}*{FPS}   
        \\
        \cline{3-15}~ & ~ & SR(\%) & PR(\%) & SR(\%) & PR(\%) & EAO & A & R & EAO & A & R & EAO & A & R & ~ \\ \hline
MKCF\cite{mkcf2015}                        & ICCV2015    & 59.1  & 78.1  & -     & -     & -      & -      & -     & -      & -     & -     & -      & -     & -     & 15     \\
RAJSSC\cite{RAJSSC2015}                    & ICCV2015    & -     & -     & -     & -     & -      & 0.52   & 1.63  & -      & -     & -     & -      & -     & -     & -      \\
KCF\cite{henriques2014high}                & TPAMI2015   & -     & 73.2  & -     & -     & -      & -      & -     & -      & -     & -     & -      & -     & -     & 172    \\
SRDCF\cite{danelljan2015learning}          & ICCV2015    & -     & -     & 60.5  & 72.9  & -      & -      & -     & -      & -     & -     & -      & -     & -     & 3.8    \\
CF2(HCF)\cite{ma2015hierarchical}          & ICCV2015    & 74.0  & 89.1  & 65.5  & 83.7  & -      & -      & -     & -      & -     & -     & -      & -     & -     & 11     \\
LCT\cite{ma2015long}                       & CVPR2015    & 61.2  & 78.0    & -     & -     & -      & -      & -     & -      & -     & -     & -      & -     & -     & 27.4   \\
Muster\cite{hong2015multi}                 & CVPR2015    & -     & -     & -     & -     & -      & -      & -     & -      & -     & -     & -      & -     & -     & 0.3    \\
DeepSRDCF\cite{danelljan2015convolutional} & ICCV2015    & 64.9  & -     & -     & -     & -      & -      & 1.05  & -      & -     & -     & -      & -     & -     & -      \\
CNN-SVM\cite{hong2015online}               & ICML2015    & 59.7  & 85.2  & -     & -     & -      & -      & -     & -      & -     & -     & -      & -     & -     & -      \\
FCNT\cite{wang2015visual}                  & ICCV2015    & 59.9  & 85.6  & -     & -     & -      & -      & -     & -      & -     & -     & -      & -     & -     & 3      \\
YCNN\cite{chen2017ycnn}                    & TCSVT2016   & 50.6  & 70.82 & 56.17 & 76.65 & -      & -      & -     & -      & -     & -     & -      & -     & -     & 45     \\
Learnet\cite{bertinetto2016learning}       & NeurIPS2016 & -     & -     & -     & -     & -      & 0.5    & -     & -      & -     & -     & -      & -     & -     & 60     \\
SINT\cite{tao2016siamese}                  & CVPR2016    & 65.5  & 88.2  & -     & -     & -      & -      & -     & -      & -     & -     & -      & -     & -     & -      \\
STCT\cite{wang2016stct}                    & CVPR2016    & 64.0  & 78.0  & -     & -     & -      & -      & -     & -      & -     & -     & -      & -     & -     & 2.5    \\
MDNet\cite{MDNet}                          & CVPR2016    & 70.8  & 94.8  & 67.8  & 90.9  & -      & -      & -     & -      & -     & -     & -      & -     & -     & 1      \\
SiamFC\cite{siamfc}                        & ECCV2016    & -     & -     & -     & -     & 0.274  & 0.524  & -     & -      & -     & -     & -      & -     & -     & 86     \\
GOTURN\cite{held2016learning}              & ECCV2016    & -     & -     & -     & -     & -      & -      & -     & -      & -     & -     & -      & -     & -     & 100    \\
CNT\cite{zhang2016robust}                  & TIP2016     & 54.5  & 76.6  & -     & -     & -      & -      & -     & -      & -     & -     & -      & -     & -     & -      \\
Staple\cite{bertinetto2016staple}          & CVPR2016    & -     & -     & -     & -     & -      & 0.538  & -     & -      & -     & -     & -      & -     & -     & 90     \\
HDT\cite{qi2016hedged}                     & CVPR2016    & 60.3  & 88.9  & 59.3  & 84.8  & -      & -      & -     & -      & -     & -     & -      & -     & -     & -      \\
DMSRDCF\cite{DMSRDCF2016}                  & ICPR2016    & -     & -     & 67.4  & -     & -      & 0.58   & 0.92  & -      & -     & -     & -      & -     & -     & 0.0659 \\
SRDCFdecon\cite{SRDCFde2016}               & CVPR2016    & -     & -     & 63.4  & -     & 0.299  & -      & -     & -      & -     & -     & -      & -     & -     & -      \\
DSST\cite{dsst2016}                        & TPAMI2016   & 67.7  & 75.7  & -     & -     & -      & -      & -     & -      & -     & -     & -      & -     & -     & 54.3   \\
C-COT\cite{C-COT}                          & ECCV2016    & -     & -     & 68.2  & 82.4  & -      & 0.54   & 0.82  & -      & -     & -     & -      & -     & -     & -      \\
SANet\cite{fan2017sanet}                   & CVPR2017    & -     & -     & 69.2  & 92.8  & 0.3895 & 0.61   & 0.69  & -      & -     & -     & -      & -     & -     & 1      \\
BranchOut\cite{han2017branchout}           & CVPR2017    & -     & -     & 67.8  & 91.7  & 0.3384 & 0.59   & 0.71  & -      & -     & -     & -      & -     & -     & -      \\
ADNet\cite{yun2017action}                  & CVPR2017    & 65.9  & 90.3  & 64.6  & 88.0  & -      & -      & -     & -      & -     & -     & -      & -     & -     & 15     \\
CACF\cite{mueller2017context}              & CVPR2017    & -     & -     & 59.8  & 81.0  & -      & -      & -     & -      & -     & -     & -      & -     & -     & -      \\
DCCO\cite{johnander2017dcco}               & CAIP2017    & -     & -     & 69.0  & 83.9  & -      & -      & -     & 0.368  & 0.54  & 0.70  & -      & -     & -     & -      \\
BACF\cite{kiani2017learning}               & ICCV2017    & 67.78  & -     & 62.98  & -     & -      & 0.59   & 1.56  & -      & -     & -     & -      & -     & -     & 35.3   \\
CSR-DCF\cite{DCF}                          & CVPR2017    & -     & -     & 58.7  & 73.3  & -      & -      & -     & 0.338  & 0.51  & 0.85  & -      & -     & -     & 13     \\
MCPF\cite{MCPF2017}                        & CVPR2017    & -     & -     & 62.8  & 87.3  & -      & -      & -     & -      & -     & -     & -      & -     & -     & 1.96   \\
Mueller \textit{et al.}\cite{mueller2017context}    & CVPR2017    & -     & -     & 59.8  & 81.0  & -      & -      & -     & -      & -     & -     & -      & -     & -     & -      \\
CFNet\cite{CFNet2017}                      & CVPR2017    & 57.4  & 70.8  & 61.6  & 73.7  & -      & -      & -     & -      & -     & -     & -      & -     & -     & 83     \\
LMCF\cite{wang2017large}                   & CVPR2017    & -     & -     & 58.9  & -     & -      & -      & -     & -      & -     & -     & -      & -     & -     & 80     \\
Obli-RaF\cite{zhang2017robust}             & CVPR2017    & -     & -     & 57.9  & 85.1  & -      & -      & -     & -      & -     & -     & -      & -     & -     & -      \\
ECO\cite{danelljan2017eco}                 & CVPR2017    & -     & -     & 70    & -     & -      & -      & -     & 0.374  & 0.54  & 0.72  & -      & -     & -     & 8      \\
TSN\cite{teng2017robust}                   & ICCV2017    & 58.0  & 80.9  & 64.4  & 86.8  & -      & -      & -     & 0.336  & 0.582 & 0.266 & -      & -     & -     & 1      \\
CFCF\cite{CFCF2018}                        & TIP2017     & -     & -     & 67.8  & 89.9  & -      & -      & -     & 0.3903 & 0.54  & 0.63  & -      & -     & -     & 1.34   \\
RFL\cite{yang2017recurrent}                & ICCV2017    & -     & -     & 58.1  & 82.5  & -      & -      & -     & 0.2222 & -  & -  & -      & -     & -     & 15     \\
DNT\cite{chi2017dual}                      & TIP2017     & 66.4  & 90.7  & 62.7  & 85.1  & -      & -      & -     & -      & -     & -     & -      & -     & -     & 25.4   \\
RSST\cite{zhang2018robust}                 & TPAMI2017   & 59.0  & 78.9  & 58.3  & 78.9  & -      & -      & -     & -      & -     & -     & -      & -     & -     & -      \\
AOGTracker\cite{lu2014online}              & TPAMI2017   & 56.19 & -     & 61.02 & -     & -      & -      & -     & -      & -     & -     & -      & -     & -     & -      \\
IBCCF\cite{li2017integrating}              & ICCV2017    & -     & -     & 63    & 78.4  & -      & -      & -     & 0.266  & 0.51  & 1.22  & -      & -     & -     & 2.19   \\
CFWCR\cite{he2017correlation}              & ICCV2017    & -     & -     & -     & -     & -      & -      & -     & 0.3905 & 0.58  & 0.81  & 0.303  & -     & -     & 4      \\
UCT\cite{uct2017}                          & ICCV2017    & -     & -     & 61.1  & 84.9  & -      & -      & -     & -      & -     & -     & -      & -     & -     & 154    \\
DSiam\cite{Dsiam2017}                      & ICCV2017    & -     & -     & -     & -     & 0.2927 & 0.5566 & -     & -      & -     & -     & -      & -     & -     & 45     \\
EAST\cite{huang2017learning}               & ICCV2017    & -     & 63.8  & 62.9 & -     & 0.34   & 0.57   & 1.03  & -      & -     & -     & -      & -     & -     & 158.9  \\
CREST\cite{CREST2017}                      & ICCV2017    & -     & -     & 62.3  & 83.7  & -      & -      & -     & 0.283  & -     & -     & -      & -     & -     & -      \\
Galoogahi \textit{et al.}\cite{kiani2017learning}   & ICCV2017    & 67.78 & -     & 62.98 & -     & -      & 0.59   & 1.56  & -      & -     & -     & -      & -     & -     & -      \\
FlowTrack\cite{FlowTrack}                  & CVPR2018    & -     & -     & 65.5  & 88.1  & 0.3405 & 0.57   & 0.95  & 0.334  & 0.578 & 0.241 & -      & -     & -     & 12     \\
Siamese-RPN\cite{siamrpn2018}              & CVPR2018    & -     & -     & 63.7  & 85.1  & 0.358  & 0.58   & 0.93  & 0.3441 & 0.56  & 1.08  & -      & -     & -     & 160    \\
VITAL\cite{song2018vital}                  & CVPR2018    & -     & -     & 68.2  & 91.7  & -      & -      & -     & 0.323  & -  & -  & -      & -     & -     & 1.5    \\
P2T\cite{gao2018p2t}                       & TIP2018     & 66.3  & 90.8  & 62.8  & 85.4  & 0.29   & 0.52   & 0.86  & -      & -     & -     & -      & -     & -     & 2      \\
Du \textit{et al.}\cite{du2017iterative}            & TIP2018     & -     & -     & 60.2  & 80.4  & -      & -      & -     & -      & -     & -     & -      & -     & -     & -      \\
Li \textit{et al.}\cite{li2019learning}             & TIP2018     & 57.7  & -     & 62.0  & -     & 0.318  & -      & -     & 0.295  & -     & -     & 0.263  & -     & -     & 82     \\
Li \textit{et al.}\cite{li2018visual-tracking}      & TIP2018     & 58.0  & 79.4  & 55.2  & -     & -      & -      & -     & -      & -     & -     & -      & -     & -     & -      \\
Liu \textit{et al.}\cite{liu2019deformable}         & TIP2018     & -     & -     & 68.3  & 91    & -      & -      & -     & 0.353  & 0.585 & 0.774 & 0.258  & 0.558 & 0.645 & -      \\
RASNet\cite{RASNet}                        & CVPR2018    & -     & -     & 64.2  & -     & 0.327  & -      & -     & -      & -     & -     & 0.281  & -     & -     & 80     \\
TM3\cite{2018tm3}                          & TIP2018     & -     & -     & 58    & 79    & -      & -      & -     & -      & -     & -     & -      & -     & -     & 4      \\
CSR-DCF\cite{CSRDCF2018}                   & TIP2018     & -     & -     & 61.5  & 81.5  & -      & -      & -     & 0.257  & -  & -  & -      & -     & -     & 35.3   \\
DRT\cite{sun2018correlation}               & CVPR2018    & -     & -     & 69.9  & 92.3  & -      & -      & -     & 0.442  & 0.569 & 0.140 & -      & -     & -     & -      \\
HP\cite{dong2018hyperparameter}            & CVPR2018    & 55.4  & 74.5  & 60.1  & 79.6  & -      & 0.578  & 1.578 & -      & -     & -     & -      & -     & -     & 69     \\
SA-Siam\cite{SA-Siam}                      & CVPR2018    & 61.0  & 82.3  & 65.6  & 86.5  & 0.31   & 0.59   & 1.26  & 0.2911 & 0.54  & 1.08  & 0.236  & 0.5   & 0.459 & 50     \\
LSART\cite{LSART2018}                      & CVPR2018    & -     & -     & 67.2  & 92.3  & -      & -      & -     & -      & -     & -     & 0.323  & 0.493 & 0.218 & 1      \\
UPDT\cite{bhat2018unveiling}               & ECCV2018    & -     & -     & 62.0  & -     & -      & -      & -     & -      & -     & -     & -      & -     & -     & -      \\
STRCF\cite{li2018learning}                 & CVPR2018    & -     & -     & 65.1  & 79.6  & -      & -      & -     & 0.279  & 0.53  & 1.32  & -      & -     & -     & 5.3    \\
MCCT\cite{wang2018multi}                   & CVPR2018    & -     & -     & 69.5  & 91.4  & -      & -      & -     & 0.393  & 0.58  & 0.73  & -      & -     & -     & 45     \\
SINT++\cite{wang2018sint++}                & CVPR2018    & 62.4  & 83.9  & 57.4  & 76.8  & -      & -      & -     & -      & -     & -     & -      & -     & -     & -      \\
Pu \textit{et al.}\cite{pu2018deep}                 & NeurIPS2018 & -     & -     & 66.8  & 89.5  & -      & -      & -     & 0.32   & -  & -  & -      & -     & -     & 1      \\
MemTrack\cite{yang2018learning}            & ECCV2018    & -     & -     & 62.6  & 82.0  & -      & -      & -     & 0.2729 & 0.53  & 1.44  & -      & -     & -     & 50     \\
DaSiamRPN\cite{DaSiamRPN2018}              & ECCV2018    & -     & -     & -  & -     & 0.446  & 0.63   & 0.66  & 0.411  & 0.61  & 0.22  & 0.326  & 0.56  & 0.34  & 110    \\
DRL-IS\cite{ren2018deep}                   & ECCV2018    & -     & -     & 67.1  & 90.9  & -      & -      & -     & -      & -   & -   & -      & -     & -     & 10.2   \\
DSLT\cite{lu2018deep}                      & ECCV2018    & -     & -     & 66.0  & 90.9  & -      & -      & -     & 0.3321 & -  & -  & -      & -     & -     & 5.7    \\
UPDT\cite{bhat2018unveiling}               & ECCV2018    & -     & -     & 62.0  & -     & -      & -      & -     & -      & -     & -     & 0.378  & 0.532 & 0.182 & -      \\
ACT\cite{chen2018real}                     & ECCV2018    & 65.7  & 90.5  & 62.5  & 85.9  & -      & -      & -     & 0.2746 & -  & -  & -      & -     & -     & 30     \\
RT-MDNet\cite{jung2018real}                & ECCV2018    & -     & -     & 65.0  & 88.5  & -      & -      & -     & -      & -     & -     & -      & -     & -     & 52     \\
StructSiam\cite{StructSiam}                & ECCV2018    & -     & -     & 62.1  & 85.1  & -      & -      & -     & 0.264  & -     & -     & -      & -     & -     & 45     \\
SACFNet\cite{zhang2018visual}              & ECCV2018    & -     & -     & 69.3  & 91.7  & 0.343  & -      & -     & 0.380  & -     & -     & -      & -     & -     & 23     \\
Dong \textit{et al.}\cite{SiamFC-tri}               & ECCV2018    & 53.5  & 71.3  & 59.2  & 78.1  & -      & -      & -     & -      & -     & -     & 0.2125 & -     & -     & 86    \\
\hline
\end{tabular}}
\end{table*}

	As shown in Fig.~\ref{figure:proscons}, different modalities of VOT have different characteristics and applications. In this section, we introduce three mainstream single-modal VOT methods, including RGB, TIR and LiDAR.
	\subsection{RGB-based Trackers}
	RGB modality refers to RGB images/videos, which are captured by commonly used vision cameras, i.e., a sensor most similar to human eye perception. 
	The core merits of RGB modality can be attributed to the abundant appearance information contained in the recordings, as well as the relatively easy access to the devices.
	Hence, trackers based on this modality are widely used in various scenarios, including autonomous driving, security surveillance, robotics, etc. 
	On the other hand, the RGB camera is a 2D sensor, and it is sensitive to illumination variations, weather changes, and background noise, which pose huge challenges for RGB-modal VOT.
    
	In the pre-deep learning era, RGB-modal trackers rely on handcraft features like Haar \cite{zhang2012real,wang2016robust}, HOG \cite{henriques2014high} and color name \cite{danelljan2014adaptive,cn2014}. And traditional classifiers like SVM \cite{hare2015struck,wang2017large}, multiple instance learning \cite{babenko2010robust} and correlation filters \cite{danelljan2014accurate,zhang2017robust,danelljan2015learning} are employed. Subsequently, with the great progress of deep learning techniques and the born of large-scale benchmark datasets, including LaSOT \cite{lasot}, GOT-10K \cite{got10k} and TrackingNet \cite{trackingnet}, various DNN-based tracking architectures have also been proposed. From the pioneering DLT \cite{wang2013learning} to the recent transformer-based methods~\cite{simtrack2022,CSWinTT2022,transT2021}, DNN-based trackers have been the research mainstream in the past decade. Consequently, we mainly review the advanced deep learning works for RGB-based VOT, categorizing them into four types and introducing them in the following. 
 \textcolor{black}{Besides, according to the development timeline, we list the RGB-based trackers' performance in several widely used benchmarks in Table~\ref{tab:RGB2024}, Table~\ref{tab:RGB2021}, Table~\ref{tab:RGBbefore2018} of this main manuscript, including the trackers from 2022 to 2024, 2019 to 2021 and 2015 to 2018. }Note that the reason for splitting the two tables is that the popular datasets for the two periods are different.
 
\subsubsection{Discriminative Correlation Filters}\label{sec:dcf}
	In DCF-based tracking, a correlation filter or a model predictor is trained online with the tracked target regions. The target is then detected in consecutive frames by convolving the trained filter via the fast Fourier transform (FFT) or convolution operations of DNN. 
	We list the milestones and nodes of DCF-based tracking developments in Fig.~\ref{figure:dcf}.
	
	Before the deep learning era, the correlation filter dominated for a long time due to its effectiveness and efficiency. It approximates dense sampling by circularly shifting operations and allows the FFT to be employed during the learning process. This technique is first proposed in MOSSE \cite{bolme2010visual} and then further explored and improved in following works like CSK \cite{henriques2012exploiting}, STC \cite{zhang2014fast}, DSST~\cite{dsst2016}, CN~\cite{cn2014}, SAMF~\cite{samf2014}, KCF \cite{henriques2014high}, SRDCF \cite{danelljan2015learning}, BACF \cite{kiani2017learning} and so on. For instance, CSK proposed a kernel correlation filter to realize dense sampling. KCF is a significant milestone of the DCF-based trackers, which proposed a fast multi-channel extension of linear correlation filters to utilize HOG features for tracking. Then, several DCF trackers employed the HOG features, like RAJSSC~\cite{RAJSSC2015}, LCT \cite{ma2015long}, MKCF~\cite{mkcf2015}, MUSTer \cite{hong2015multi}, RPT \cite{li2015reliable}, CFLB \cite{kiani2015correlation}, SRDCF \cite{danelljan2015learning}, CFAT \cite{bibi2016target}, RCF \cite{sui2016real}, Staple \cite{bertinetto2016staple}, SCF \cite{liu2016structural}, SRDCFde~\cite{SRDCFde2016}, CACF \cite{mueller2017context}, BACF \cite{kiani2017learning}, LMCF \cite{wang2017large}, PTAV \cite{fan2017parallel}, STRCF \cite{li2018learning}, LSART \cite{sun2018learning}, CSR-DCF~\cite{CSRDCF2018}, LDES~\cite{LDES2019}, DSAR-CF~\cite{DSARCF2019} and CMKCF~\cite{CMKCF2020}. \textcolor{black}{An important direction among these methods involved extending beyond closed-form computations to enable learning from a search region wider than the template, exemplified by SRDCF, CCOT, BACF, and CSRDCF. This advancement was achieved through formulations utilizing conjugate gradient descent (CGD)~\cite{danelljan2015learning,danelljan2016beyond} and alternating direction method of multipliers (ADMM)~\cite{kiani2017learning,CSRDCF2018}.}

	Stepping into the early stage of the deep learning era, the convolution neural network (CNN) based features gradually replaced the handcraft features, stimulating many non-end-to-end trainable trackers, 
	including DeepSRDCF~\cite{danelljan2015convolutional}, DMSRDCF~\cite{DMSRDCF2016},  HCF~\cite{ma2015hierarchical}, C-COT~\cite{danelljan2016beyond}, HDT \cite{qi2016hedged}, ECO~\cite{danelljan2017eco}, DeepLMCF~\cite{wang2017large}, CFWCR~\cite{he2017correlation}, Obli-RaF~\cite{zhang2017robust}, MCPF~\cite{MCPF2017}, CFCF~\cite{CFCF2018}, IBCCF~\cite{li2017integrating}, MCCT~\cite{wang2018multi}, LSART~\cite{LSART2018}, TM3~\cite{2018tm3},  STRCF~\cite{li2018learning}, UPDT \cite{bhat2018unveiling}, DRT \cite{sun2018correlation},  LADCF~\cite{LADCF2019}, GFS-DCF \cite{xu2019joint}, ASRCF~\cite{ASRCF2019}, RPCF~\cite{sun2019roi}, fDeepSTRCF~\cite{fDeepSTRCF2020}, RRCF~\cite{RRCF2020}, CMKCF~\cite{CMKCF2020}, WSCF~\cite{WSCF2020} and BWRR~\cite{BWRR2021}. 
	DeepSRDCF \cite{danelljan2015convolutional} investigated the usage of convolutional layer activations in DCF-based tracking frameworks. HCF \cite{ma2015hierarchical} illustrated the hierarchies of convolutional layers and exploited these multiple levels of abstraction for visual tracking. C-COT \cite{danelljan2016beyond} employed an implicit interpolation model to enable efficient integration of multi-resolution deep feature maps. ECO \cite{ECO} improved C-COT by introducing a factorized convolution operator to dramatically reduce the number of parameters and tackle the over-fitting issues. DeepLMCF \cite{wang2017large} directly used the CNN features to demonstrate its proposed large margin structured SVM classifier. MCCT \cite{wang2018multi} explored a way to combine different appearance features with multiple experts. RPCF \cite{sun2019roi} introduced ROI-based pooling in the correlation filter by enforcing additional constraints on the learned filter weights. These methods mainly regard the DNN as a feature extractor, exploring the better usage of strong representation without training it together with the following classifiers. 
	
\begin{figure}[tbp]
		\begin{center}
			\includegraphics[width=\columnwidth]{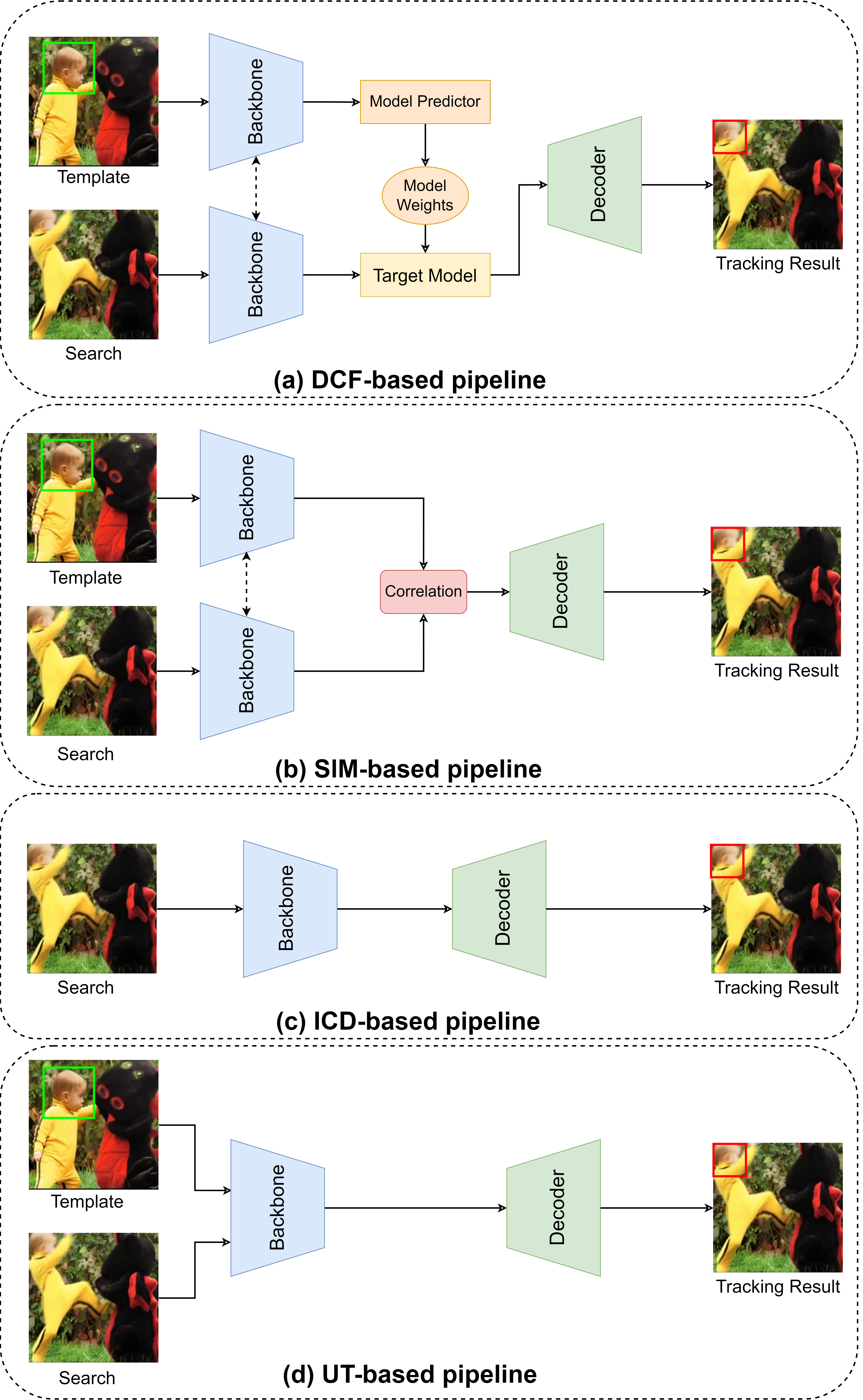}
		\end{center}
   \vspace{-10.5pt}
		\caption{Visualization of the four schemas of RGB-based VOT, which consists of Discriminative Correlation Filters (DCF), Siamese, Instance Classification/Detection (ICD) and One-Stream Transformers (OST).}
   \vspace{-10.5pt}
		\label{figure:pipeline}
	\end{figure}

	Next, we review the end-to-end trainable DCF-based deep trackers. Unlike traditional correlation filters, here, the trackers employ several convolutional layers of DNN to replace the FFT operation and generate the correlation weights. We abstract a general DNN-based DCF schema as shown in Fig~\ref{figure:pipeline}(a), where a model predictor is used to generate the model weights like correlation filters and then operated on the search features to obtain the target model for decoding the results. \textcolor{black}{ATOM \cite{danelljan2019atom} is a pioneering work in this line, which first used two shared ResNet-18 to extract features for template and search regions and then designed a target state estimation predictor and a target classifier. More importantly, it proposes reusing automatic differentiation to implement an efficient gradient descent method for training a DCF based on CGD and Gauss-Newton.} The correlation weights are obtained by fully connected layers and applied channel-wise multiplication. DiMP \cite{DiMP} exploited both target and background appearance information for target model prediction, improving the discrimination of the DCF trackers. It proposed a model predictor to learn the target model, which is updated online. PrDiMP \cite{danelljan2020probabilistic} improved DiMP by a probabilistic regression formulation which modeled the uncertainty in the annotations themselves to counter noise in the annotations and ambiguities in the regression task. KYS \cite{bhat2020know} learned to effectively utilize the scene information by directly maximizing tracking performance on video segments upon DiMP. To realize long-term tracking, LTMU \cite{dai2020high} designed a meta-updater to decide the proper time of model update with a cascaded LSTM module. To tackle the problem of distractors, KeepTrack \cite{mayer2021learning} kept track of distractor objects to continue tracking the target. UATracker \cite{zhou2021model} proposed a sampling method based on uncertainty adjustment to select representative sample frames to feed the discriminative branch of DiMP. Recently, with the success of Transformers, researchers also integrate it into DCF trackers. ToMP \cite{mayer2022transforming} employed a Transformer-based model prediction module to predict the target model weights, which is an encoder-decoder structure. 
	
In summary, DCF-based methods have a long exploration history from the early handcraft to the current DNN-based end-to-end learnable approaches.  
The correlation filter demonstrates its vitality in tackling tracking, deriving hundreds of methods.
Moreover, most of this pipeline needs to update the learned model/filter online to keep track, regardless of using handcraft features or DNNs. This model update process is beneficial for adapting to the target appearance changes and improving the tracking accuracy, but it will limit the DNN-based methods' tracking speed.
	
	\subsubsection{Siamese Trackers}
 	 \begin{figure*}[tbp] \vspace{-8.5pt}
		\begin{center}
			\includegraphics[width=2\columnwidth]{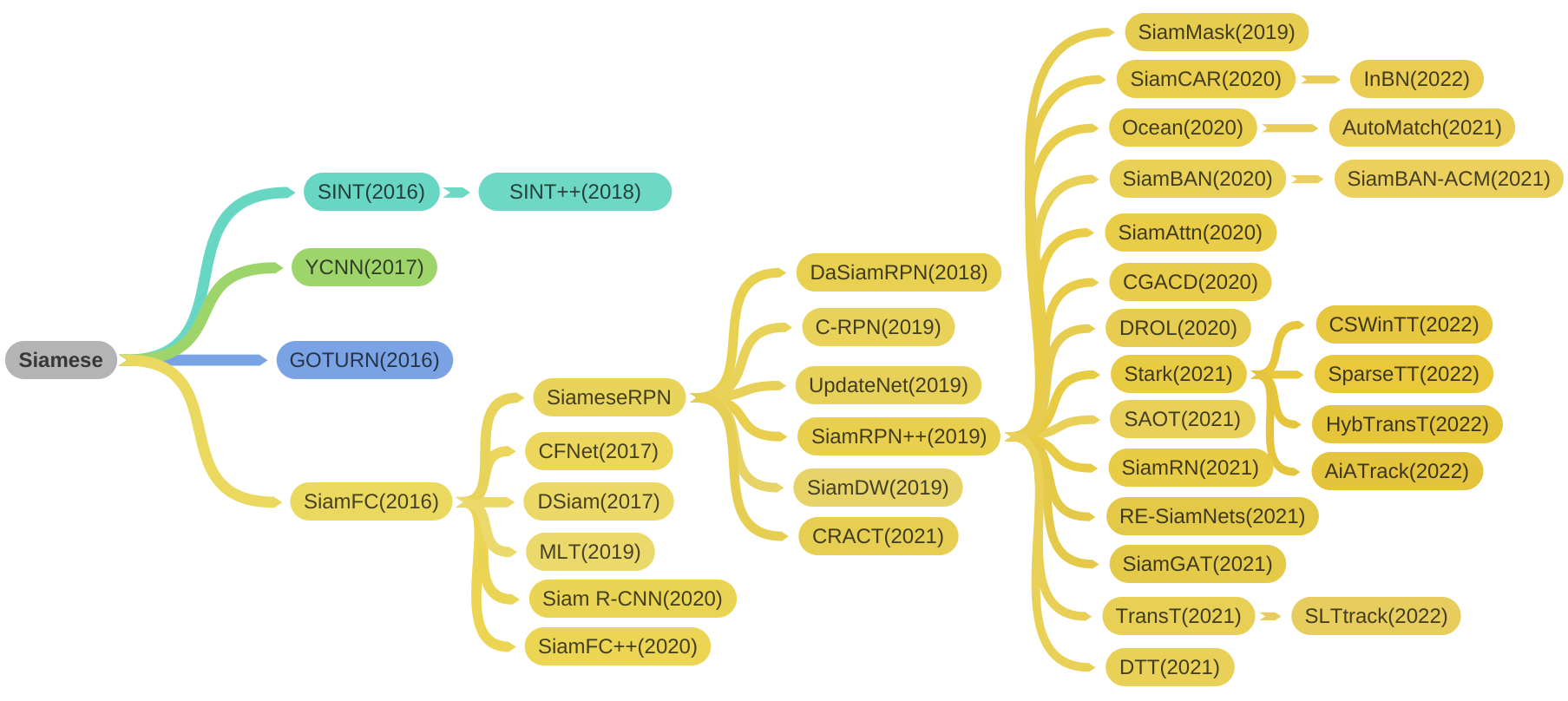}
		\end{center}
   \vspace{-10.5pt}
		\caption{Develop lineage of representative Siamese-based RGB trackers.}
   \vspace{-5.5pt}
		\label{figure:Siamese}
	\end{figure*}
	 Siamese paradigm has been the most prevalent pipeline in recent years. It regards the tracking task as a more general similarity learning problem in an initial offline phase, and then this function is evaluated online during tracking without model updating. \textcolor{black}{More specifically, siamese models commonly employ a shared Siamese network as a feature extractor, tasked with encoding inputs from both the template and search regions. The extracted features are then fused using correlation similarity or specific fusion modules}, as shown in Fig.~\ref{figure:pipeline}(b). The target is localized by choosing the most similar candidates or regressing the bounding boxes with various decoders. 
	 
	 Deep Siamese Network (DSN) was first employed for signature verification task \cite{bromley1993signature} and then applied to many tasks such as face verification \cite{schroff2015facenet,parkhi2015deep}, self-supervised learning \cite{chen2021exploring}, video object segmentation \cite{oh2019video,wang2022delving} and so on. The researchers of VOT also exploited DSN in this area. There were four earliest methods, GOTURN \cite{held2016learning}, YCNN \cite{chen2017ycnn}, SINT \cite{tao2016siamese} and SiamFC \cite{siamfc} born almost simultaneously. Specifically, GOTURN proposed to use DSN to extract features for template and search regions. Then, the template and search representations were simply concatenated together to directly regressed the target location in the search region by fully connected layers.
	 The whole model is trained offline and tracked online without model updating. Similarly, YCNN introduced a DSN as the features extractor and concatenated the obtained template and search features to a response map with fully connected layers. SINT learned a matching function between the features of template and search candidate proposals which were extracted by a DSN. SiamFC proposed a fully convolutional DSN with a cross-correlation layer, which regarded the template features as the convolutional kernel to convolve the search features and computed the similarity map. Based on SiamFC, CFNet \cite{CFNet2017} designed an asymmetric Siamese network, which applied the correlation filter as a differentiable layer in a deep neural network, learning deep features that are tightly coupled to the Correlation Filter. DSiam \cite{Dsiam2017} proposed a dynamic Siamese network to explore the model update problem of SiamFC with a fast general transformation learning model and elementwise multi-layer fusion.
	 
One of the most significant milestones of siamese trackers is SiameseRPN \cite{siamrpn2018}, which absorbed the region proposal subnetwork from the object detection \cite{ren2015faster} area to generate proposals for VOT. SiameseRPN was end-to-end trained offline with large-scale image pairs and online tracked as a local one-shot detection problem. It built a bridge between object detection and VOT and obtained a strong performance with a high speed at that time. After that, plenty of Siamese methods were proposed, like DaSiamRPN \cite{DaSiamRPN2018}, SiamRPN++ \cite{siamrpn++2019}, SiamMask \cite{siammask2019}, C-RPN \cite{C-RPN2019}, SiamDW \cite{siamdw2019}, UpdateNet \cite{updatenet2019}, SiamCAR \cite{siamcar2020}, Siam R-CNN \cite{SiamRCNN2020}, SiamFC++~\cite{siamfc++2020},  SiamAttn \cite{SiamAttn2020}, CRACT~\cite{cract2021}, SiamBAN~\cite{siamban2022}, CGACD~\cite{CGACD2020}, Ocean~\cite{ocean2020}, RE-SiamNets~\cite{RESiamNets2021}, SAOT~\cite{saot2021}, SiamBAN-ACM~\cite{ACM2021}, TransT~\cite{transT2021}, Stark~\cite{stark2021}, SiamRN~\cite{siamrn2021}, DTT\cite{DTT2021}, SiamGAT~\cite{guo2021graph}, InBN~\cite{InBN2022}, AutoMatch~\cite{automatch2021}, SLTtrack~\cite{SLTtrack2022}, CSWinTT~\cite{CSWinTT2022}, SparseTT~\cite{sparsett2022}, HybTransT~\cite{HybTransT2022}, AiATrack~\cite{aiatrack2022}, and so on. We present typical Siamese trackers according to their development vein in Fig. \ref{figure:Siamese}.
	
	One aspect is to exploit deeper and wider networks, like SiamRPN++~\cite{siamrpn++2019} and SiamDW \cite{siamdw2019}. SiamRPN++ proposed a spatial aware sampling strategy to take advantage of features from deeper networks (ResNet-50 or deeper) and the depthwise correlation operation to further improve the accuracy and reduce the model size. SiamDW also investigated how to leverage deeper and wider convolutional neural networks to enhance tracking robustness and accuracy. Although RPN-based trackers \cite{siamrpn2018,siamrpn++2019,siamdw2019} have achieved great success, the main drawback is that they are sensitive to hyper-parameters associated with anchors. Then, some works devoted to eliminating the negative effects of anchors with anchor-free trackers such as Ocean~\cite{ocean2020}, SiamFC++~\cite{siamfc++2020}, SiamCAR~\cite{siamcar2020}, and SiamBAN~\cite{siamban2022}. For example, SiamCAR~\cite{siamcar2020} designed an anchor-free and proposal-free tracker, which decomposed tracking into two subproblems: one classification problem and one regression task, and learned them simultaneously in an end-to-end manner. Another prevalent research direction is to explore the correlation operation of Siamese frameworks, such as the depthwise correlation of SiamRPN++, and others like SiamGAT~\cite{guo2021graph} and SiamBAN-ACM~\cite{ACM2021}. SiamGAT introduced the graph attention mechanism to solve the problem of the fixed-scale cropped template region of cross correlation. SiamBAN-ACM proposed a learnable asymmetric convolution module, which learned to capture the semantic correlation information better. Since most of the Siamese trackers are trained offline and tracked online without model updates for efficiency, many works are devoted to exploring this point. MLT~\cite{mlt2019}  incorporated SiamFC with a meta-learner network to provide the matching network with new appearance information of the target objects. DROL~\cite{drol2020} equipped offline siamese networks with an online module with an attention mechanism to extract target-specific features under L2 error. UpdateNet~\cite{updatenet2019} proposed to replace the handcrafted linearly combined update function of Siamese trackers with a CNN that learns to update.
	
	Very recently, with the development of vision Transformer, Siamese trackers also employ it in this task to explore stronger feature representations and template-search relationship modeling. TransT~\cite{transT2021} presented a Transformer tracking method that combined the template and the search region features solely using the attention mechanism of Transformer. Stark~\cite{stark2021} employed an encoder-decoder Transformer to model the global feature dependencies of both spatial and temporal information, avoiding many postprocessing steps such as cosine window. Then, many recent methods are proposed to improve the Transformer structure to adapt to VOT tasks. For example, CSWinTT~\cite{CSWinTT2022} designed a new transformer architecture with multi-scale cyclic shifting window attention, replacing the original pixel-to-pixel attention strategy with window-level attention. SparseTT~\cite{sparsett2022} used a sparse attention mechanism to find the most relevant information in the search regions, improving self attentions. AiATrack~\cite{aiatrack2022} proposed an attention in attention module to enhance the correlation operations.
	
	In a word, the siamese schema greatly improves the efficiency of the DNN-based method while maintaining the performance by the offline-trained process and no-finetune online tracking. This schema transfers the tracking problem into a template-search similarity matching problem, thus settling the possible performance degradation due to no model updates.
	
	\subsubsection{Instance Classification/Detection}
	One of the most intuitive ways of employing deep learning technology is to use a neural network to memorize the target information and find the target via binary classification. That is to say, this way models the VOT as an instance classification or detection problem that only looks for a particular instance, which may belong to any known or unknown object class. 
	We simplify the ICD pipeline in Fig.~\ref{figure:pipeline}(c), where only the search region will be processed in one forward propagation. The backbone extracts the search features and the decoder outputs the final classification/detection results. Although the forward pipeline seems easier than other schemas, the model initialization and update are indispensable to encode the template information into the network.

	There was an early tracking algorithm \cite{fan2010human} based on the ICD paradigm, which trained a CNN offline before tracking and fixed it afterward. However, this work can only handle pre-defined target object classes, e.g., humans, but not for general VOT. DLT \cite{wang2013learning} is the pioneering work of ICD trackers and end-to-end deep trackers for VOT. It first trained a stacked denoising autoencoder offline with extra data to learn generic image features and transferred the learned model to the online tracking process via online finetuning. DeepTrack \cite{li2014robust} proposed a target-specific CNN for object tracking without pre-training, where the CNN is trained incrementally during tracking with new examples obtained online. To cater to the date-poor problem, it enhanced the ordinary Stochastic Gradient Descent approach in CNN training with a temporal selection mechanism, generating positive and negative samples within different periods. SO-DLT \cite{wang2015transferring} exploited the power of CNN based on DLT by using CNN as the feature extractor. From then on, CNN became the main choice of the backbone. Hong \textit{et al.}\cite{hong2015online} explored the usage of a pretrained CNN, using an SVM layer as the classifier to end-to-end learn the network by predicting the target saliency maps. STCT \cite{wang2016stct} transferred rich features of pre-trained deep CNNs for online tracking by casting online training CNN as learning ensembles to remove feature correlation and avoid over-fitting effectively. UCT \cite{uct2017} proposed to treat both the feature extractor and the ridge regression as convolution operations inside a unified convolutional network. MAML~\cite{wang2020tracking} was a representative instance detection method, which retrofitted robust detectors like FCOS~\cite{tian2019fcos} and RetinaNet~\cite{lin2017focal} to trackers with proper initialization by model-agnostic meta-learning strategy.
	MDNet \cite{nam2016learning} is a typical milestone of ICD trackers, which proposed a multi-domain learning framework based on CNNs. The network is pretrained on visual tracking data for obtaining a generic target representation and is updated online for learning target-specific features. Following MDNet, there are many trackers proposed to improve its performance from multiple aspects like the attentions (SANet \cite{fan2017sanet}), prediction head (DRL-IS \cite{ren2018deep}, ADNet \cite{yun2017action}, BranchOut \cite{han2017branchout}), class imbalance (DSLT \cite{lu2018deep}, VITAL \cite{song2018vital}), acceleration (DeepAttTrack \cite{pu2018deep}, RTMDNet \cite{jung2018real}, Meta-Tracker \cite{park2018meta}), and refinement (ACT \cite{chen2018real}).
 \begin{figure}[tbp]
		\begin{center}
			\includegraphics[width=\columnwidth]{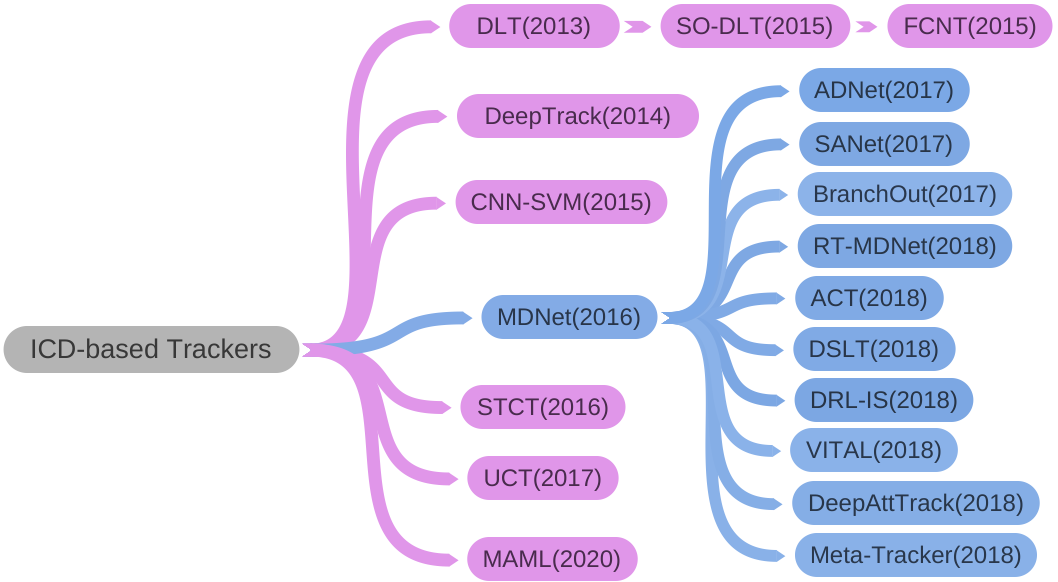}
		\end{center}
   \vspace{-15.5pt}
		\caption{Develop lineage of representative ICD-based RGB trackers.}
   \vspace{-10.5pt}
		\label{figure:icd}
	\end{figure}
	
	We present the typical ICD trackers with their development process in Fig.~\ref{figure:icd}. Note that trackers in this category usually finetune a pretrained feature extraction network with the ground truth of the target in the first frame and continuously update the network with tracked regions. Therefore, the ICD models are end-to-end trainable, which are first trained offline to ensure a good initialization and updated online to adapt to specific targets. In fact, the ICD-based trackers mainly emerged in a data-insufficient time, where large-scale tracking datasets haven't been proposed. Hence, most of them are dedicated to research on how to learn a CNN under the data-poor condition due to the lack of sufficient training samples.

\subsubsection{One-stream Transformers}\label{sec:unifiedTrans}
	\begin{figure}[tbp]
		\begin{center}
			\includegraphics[width=\columnwidth]{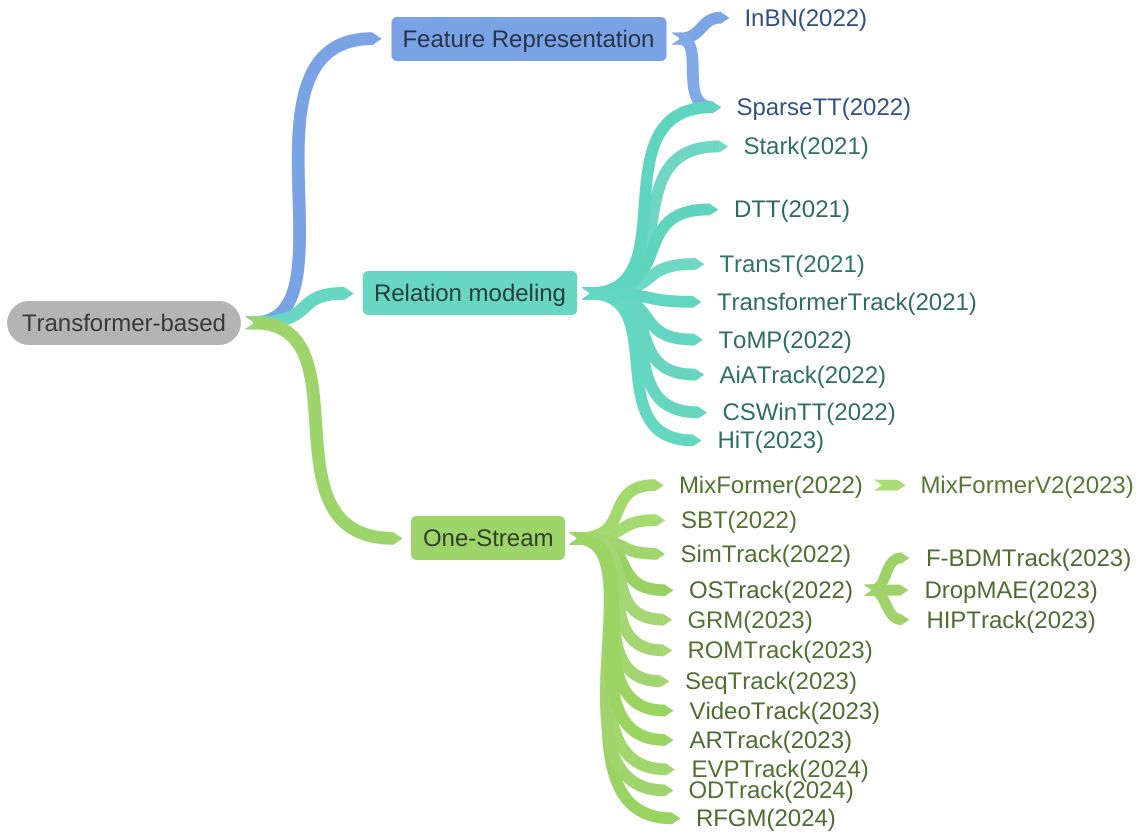}
		\end{center}
   \vspace{-10.5pt}
		\caption{\textcolor{black}{Transformer-based VOT methods of RGB modality.}}
  \vspace{-10.5pt}
		\label{figure:transformers}
	\end{figure}
	Before the vision Transformer appeared, trackers of the last five years were mainly dominated by the DCF and Siamese methods, where the feature extractors are usually shared between the template and search region. Then, a matcher or a model predictor will be employed to embed the template information into the search representations. In these pipelines, the template and search region interaction happens upon their extracted high-level features. 
		
    \begin{figure*}[tbp]
		\begin{center}
			\includegraphics[width=2\columnwidth]{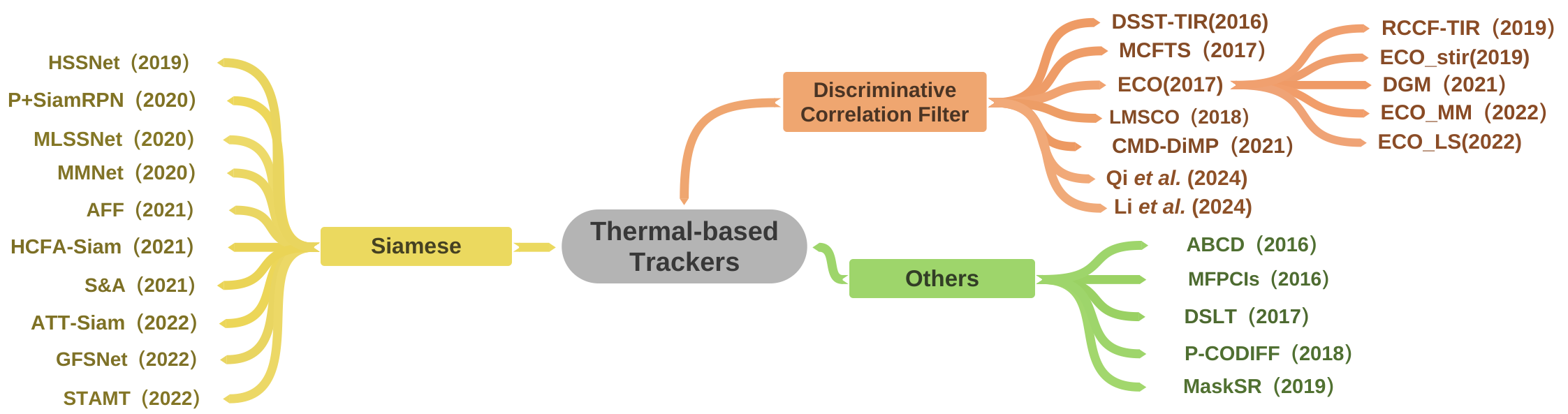}
		\end{center}
		\caption{\textcolor{black}{Develop lineage of Thermal-based trackers. The methods are split into Discriminative Correlation Filter, Siamese and Others.}}
  \vspace{-12.5pt}
		\label{figure:thermal}
	\end{figure*}
Transformer has the attractive characteristic of relation modeling and feature extraction, which is demonstrated in vast vision tasks like segmentation~\cite{zheng2021rethinking,li2021mail}, detection~\cite{chen2021pix2seq}, video classification~\cite{wang2021actionclip,fan2021multiscale}, vision-language learning~\cite{kim2021vilt} and so on. In the recent two years (2021-2022), Transformer has also shown its power in VOT, as shown in Fig.~\ref{figure:transformers}.
Initially, Transformer in VOT is used for either relation modeling (replacing the conventional cross-correlation operation) like TransformerTrack~\cite{wang2021transformer}, Stark~\cite{stark2021}, CSWinTT~\cite{CSWinTT2022}, TransT~\cite{transT2021}, ToMP~\cite{mayer2022transforming}, DTT~\cite{DTT2021}, SparseTT~\cite{sparsett2022}, AiATrack~\cite{aiatrack2022} or feature representation (substituting the CNN backbones like ResNet) such as InBN~\cite{InBN2022} and SparseTT~\cite{sparsett2022}, both the two ways are demonstrated to be effective with large-margin improvements of performance. Note that although the methods mentioned here introduce Transformer, they are classified into the aforementioned categories based on their pipelines.
	
Recently, we have observed another appealing and emerging trend of VOT, where the previous extractor-matcher separated two-stream pipeline gradually becomes a One-Stream Transformer (OST) pipeline as shown in Fig.~\ref{figure:pipeline}(d). \textcolor{black}{Plenty of emerging trackers follow this OST paradim as shown in Fig~\ref{figure:transformers}, including MixFormer~\cite{mixformer2022}, MixFormer V2~\cite{mixformerv2}, OSTrack~\cite{ostrack2022}, SBT~\cite{sbt2022}, SimTrack~\cite{simtrack2022}, DropMAE~\cite{dropmae}, F-BDMTrack~\cite{yang2023foreground}, GRM~\cite{grm}, HIPTrack~\cite{hiptrack}, RFGM~\cite{rfgm}, ROMTrack~\cite{romtrack}, SeqTrack~\cite{seqtrack}, ARTrack~\cite{artrack}, VideoTrack~\cite{videotrack}, ODTrack~\cite{odtrack} and EVPTrack~\cite{evptrack}. }
\textcolor{black}{Specifically, leveraging the attention mechanism of Transformers, MixFormer~\cite{mixformer2022} proposes to extract target-specific features and facilitate interactions between the target and search regions concurrently using a mixed attention module, dispensing with conventional Siamese CNN feature extraction.} Similarly, in the same period, SBT~\cite{sbt2022} proposed a target-specific tracker based on the self-/cross-attention scheme, which deeply incorporated self-image feature extraction and cross-image feature correlation in multiple layers of the feature network. OSTrack~\cite{ostrack2022} unified the feature learning and relation modeling jointly within a Transformer, bridging the template-search image pairs with bidirectional information flows. SimTrack~\cite{simtrack2022} also leveraged a Transformer backbone for joint feature learning and interaction by serializing the input template and search images and concatenating them directly before the one-branch Transformer backbone. \textcolor{black}{DropMAE~\cite{dropmae} investigated masked autoencoder video pre-training for VOT with the OST paradigm. F-BDMTrack~\cite{yang2023foreground} and GRM~\cite{grm} focused on solving the problem of insufficient consideration of foreground-background relationships. HIPTrack~\cite{hiptrack} and RFGM~\cite{rfgm} proposed to incorporate historical information into OST-based trackers. SeqTrack~\cite{seqtrack} and ARTrack~\cite{artrack} converted the four values of a bounding box into a sequence of discrete tokens and used a decoder to generate this sequence token-by-token. VideoTrack~\cite{videotrack}, ODTrack~\cite{odtrack}, and EVPTrack~\cite{evptrack} exploited the contextual relationships of video frames, i.e., the spatio-temporal information encoding of OST pipelines. These OST-based trackers have achieved state-of-the-art performance, indicating the potential of this paradigm.}

Generally, the RGB modality is the most commonly used modality for VOT in various real-world applications due to its easy accessibility and rich appearance information. Hence, RGB-based VOT has a long exploration history and has given birth to a large number of methods. Nevertheless, challenges still exist, such as appearance variations and background noise. In addition, the challenges of long-term stability and accuracy-efficiency dilemma remain unresolved. Robust feature representation and strong relational modeling are key to solving these problems, which have been the most popular research topics for a long time. The recently emerged OST paradigm is promising because it leverages Transformer's feature representation and relational modeling capabilities to achieve superior performance while maintaining simplicity. Besides, to ensure efficiency, powerful but lightweight backbone architectures are always needed to accomplish this task. Moreover, specific issues like redetection modules for long-term tracking, model update strategies for different paradigms, and integration with other modalities are open, interesting, and valuable directions to explore for RGB-based VOT.

\subsection{Thermal infrared-based Trackers}
	Thermal infrared (TIR) tracking refers to object tracking based on TIR video sequences, which are the radiated electromagnetic radiations within a specific wavelength spectrum and emitted by objects with temperatures above absolute zero. This modality can effectively handle some harsh situations like the camouflage and the poor illumination situations due to its imaging principle, enabling robust VOT under these conditions. 
	Compared with RGB modality, TIR is helpful in privacy preserving and more robust to illumination changes. The application scenarios of TIR-based VOT include marine rescue, video surveillance, and night driving assistance.
	Therefore, TIR-based tracking has become a popular research topic considering the above benefits and the development of infrared equipment. However, since the TIR modality has no visible colors and textures, it is hard to provide discriminative feature representations and resist the distractors of similar objects, making TIR-based VOT challenging.

    Traditional TIR-based algorithms often combine handcrafted features with classical machine learning techniques. 
    For instance, MILTracker~\cite{MILTracker} combined multiple instance learning with an adaptive motion prediction method to boost the tracker's efficiency.  
    RLRST~\cite{RLRST} created the multiple feature pseudo-color images with its kernel density estimation mechanism to monitor the characteristics from feature maps. ABCD~\cite{ABCD} employed template-based trackers based on distribution fields for TIR tracking, exploiting background information for an online model update. MFPCIS~\cite{MFPCIS} sought to maximize the likelihood estimation for the residuals based on robust low-rank sparse representation. To overcome partial deformations, DSLT~\cite{DSLT} used structural SVM with a part-based matching method that integrated the co-difference feature of multiple parts. P-CODIFF~\cite{P-CODIFF} also used co-difference matrix and developed a part-based tracker that sampled at salient corner points. MaskSR~\cite{MaskSR} combined sparse coding together with high-level semantic features extracted by DNNs within a particle filter framework for TIR target tracking. 

	In recent years, TIR-based VOT methods mainly follow two mainstream paradigms: the DCF and Siamese schemas similar to the RGB tracking. We categorize the TIR trackers in Fig.~\ref{figure:thermal} and introduce the corresponding methods in the following. Besides, we report the results of TIR-based methods in Table~\ref{tab:thermal}.
\begin{table*}[!ht]
    \centering
    \vspace{-10.5pt}
        \caption{The performance of other thermal-based trackers on thermal datasets. (SR: success rate; PR: precision rate; EAO: expected average overlap; A: accuracy; R: robustness; FPS: mean speed, taking the maximum speed that can be achieved in the corresponding paper)}
        \vspace{-10.5pt}
    \label{tab:thermal}
    \footnotesize
    \scalebox{0.94}{\begin{tabular}{c c c|c c c|c c|c c c|c c|c c c}
    \hline
        \multicolumn{2}{c}{\multirow{2}*{Trackers}} & \multirow{2}*{Publication} & \multicolumn{3}{c|}{VOT-TIR 2015} & \multicolumn{2}{c|}{LSOTB-TIR} & \multicolumn{3}{c|}{VOTT-TIR 2017} & \multicolumn{2}{c|}{PTB-TIR} & \multicolumn{3}{c}{VOT-TIR 2017}  \\ \cline{4-16}
        \multicolumn{2}{c}{~} & ~ & EAO & A & R & SR & PR & EAO & A & R & PR & SR & EAO & A & R  \\ \hline
        \multirow{5}*{DCF} & MCFTS\cite{MCFTS}  & KBS2017 & 0.218 & 0.59 & 0.41 & 0.479 & 0.635 & - & - & - & 0.69 & 0.492 & - ~ & - & -  \\ 
        ~ & ECO-stir\cite{ECO-stir} & TIP2019 & - & - & - & 0.616 & 0.75 & - & - & - & 0.83 & 0.617 & \textbf{0.436} & 0.65 & 0.8 \\ 
        ~ & DGM\cite{DGM} & SPL 2021 & - & - & - & - & - & - & - & - & 0.848 & 0.639 & - & - & -  \\ 
        ~ & ECO\_MM\cite{ECO-MM} & TMM2022 & 0.303 & 0.64 & 0.24 & - & - & - & - & - & - & - & 0.29 & 0.61 & 2.31  \\ 
        ~ & ECO\_LS\cite{ECO-LS} & IVP2022 & 0.333 & 0.663 & 0.837 & - & - & \textbf{0.323} & 0.647 & 0.771 & \textbf{0.867} & \textbf{0.648} & - & - & -  \\ 
        ~ & \textcolor{black}{STFT-Net}\cite{qi2024exploring} & KBS2024 & - & - & - & \textbf{0.705} & 0.821 & - & - & - & - & - & - & - & -  \\ 
        ~ & \textcolor{black}{UTFCNet}\cite{li2024two} & ITGRS2024 & - & - & - & 0.647 & \textbf{0.857} & - & - & - & - & - & - & - & -  \\ 
        \hline
        \multirow{8}*{Siamese} 
        ~ & HSSNet\cite{HSSNet} & KBS2019 & 0.311 & 0.67 & 0.25 & 0.435 & 0.566 & 0.268 & 0.58 & 0.626 & 0.689 & 0.468 & 0.26 & 0.58 & 3.33 \\ 
        ~ & MLSSNet\cite{MLSSNet} & TMM2020 & 0.329 & 0.57 & 0.24 & 0.459 & 0.596 & 0.293 & 0.564 & 0.668 & 0.731 & 0.516 & 0.278 & 0.56 & 3.11  \\ 
        ~ & MMNet\cite{MMNet} & AAAI2020 & 0.344 & 0.61 & 0.21 & - & - & 0.307 & 0.62 & 0.677 & 0.783 & 0.557 & 0.32 & 0.58 & 2.90 \\ 
        ~ & AFF\cite{AFF} & ITME2021 & - & - & - & 0.669 & 0.8 & - & 0.504 & 0.71 & - & - & - & - & -  \\ 
        ~ & HCFA-Siam\cite{HCFA-Siam} & TIM2021 & - & 0.655 & - & - & - & - & - & - & - & - & - & - & -  \\ 
        ~ & S\&A\cite{S&A} & IP\&T2021 & 0.344 & 0.71 & 0.275 & 0.641 & 0.58 & - & - & - & 0.759 & 0.591 & - & - & -  \\ 
        ~ & GFSNet\cite{GFSNet} & IP\&T2022 & \textbf{0.365} & 0.67 & 0.21 & 0.683 & 0.798 & - & - & - & - & - & - & - & -  \\ 
        ~ & STAMT\cite{STAMT} & NC 2022 & - & - & - & 0.579 & 0.712 & - & - & - & 0.781 & 0.573 & - & - & -  \\ 
        \hline
        \multirow{2}*{others}  & ABCD\cite{ABCD} & CVPR2016 & - & 0.45 & 0.312 & - & - & - & - & - & - & - & - & - & -  \\ 
        ~ & MaskSR\cite{MaskSR} & RS 2019 & - & - & - & - & - & 0.26 & 0.517 & 0.586 & - & - & - & - & -  \\ \hline
    \end{tabular}}
     \vspace{-10.5pt}
\end{table*}

\subsubsection{Discriminative Correlation Filters}
	DCF paradigm has gained tremendous popularity and achieved significant success in VOT, as introduced in Section~\ref{sec:dcf}.
	Its basic principle is applying a correlation filter between the template and candidate regions, then selecting the maximum value in the response map to locate the target in the current frame. Many researchers have attempted to solve the TIR tracking based on DCF. For instance, WCF~\cite{WCF} introduced a weighted correlation filter into a tracking-by-detection framework to fuse different target features and locate the targets. TBOOST~\cite{TBOOST} used a continuously switching mechanism to select a set of base trackers constructed from several dynamical MOSSE~\cite{bolme2010visual} filters. 
	
  In recent years, some DCF-based trackers built on deep learning techniques. For instance, DSST-TIR~\cite{DSST-TIR} integrated the TIR-specific features learned by a CNN architecture into the DCF paradigm. 
  MCFTS~\cite{MCFTS} proposed an ensemble method that merged the features of the multiple convolution layers by multiple weak trackers constructed by a correlation filter. 
  LMSCO~\cite{LMSCO} integrated DCF with structured SVM and employed spatial regularization and implicit interpolation to obtain continuous deep feature maps.
  One of the most significant milestones in DCF-based deep trackers is Efficient Convolution Operators (ECO)~\cite{ECO}, which is a RGB-modal tracker that proposed a factorized convolution operator to dramatically reduce the number of parameters for using DNN-based features. The most recent TIR tracking methods, like RCCF-TIR~\cite{RCCF-TIR}, ECO-stir~\cite{ECO-stir}, DGM~\cite{DGM}, ECO-MM~\cite{ECO-MM} and ECO-LS~\cite{ECO-LS}, were all developed based on ECO. Specifically, RCCF-TIR jointed continuous correlation filter with a feature fusion strategy based on average peak-to-correlation energy to increase robustness towards various challenges. ECO-stir tried translating the abundantly labeled RGB data to synthetic TIR data, compensating for the data-poor situation for TIR-based VOT. DGM improved ECO with a long-term cross-frame graph memory feature to enrich deep features with time-variant TIR object appearance information.
  ECO-MM introduced a dual-level feature model which contained both target-specific features and fine-grained similarity information. ECO-LS proposed to obtain the object contour and then continuously encoded the historical level set function, which significantly improved the convergence speed. CMD-DiMP~\cite{CMD} was a recent method that built on a newer and stronger RGB-based VOT baseline DiMP~\cite{DiMP}. It employed cross-modal distillation to distill representations from the RGB modality to the TIR modality, taking advantage of a large amount of unlabeled paired RGB-TIR data. \textcolor{black}{Qi \textit{et al.}~\cite{qi2024exploring} proposed STFT-Net that involved a Transformer-based encoder–decoder to fuse spatio-temporal information. Li~\textit{et al.}~\cite{li2024two} proposed UTFCNet which was a two-stage spatio-temporal feature correlation network to solve the problem of similar target distractor and background occlusion problem.}
  
  In general, the DCF-based TIR trackers have achieved successful performance on benchmark datasets, such as VOT2015~\cite{VOT-2015} and PTB-TIR~\cite{PTB-TIR}, as shown in Table~\ref{tab:thermal}. 

\subsubsection{Siamese Trackers}
     Subsequently, we will review some TIR trackers based on the Siamese paradigm as shown in Fig.~\ref{figure:pipeline}(b). The core idea of this schema is to transform the tracking task into the template-search similarity measuring problem. 
     
     As a pioneer work, HSSNet~\cite{HSSNet} first introduced the Siamese CNN into the TIR tracking, which used the Siamese network to provide hierarchical features for improving the discriminative ability of TIR tracking. P+SiamRPN~\cite{P+SiamRPN} introduced SiameseRPN~\cite{siamrpn2018} into TIR tracking with a CNN-based exemplar prediction model that fully utilizes spatial information and temporal information around a TIR target. 
     Unlike RGB images, TIR images do not have rich color and texture information. Therefore, intra-class TIR objects usually have similar visual and semantic patterns, causing the distractor drifting problem. This is a key challenge of TIR-based VOT. Several methods were aware of this problem, like MLSSNet~\cite{MLSSNet}, MMNet~\cite{MMNet}, AFF~\cite{AFF}, HCFA-Siam~\cite{HCFA-Siam} and STAMT~\cite{STAMT}.
     Specifically, MLSSNet proposed a multi-level TIR tracker, which not only used the global semantic features but also captured local structural representations of TIR objects to provide discriminative features for intra-class object distinguishing.
     MMNet explored a multi-task learning framework for TIR tracking, which learned the TIR-specific discriminative features and fine-grained correlation features with the guidance of an auxiliary classification network. AFF integrated SiamRPN++~\cite{siamrpn++2019} with an adaptive feature fusion module to fuse the features of different layers, enriching the features from different semantic levels. 
     Similarly, HCFA-Siam proposed to fuse the shallow spatial information and deep semantic information by a hierarchical architecture and employed the channel-attention and an adaptive updated negative template pool to distinguish the TIR target from similar appearance distractors.
     STAMT designed a structural target-aware model that assigned more attention to the target area to avoid interference with similar objects.
     
     Except for the distraction problem, there are some other research directions for Siamese-based TIR trackers. For example, GFSNet~\cite{GFSNet} focused on the network generalization problem, introducing the self-adaption structure into the Siamese network. ATT-Siam~\cite{li2022infrared} built a small TIR object tracking framework on SiamFC~\cite{siamfc} by applying the spatial and channel attention mechanisms. Yao \textit{et al.}~\cite{S&A} designed an enhanced Siamese network to solve the scale and appearance variation problem by introducing the dilated convolution module and a template library update strategy. The Siamese paradigm has been a remarkable success in TIR-based VOT due to its simplicity and efficiency. It still has room for improvement, such as using more powerful backbone architectures like Transformer.

     In general, the TIR modality is a promising candidate for working under some low-light situations since its imaging is relatively independent of the illumination conditions. There has been a great interest in TIR-based VOT research for practical applications. However, there are still two major challenges with TIR-based VOT. First, since TIR images do not have appearance information like color and texture, it is difficult to distinguish intra-class TIR objects, causing the distractor drift problem. Second, most DNN-based TIR trackers use RGB-pretrained models, which limit the feature discriminability due to the modality distinctions. Therefore, researchers can explore the two challenges for the future development of TIR tracking. In addition, stronger backbones, better schemas, and the combination with RGB images can be investigated.
\begin{table*}[h]
    \centering
        \caption{The performance of different LiDAR-based trackers on the KITTI dataset. The results are evaluated by \textit{Success}/\textit{Precision}.}
         \vspace{-10.5pt}
    \label{tab:lidar_kitti}
    \footnotesize
    \begin{tabular}{c|c|ccccc}
    \hline
        \multirow{2}*{Trackers}  &\multirow{2}*{publication} & \multicolumn{5}{c}{KITTI}     \\
        & &Car(6424) &Pedestrian(6088) &Van(1248) &Cyclist(308) &Mean(14068) \\
          \hline
         SC3D~\cite{SC3D} &CVPR2019 &41.3/57.9 &18.2/37.8 &40.4/47.0 &41.5/70.4 &31.2/48.5   \\
         P2B~\cite{p2b} &CVPR2020  &56.2/72.8 &28.7/49.6 &40.8/48.4 &32.1/44.7 &42.4/60.0   \\
         BEV~\cite{bev} &None2020 &64.0/83.5 &17.89/47.81 &- &43.23/81.15 &-\\
         MLVSNet~\cite{mlvsnet} &ICCV2021 &56.0/74.0 &34.1/61.1 &52.0/61.4 &34.3/44.5 &45.7/66.6  \\
         3D-SiameseRPN~\cite{3dsiameserpn} &IEEE Sensors &58.2/76.2 &35.2/56.2 &45.6/52.8 &36.1/49.0 &46.6/64.9  \\
         LTTR~\cite{lttr} &BMVC2021 &65.0/77.1 &33.2/56.8 &35.8/45.6 &66.2/89.9 & 48.7/65.8  \\
         PTT~\cite{ptt} &IROS2021 &67.8/81.8 &44.9/72.0 &43.6/52.5 &37.2/47.3 &55.1/74.2 \\
         PointSiamRCNN~\cite{pointsiamrcnn} &IROS2021 &51.5/68.9 &- &- &- &- \\
         BAT~\cite{bat} &ICCV2021 &60.5/77.7 &42.1/70.1 &52.4/67.0 &33.7/45.4 &51.2/72.8\\
         V2B~\cite{v2b} &NeurIPS2021 &70.5/81.3 &48.3/73.5 &50.1/58.0 &40.8/49.7 &58.4/75.2 \\
         VPIT~\cite{oleksiienko2022vpit} &None2022 &50.5/64.5 &- &- &- &- \\
         GPT~\cite{park2022gpt} &AAAI2022 & 59.1/75.6 &35.2/63.6 &49.6/60.6 &34.3/46.3 &47.4/68.4  \\
         PTTR~\cite{zhou2022pttr} &CVPR2022 &65.2/77.4 &50.9/81.6 &52.5/61.8 &65.1/90.5 &58.4/77.8 \\
         PTTR++~\cite{pttr++} &None2022 &73.4/84.5 &55.2/84.7 &55.1/62.2 &71.6/92.8 &63.9/82.8 \\
         M2Track~\cite{m2track} &CVPR2022 &65.5/80.8 &61.5/88.2 &53.8/70.7 &73.2/93.5 &62.9/83.4  \\
         STNet~\cite{stnet} &ECCV2022 &72.1/84.0 &49.9/77.2 &58.0/70.6 &73.5/93.7 &61.3/80.1\\
         \textcolor{black}{DMT}~\cite{xia2023lightweight} &T-ITS2023 &66.4/79.4 &48.1/77.9 &53.3/65.6 &70.4/92.6 &55.1/75.8\\
         \textcolor{black}{GLT-T}~\cite{gltt} &AAAI2023 &68.2/82.1 &52.4/78.8 &52.6/62.9 &68.9/92.1 &60.1/79.3 \\
         \textcolor{black}{OSP2B}~\cite{nie2023osp2b} &IJCAI2023 &67.5/82.3 &53.6/85.1 &56.3/66.2 &65.6/90.5 &60.5/82.3  \\
         \textcolor{black}{CXTrack}~\cite{xu2023cxtrack} &CVPR2023 &69.1/81.6 &67.0/91.5 &60.0/71.8 &74.2/94.3 &67.5/85.3 \\
         \textcolor{black}{CorpNet}~\cite{cropnet} &CVPRW2023 &73.6/84.1 &55.6/82.4 & 58.7/66.5 &74.3/94.2 &64.5/82.0\\
         \textcolor{black}{SyncTrack}~\cite{ma2023synchronize} &ICCV2023 &73.3/85.0 &54.7/80.5 &60.3/70.0 &73.1/93.8 &64.1/81.9 \\
         \textcolor{black}{SCVTrack}~\cite{scvtrack} &None2023 &68.7/81.9 &62.0/89.1 &58.6/72.8 &77.4/94.4 &65.1/84.5 \\
         \textcolor{black}{BEVTrack}~\cite{yang2023bevtrack} &None2023 &71.7/84.1 &68.4/92.7 &\textbf{65.7}/76.2 &74.6/\textbf{94.7}  &69.8/87.4 \\
         \textcolor{black}{M3SOT}~\cite{liu2023m3sot} &None2023 &\textbf{75.9}/\textbf{87.4} &66.6/92.5 &59.4/74.7 &70.3/93.4 &70.3/\textbf{88.6} \\
         \textcolor{black}{StreamTrack}~\cite{streamtrack} &None2023 &72.6/83.7 &\textbf{70.5}/94.7 &61.0/76.9 &\textbf{78.1}/94.6 &70.8/88.1 \\
         \textcolor{black}{MTMTrack}~\cite{mtmtrack} &RAL2023 &73.1/84.5 &70.4/\textbf{95.1} &60.8/74.2 &76.7/94.6 &\textbf{70.9}/88.4 \\
         \textcolor{black}{MoCUT}~\cite{cutrack} &ICLR2024 &67.6/80.5 &63.3/90.0 &64.5/\textbf{78.8} &76.7/94.2 &65.8/85.0  \\
         \hline
    \end{tabular}
     \vspace{-2.5pt}
\end{table*}
	\subsection{LiDAR-based Trackers}
	LiDAR modality refers to the point clouds generated by LiDAR sensors, which can directly sense the geometry and depth accurately. Trackers based on LiDAR can intrinsically capture the geometry information of targets. Thus they could provide the 3D locations and bounding boxes information of targets. Moreover, the point clouds generated by LiDAR sensors are less sensitive to environmental changes like light and weather conditions, thereby can be a robust modality for performing object tracking. However, the LiDAR modality has the problems of sparsity, density variance, and implicit geometric features in data locality, which bring extra challenges for the tracking framework designing. Therefore, the LiDAR modality based methods require relatively complex data pre-process and is sensitive to the sampling quality of raw data. The points sampling is more difficult and expensive comparing with RGB modality. 
  \begin{figure}[tbp]
		\begin{center}			\includegraphics[width=\columnwidth]{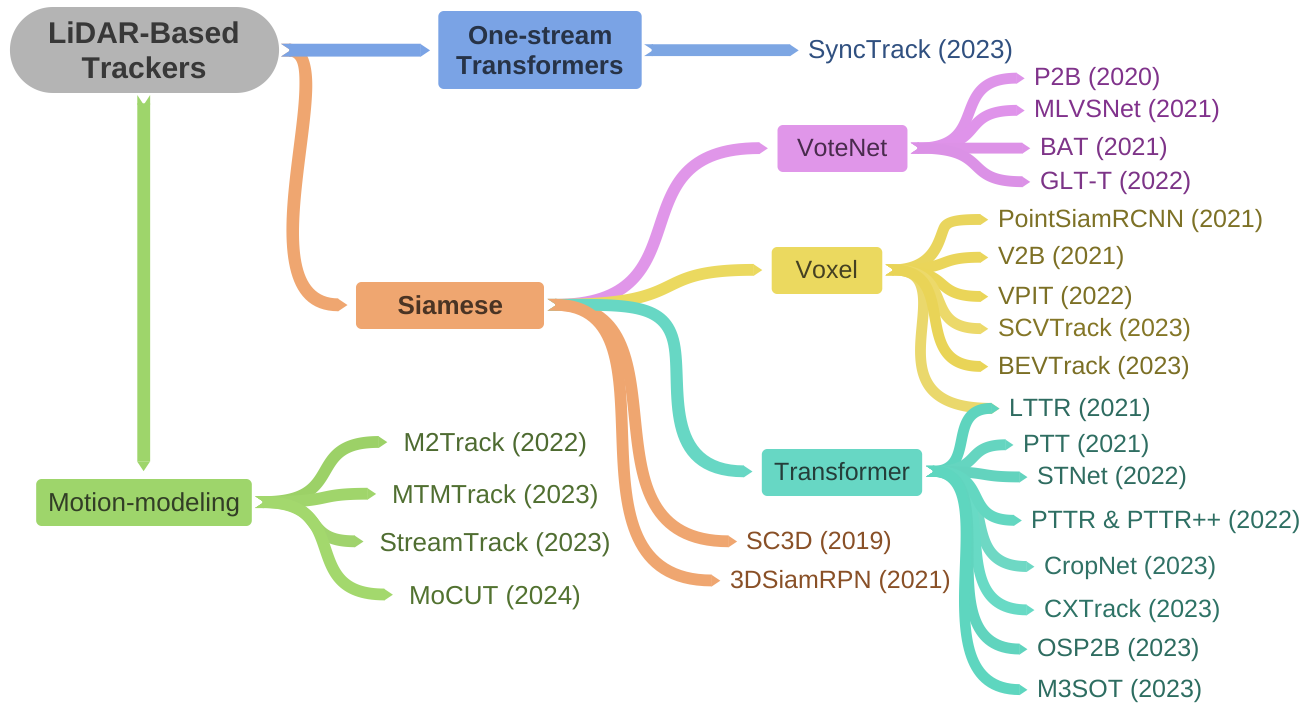}
		\end{center}
      \vspace{-5.5pt}
		\caption{\textcolor{black}{Develop lineage of LiDAR-based trackers. The methods are split into Siamese, Motion-Modeling and Single-Branch framework, where the first one has dominated till now. Then, we make fine-grained categorization to conclude the common labels of related methods, like using VoteNet, voxelizing the features and building models based on Transformers.}}
  \vspace{-12.5pt}
		\label{figure:lidar}
	\end{figure}
	
	The LiDAR-based 3D VOT is a new task that emerged in recent years, and thus all models are designed based on deep learning methods. \textcolor{black}{From the pioneering tracker SC3D~\cite{SC3D}, most of the tracking methods depend on the Siamese paradigm and appearance matching, which is quite similar to that in RGB tracking methods (Fig.~\ref{figure:pipeline}(b))}. \textcolor{black}{Specifically, a shared Siamese backbone is leveraged for point clouds feature extracting of template and search regions. The extracted features are then fused to generate seed points for subsequent target localization.} \textcolor{black}{Recently, few attempts have been made to explore efficient and effective LiDAR network design besides appearance matching and Siamese structure such as motion-modeling and one-stream Transformers. 
    Therefore, we introduce the current methods from three perspectives, including Siamese structure, motion-modeling and one-stream Transformers, as shown in Fig.~\ref{figure:lidar}.}
	\begin{table*}[h]
    \centering
     \caption{The performance of different LiDAR-based trackers on the nuScenes dataset. The results are evaluated by \textit{Success}/\textit{Precision}.}
    \label{tab:lidar_nuscenes}
    \scriptsize
    \resizebox{\textwidth}{!}{
    \begin{tabular}{cc|cccccccc}
    \hline
    &Trackers &SC3D~\cite{SC3D} &P2B~\cite{p2b} & BAT~\cite{bat}  & V2B~\cite{v2b} &PTTR~\cite{zhou2022pttr} & M2Track~\cite{m2track} &PTTR++~\cite{pttr++} &STNet~\cite{stnet} \\ 
    &publication & CVPR2019 &CVPR2020 &ICCV2021 &NeurIPS2021 &CVPR2022 &CVPR2022 &None2022 &ECCV2022 \\\hline
     \multirow{6}*{\rotatebox{90}{nuScenes1}} &Car(64159) &22.31 / 21.93 &38.81 / 43.18 &40.73 / 43.29 &- &51.89 / 58.61 &55.85 / 65.09 &59.96 / 66.73 &- \\
     &Pedestrian(3322) &11.29 / 12.65 &28.39 / 52.24 &28.83 / 53.32 &- &29.90 / 45.09  &32.10 / 60.92 &32.49 / 50.50 &- \\
     &Truck(13587) & 30.67 / 27.73 &42.95 / 41.59 &45.34 / 42.58 &- &45.30 / 44.74 &57.36 / 59.54 &59.85 / 61.20 &-\\
     &Trailer(3352) &35.28 / 28.12 &48.96 / 40.05 &52.59 / 44.89 &- &45.87 / 38.36 &57.61 / 58.26 &54.51 / 50.22 &- \\
     &Bus(2953) & 29.35 / 24.08 &32.95 / 27.41 &35.44 / 28.01 &- &43.14 / 37.74  &51.39 / 51.44 &51.39 / 51.44 &- \\
     &Mean(117278) &20.70 / 20.20  &36.48 / 45.08 &38.10 / 45.71 &- &44.50 / 52.07 &49.23 / 62.73 &51.86 / 60.63  &- \\ \hline
    \multirow{5}*{\rotatebox{90}{nuScenes2}} &Car(15578) &25.0 / 27.1 &27.0 / 29.2 &22.5 / 24.1 &31.3 / 35.1 &- &- &- &32.2/36.1 \\
    &Pedestrian(8019) &14.2 / 16.2 &15.9 / 22.0 &17.3 / 24.5 &17.3 / 23.4 &- &- &- &19.1 / 27.2 \\
    &Truck(3710) &25.7 / 21.9 &21.5 / 16.2  &19.3 / 15.8 &21.7 / 16.7 &- &- &- &22.3 / 16.8 \\
    &Bicycle(501) &17.0 / 18.2 &20.0 / 26.4 &17.0 / 18.8 &22.2 / 19.1 &- &- &- &21.2 / 29.2 \\
    &Mean(27808) &21.8 / 23.1 &22.9 / 25.3 &20.5 / 23.0 &25.8 / 29.0 &- &- &- &26.9 / 30.8\\ \hline \hline
    &Trackers &\textcolor{black}{CorpNet~\cite{cropnet}} &\textcolor{black}{SyncTrack}~\cite{ma2023synchronize} & \textcolor{black}{CXTrack}~\cite{xu2023cxtrack}  & \textcolor{black}{SCVTrack}~\cite{scvtrack} &\textcolor{black}{M3SOT}~\cite{liu2023m3sot} & \textcolor{black}{StreamTrack}~\cite{streamtrack} &\textcolor{black}{MTMTrack}~\cite{mtmtrack} &\textcolor{black}{MoCUT}~\cite{cutrack} \\ 
    &publication & CVPRW2023 &ICCV2023 &CVPR2023 &None2023 &None2023 &None2023 &RAL2023 &ICLR2024 \\\hline
    \multirow{6}*{\rotatebox{90}{\textcolor{black}{nuScenes1}}} &Car(64159) &-  &-  &-  &58.9 / 67.7 &- &62.05 / 70.81 &57.05 / 65.94 &57.32 / 66.01 \\
     &Pedestrian(3322) &-  &- &-  &34.5 / 61.5 &-  &38.43 / 68.58 &37.08 / 72.80 &33.47 / 63.12 \\
     &Truck(13587) & -  &-  &- &60.6 / 61.4 &- &64.67 / 66.60 &59.37 / 63.69 &61.75 / 64.38\\
     &Trailer(3352) &-  &-  &- &59.5 / 60.1 &- &66.67 / 64.27 &59.73 / 60.44 &60.90 / 61.84 \\
     &Bus(2953) & - &-  &- &54.3 / 53.6 &-  &60.66 / 59.74 &55.46 / 54.31 &57.39 / 56.07 \\
     &Mean(117278) &-   &-  &- &52.1 / 64.7 &- &55.75 / 69.22 &51.70 / 67.17  &51.19 / 64.63 \\ \hline
    \multirow{5}*{\rotatebox{90}{\textcolor{black}{nuScenes2}}} &Car(15578) &35.0 / 38.4 &36.7 / 38.1 &29.6 / 33.4 &- &34.9 / 39.9 &- &- &- \\
    &Pedestrian(8019) &21.3 / 33.6 &19.1 / 27.8 &20.4 / 32.9 &- &23.3 / 25.6 &- &- &- \\
    &Truck(3710) &39.7 / 36.3 &39.4 / 38.6  &27.6 / 20.8 &- &30.4 / 27.0 &- &- &- \\
    &Bicycle(501) &26.9 / 43.5 &23.8 / 30.4 &18.5 / 26.8 &- &16.5 / 22.6 &- &- &- \\
    &Mean(27808) &31.8 / 36.8 &31.8 / 35.1 &26.5 / 31.5 &- &30.6 / 33.7 &- &- &-  \\ \hline
    
    \end{tabular}
    }
\end{table*}
	\subsubsection{Siamese Trackers}
	Siamese is the prevalent pipeline in LiDAR-based VOT. It follows the idea of a siamese backbone structure and similarity/correlation matching as that in RGB-based VOT. Differently, the 3D trackers pay more attention to the sparsity of point clouds and aim to capture the geometric information contained in point clouds for localization and classification. SC3D~\cite{SC3D} is the first one definite the 3D VOT task. It adopted a shared backbone consisting of 1D convolution blocks for feature extracting and matched the cosine similarity of extracted features between the candidates and target. A regularization was also utilized to regularize the training with the shape completion network, embedding the shape information into the latent representation that could be useful in the discrimination process. The decoder of SC3D was made up of two fully connected layers that decode the latent features into the reconstructed points. SC3D leveraged the filters to perform real-time tracking when doing inferences, providing the candidates for the tracker based on Kalman Filters, Particle Filters, and Gaussian Mixture Models. P2B~\cite{p2b} improved the SC3D by proposing an end-to-end trainable framework in a tracking-by-detecting scheme. The VoteNet~\cite{votenet} was used for proposal generation, and the candidate with the highest scores was chosen as the target. P2B decreased the training time and improved the inference speed compared to the SC3D. Numerous methods followed P2B and made improvements based on it. For example, MLVSNet~\cite{mlvsnet} used
    the CBAM module~\cite{woo2018cbam} to enhance the vote cluster features with channel attention and spatial attention. BAT~\cite{bat} improved the P2B with a box-aware feature fusion module, generating better target-specific search areas in a box-aware manner. The object features were enhanced by encoding the bounding box-related information in the first frame into the features. Besides, 3D-SiamRPN~\cite{3dsiameserpn} used a region proposal subnetwork for the proposal generation of tracking. 
    
    All the above methods relied only on the point-wise features, 
    while recent researches explore the usage of point clouds voxelization in trackers.
    V2B~\cite{v2b} voxelized the encoded point-wise features by averaging the 3D coordinates and features of the points into the same voxel bin. Then, the voxel-to-BEV target localization network was designed to regress the center of the target both on a 2D map and \textit{z}-axis dimension in an anchor-free manner. Moreover, the VPIT was the first method using voxel pseudo images for 3D VOT. The pillar-based voxelization was adopted to convert the 3D point clouds into 2D images, and these kinds of pseudo 2D images were fed into the 2D Siamese visual trackers. BEV coordinates were also considered as the unified view of creating the 2D pseudo images. Lastly, PointSiamRCNN~\cite{pointsiamrcnn} combined the volumetric representation with the regional proposal network in the tracker design, utilizing the 2D tracking head for generating high-quality 3D proposals based on the BEV feature map, which was quite similar to the idea of V2B~\cite{v2b}. \textcolor{black}{The SCVTrack~\cite{scvtrack} combines voxelization representation with shape completion, harnessing the historic frames to tackle the sparsity and incompleteness of point clouds. 
    The BEVTrack~\cite{yang2023bevtrack} leveraged the VoxelNet to extract the voxelized features, then modeled the relations between consecutive frames with a BEV-based motion network.}  
   
    In recent years, the success of Vision Transformer in both 2D~\cite{dosovitskiy2020image, deit} and 3D vision~\cite{zhao2021point, guo2021pct} stimulates numerous attempts to embed the Transformers into the siamese tracker designing. LTTR~\cite{lttr} proposed a Transformer architecture exploring both the intra- and inter-relations of the same and different point clouds. Self-attention and cross-attention modules were utilized in the feature matching of the model. PTT~\cite{ptt} was similar to the LTTR in adopting the Transformer blocks for the feature matching, weighted the features based on the attention after calculating the template and search region features cosine similarity to improve the tracking performance. PTT embedded the Transformer blocks into the seed voting stage and proposals generating network to enhance the features. PTTR~\cite{zhou2022pttr} was a Transformer-based 3D tracking framework that follows a coarse-to-fine tracking paradigm to refine the coarse global prediction with pooling operations. They designed novel Point Relation Transformer modules to aggregate and match features effectively. Moreover, PTTR improved the point cloud sampling with relation-aware modules to consider the template-relevant relations when doing search region points sampling. PTTR++~\cite{pttr++} improved the original PTTR~\cite{zhou2022pttr} with BEV representations to complement the original point-wise representations, constructing the point-BEV fusion with cross-view feature mapping and selective feature fusion modules. Subsequently, STNet~\cite{stnet} built a pure Transformer backbone to replace the PointNet++ backbone for feature extracting. A standalone Transformer-based matching network which consisted of self-attention modules and cross-attention modules, was proposed to augment the features for decoding. \textcolor{black}{CXTrack~\cite{xu2023cxtrack} proposed a target-centric Transformer architecture to aggregate both target and contextual information. Among them, some works used the cross-attention mechanism for feature fusion among the template and search region. CorpNet~\cite{cropnet} harnessed the self-attention modules to extract features, and cross-attention to perform multi-level feature interaction between the template and search region. Feng et al.~\cite{feng2023multi} also proposed to utilize cross-attention to fuse the two kinds of features based on the pillar representation. }

    \subsubsection{\textcolor{black}{Motion-modeling Trackers}}
    \textcolor{black}{The motion information between successive 3D frames serves as a strong complement for efficient object tracking. Motion modeling combines point clouds from consecutive frames as inputs to retain motion information for the network. By utilizing segmented foreground points, it estimates motion to predict relative movement, resulting in exceptional tracking performance. M2Track~\cite{m2track} proposed a novel motion-centric paradigm for real-time 3D point clouds VOT without any appearance matching. M2Track was made up of two stages and at the first stage, the model took in the target points and obtained a coarse bounding box at the current 3D frame via motion prediction and transformation. In the second stage, the model refined the coarse bounding box with a motion-assisted shape completion network. MTM-Tracker~\cite{mtmtrack} improved motion modeling by combining with appearance matching. It first located the target coarsely using historical boxes as motion cues. Subsequently, MTM-Tracker incorporated a feature interaction module to derive motion-sensitive features from successive point clouds, aligning them to enhance target trajectory. StreamTrack~\cite{streamtrack} focused on the long-range continuous motion property of tracked objects and proposed to fetch multiple historic features as a memory bank to interact with the current frame. MoCUT~\cite{cutrack} built a category-unified model based on motion-centric 3D SOT paradigms, which followed previous works by concatenating point clouds from historic frames and segmenting the foreground points. However, MoCUT was trained and tested on all categories with a single model, which is different from the category-specific models in previous works. }
    
    \subsubsection{\textcolor{black}{One-stream Transformers}}
\begin{table}[h]
    \centering
    \caption{The performance of different LiDAR-based trackers on the Waymo Open dataset. The results are evaluated by \textit{Success}/\textit{Precision}.}
    \label{tab:lidar_waymo}
    \resizebox{\columnwidth}{!}{
    \begin{tabular}{c|c|c|ccc}
    \hline
         &\multirow{2}*{Trackers} &\multirow{2}*{publication} &\multicolumn{3}{c}{Waymo Open}  \\
         & & &Vehicle &Pedestrian &Mean\\ \hline
    \multirow{3}{*}{\rotatebox{90}{S1}}&P2B~\cite{p2b} &CVPR2020 &28.32 / 35.41 &15.60 / 29.56 &24.18 / 33.51\\
    &BAT~\cite{bat} &ICCV2021 &35.62 / 44.15 &22.05 / 36.79 &31.20 / 41.75 \\
    &M2Track~\cite{m2track} &CVPR2022 &42.62 / 61.64 &42.10 / 67.31 &43.13 / 63.48 \\
    &\textcolor{black}{SCVTrack}~\cite{scvtrack} &None2023 &46.4 / 63.0 &44.1 / 68.2 &45.7 / 64.7\\
    &\textcolor{black}{StreamTrack}~\cite{streamtrack} &None2023 &60.23 / 72.61 &47.07 / 70.44 &55.95 / 71.90 \\
    \hline
    \multirow{3}{*}{\rotatebox{90}{S2}} &SC3D~\cite{SC3D} &CVPR2019 &46.5 / 52.7 &26.4 / 37.8 &- \\
    &P2B~\cite{p2b} &CVPR2020 &55.7 / 62.2 & 35.3 / 54.9 &- \\
    &PTTR~\cite{zhou2022pttr} &CVPR2022 &58.7 / 65.2 &49.0 / 69.1 &- \\ 
    &\textcolor{black}{OSP2B}~\cite{nie2023osp2b} &IJCAI2023  &59.2 / 67.3 &46.6 / 67.4 & - \\ \hline
    \multirow{4}{*}{\rotatebox{90}{S3}} &P2B~\cite{p2b} &CVPR2020 &52.6 / 61.7 &17.9 / 30.1 &33.0 / 43.8\\
     &BAT~\cite{bat} &ICCV2021 &54.7 / 62.7 &18.2 / 30.3 &34.1 / 44.4\\
     &V2B~\cite{v2b} &NeurIPS2021 &57.6 / 65.9 &23.7 / 37.9 &38.4 / 50.1 \\
     &STNet~\cite{stnet} &ECCV2022 &59.7 / 68.0 &25.5 / 39.9 &40.4 / 52.1 \\
     &\textcolor{black}{CXTrack}~\cite{xu2023cxtrack} &CVPR2023 &57.1 / 66.1  &30.7 / 49.4 & 42.2 / 56.7\\
     &\textcolor{black}{M3SOT}~\cite{liu2023m3sot} &None2023 &64.5 / 74.7 &32.8 / 52.1 &46.6 / 61.9 \\
     &\textcolor{black}{MoCUT}~\cite{cutrack} &ICLR2024 &61.9 / 69.7 &32.4 / 49.9 &-  \\
     \hline
    \end{tabular}}
\end{table}
    \textcolor{black}{With the advent of vision Transformer, numerous works have attempted to build LiDAR-based trackers based on the Transformer to harness the long-range relation modeling capacity of Transformers~\cite{ptt, pttr++, lttr, stnet, xu2023cxtrack, ma2023synchronize, cropnet}. Recently, we observed the trend of using a One-Stream Transformer (OST) architecture in LiDAR SOT following RGB VOT~\cite{ostrack2022, sbt2022, mixformer2022, mixformerv2, simtrack2022}. The OST streamlines the extraction and matching of template and search region features by employing joint relation modeling with Transformers. }
     \textcolor{black}{The pioneering work, SyncTrack~\cite{ma2023synchronize} synchronized feature extracting and matching networks using self-attention without siamese forward propagation, which simplified the model design in a single-branch and single-stage manner. The attending mechanism in the Transformer provides feature matching for the template and search region embeddings. }

    The results of LiDAR-based trackers are shown in Table~\ref{tab:lidar_kitti} of the main manuscript and Table~\ref{tab:lidar_nuscenes} and Table~\ref{tab:lidar_waymo} here. Note that the evaluation of nuScenes~\cite{nuscenes} dataset has two settings. The first one is training on the nuScenes training split and the categories include \textit{Car, Pedestrian, Truck, Trailer} and \textit{Bus}, which is the \textbf{nuScenes1} row in the Table~\ref{tab:lidar_nuscenes}. The other setting utilizes pretrained models on KITTI~\cite{KITTI2012} training dataset to evaluate the nuScenes testing set without model training on nuScenes training set. And the results are shown in the \textbf{nuScenes2} row in Table~\ref{tab:lidar_nuscenes}. Moreover, we report three settings of Waymo Open dataset results in Table~\ref{tab:lidar_waymo}. The three settings are from~\cite{m2track}, ~\cite{zhou2022pttr} and ~\cite{stnet}, respectively. Note that they use a different number of frames for training and testing, and S3 results are tested based on pretrained models on KITTI without training.
    
    Overall, the LiDAR modality can effectively provide the 3D geometry of the targets and is robust to viewpoint changes and environment changes like illumination, thus essential in the VOT applications that require 3D information. Despite the successful development of LiDAR-based VOT, challenges still exist and pose threats to successful tracking. The LiDAR points clouds have a deficiency of density variance, points sparsity, insufficient appearance information, and implicit features in data locality. Therefore, the LiDAR-based methods require relatively complex data pre-processing and are sensitive to the sampling quality of raw data. Combining with the RGB images can promisingly improve the model performance due to the complementary appearance information of the RGB modality. Besides, the single-branch and one-stream frameworks like the paradigm of RGB-based VOT (Sec.~\ref{sec:unifiedTrans}) will be a good candidate for LiDAR-based tracking, which simplifies the tracking paradigms a lot and has fewer model parameters and computational overheads. Moreover, with the development of large-scale 3D datasets like nuScenes~\cite{nuscenes}, the LiDAR-based pre-training will facilitate the tracking task with more effective visual representations. Instead of training from scratch, the trackers can be fine-tuned upon the models that are pretrained on large-scale LiDAR datasets. 
   

%% file: multi_modal.tex
	\section{Multi-modal Object Tracking}~\label{sec:multimodal}
	As shown in Fig.~\ref{figure:proscons}, we illustrate the characteristics and applications of four kinds of multi-modal VOT, including RGB-Depth, RGB-Thermal, RGB-LiDAR and RGB-Language. In this section, we elaborate their research progress in detail.
 \begin{table*}[h]
    \centering
    \caption{The performance of trackers with RGB-Depth dataset. We utilize \textit{Success}, \textit{AUC} and two kinds of \textit{F-Score} to evaluate models performance on PTB, STC and DepthTrack, respectively.}
    \label{tab:rgbd}
    \footnotesize
    \begin{tabular}{ccc|c|c|cc}
    \hline
        \multirow{2}*{Trackers} &\multirow{2}*{Publication}   &\multirow{2}*{Fusion}  &PTB &STC &\multicolumn{2}{c}{DepthTrack} \\
        & & &Success &AUC &Sequence-based F-Score &Frame-based F-Score\\ \hline
        AMCT~\cite{amct} &JDOS2012  &middle  &- &- &- &-\\
        PT~\cite{ptb} &ICCV2013  &early  &0.733 &0.35 \\
        MCBT~\cite{mcbt} &Neurocomputing2014  &early  &- &- &- &-\\
        DS-KCF~\cite{dskcf} &BMVC2015  &early &0.693 &0.34 &0.076 &0.075\\
        OL3DC~\cite{ol3dc} &Neurocomputing2015  &early &- &- &- &- \\
        DS-KCF-Shape~\cite{dskcf-shape} &J.RTIP2016  &early &0.719  &0.39 &0.071 &0.071 \\
        3D-T~\cite{3dt} &CVPR2016 & early &0.750 &- &- &-\\
        OAPF~\cite{oapf} &CVIU2016  &middle &0.731 &0.26 &- &- \\
        DLS~\cite{dls} &ICPR2016   &middle &0.740 &- &- &-\\
        STC~\cite{stc} &IEEE TCYB-2018  &late  &0.698 &0.40 &- &-\\
        CA3DMS~\cite{ca3dms} &TMM-2018 &early  &0.737 &0.43 &0.223 &0.216\\
        OTR~\cite{otr} &CVPR2019  &middle &0.769 &- &- &-\\
        DAL~\cite{dal} &ICPR2019 & middle &0.807 &0.64 &0.429 &0.433\\
        TSDM~\cite{tsdm} &ICPR2021 & middle  &0.792 &- &- &-\\
        DeT~\cite{depthtrack} &ICCV2021  &middle  &- &- &0.532 &0.500\\
        \textcolor{black}{ViPT}~\cite{zhu2023vipt} &CVPR2023 &middle &- &- &0.594 &0.592 \\
        \textcolor{black}{HMAD}~\cite{hmad} & ACMMM Asia2023 &middle &- &- &0.611 &0.597 \\
        \hline
    \end{tabular}
\end{table*}
	\begin{figure}[tbp]
		\begin{center}
			\includegraphics[width=1\columnwidth]{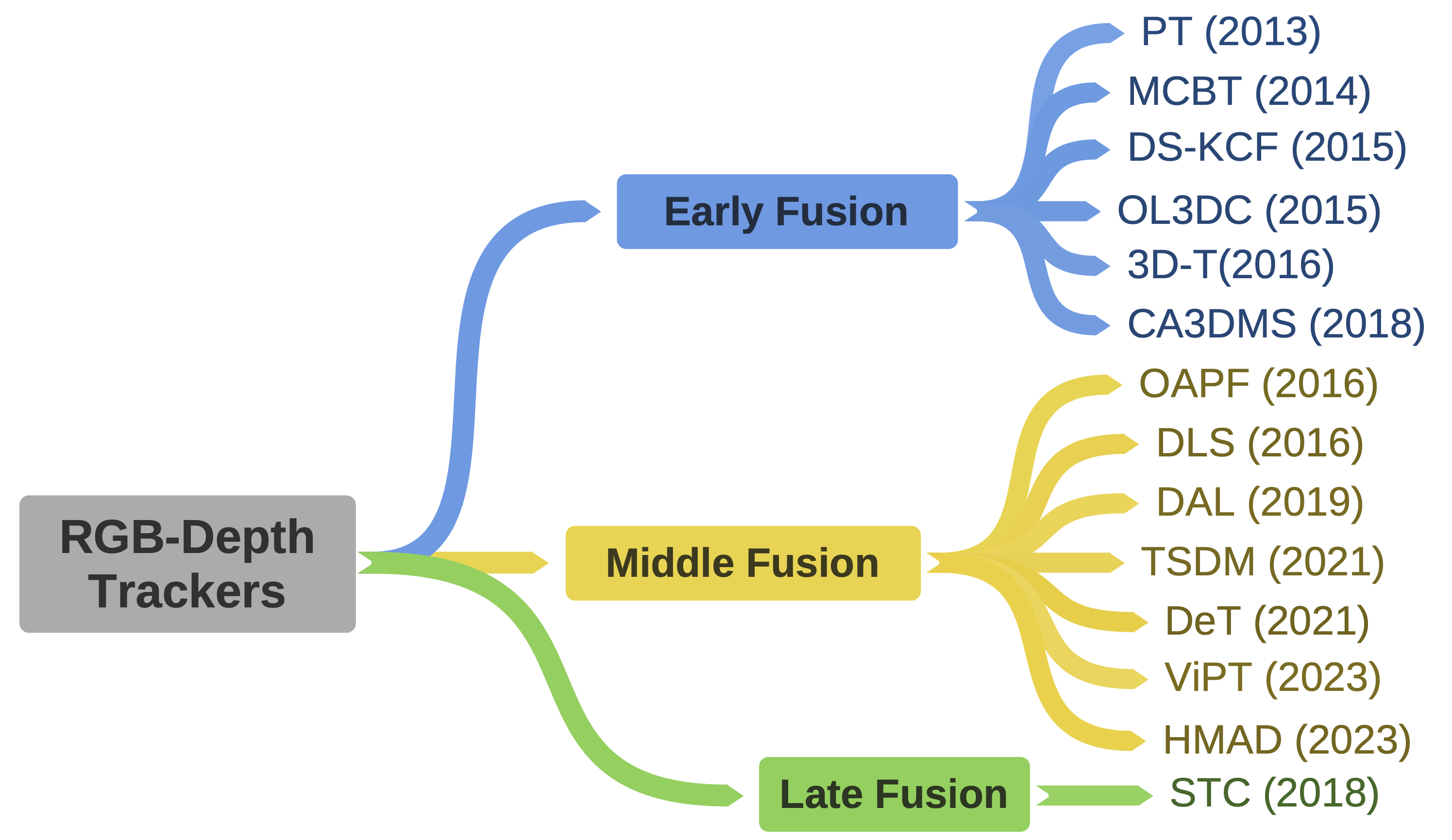}
		\end{center}
  \vspace{-15.5pt}
		\caption{\textcolor{black}{Develop lineage of RGB-Depth based trackers. We categorize the methods based on the feature fusion manners, which can be summarized into Early Fusion, Middle Fusion, and Late Fusion specifically.}}
  \vspace{-15.5pt}
		\label{figure:rgb-d}
	\end{figure}
\subsection{RGB-Depth Trackers}\label{sec:rgb+depth}
Depth information is significant in providing hints for reasoning the occlusion and sensing the distance and geometric information for better perceiving capability of the trackers. Thanks to the developments of the depth sensors such as Microsoft Kinect and Intel RealSense, much research has focused on improving RGB trackers with complementary depth information. Early RGB-Depth (RGB-D) VOT research always pays attention to conventional hand-crafted methods based on RGB trackers, extracting hand-crafted features from depth to fuse with RGB features for localizing. For example, AMCT~\cite{amct} leveraged the particle filter that is very popular in RGB tracking to predict. As one of the pioneering works, PT~\cite{ptb} established the benchmark dataset and metrics for evaluating the RGB-D trackers, as well as built various baseline algorithms like adding extra HOG features of depth maps into the 2D visual trackers.  Recently, deep learning methods have been introduced into RGB-D tracking to build a more accurate and robust model. Since 2019, the VOT-RGBD~\cite{vot} challenges have been proposed to stimulate the development of RGB-D tracking methods and create new benchmark datasets and metrics for the visual tracking community. We will introduce the mainstream RGB-D trackers from the perspective of RGB and depth features fusion strategies, and we divide the RGB-Depth trackers into three classes, namely early fusion, middle fusion and late fusion, which is also illustrated in Fig.~\ref{figure:rgb-d}.

Early fusion means fusing the RGB and depth modalities in the level of input data, concatenating the input RGB images and depth images for feature extraction. PT~\cite{ptb} and MCBT~\cite{mcbt} are classical methods of early fusion. Specifically, PT~\cite{ptb} combined the RGB and depth images for feature extraction and built the discriminative model with support vector machine (SVM)~\cite{svm}. 
MCBT~\cite{mcbt} proposed to combine multiple cues such as optical flow, color and depth information. This combination provided complementary information for object tracking, avoiding the tracking failure caused by insufficient knowledge of critical features. Moreover, DS-KCF~\cite{dskcf, dskcf-shape} extended the RGB tracker KCF~\cite{henriques2014high} by combining the RGB and depth inputs, and tackled the problems of scale changes and occlusions efficiently. OL3DC~\cite{ol3dc} directly learned a set of 3D context key points on combining RGB and depth images, building the spatial-temporal co-occurrence correlations with the template. 
It regards the synchronization as matching by generating the depth images of the next frame based on the optical flow between the current and the next frame. Moreover, CA3DMS~\cite{ca3dms} mapped each pixel of RGB image into 3D space according to depth images and leveraged the extended mean-shift tracker in 2D vision to address online adapting and occlusion.

Middle fusion refers to the paradigm that fuses the separately extracted features of different modalities and utilizes the fused features for tracking. 
For example, OAPF~\cite{oapf} mainly focused on the occlusion problem of RGB-D tracking and it fused multi-modal features based on the likelihood. The likelihood of a non-occluded particle was computed to represent the product of all features. DLS~\cite{dls} proposed decomposed the tracking task into detection and segmentation, and employed a simple 2D appearance model to merge the features. Moreover, many RGB-D fusion methods based on deep networks are proposed. DAL~\cite{dal} integrated the depth features into the state-of-the-art deep DCF architecture to model the target appearance. 
TSDM~\cite{tsdm} consisted of an RGB module and two depth modules to combine 2D appearance features with 3D geometric features. SiamRPN++~\cite{siamrpn++2019} module was leveraged in the RGB tracking part in TSDM. Moreover, DeT~\cite{depthtrack} designed an additional depth branch for depth features and was the first method to use a standalone depth network. Then, the depth and RGB features were directly concatenated and merged for tracking. \textcolor{black}{ViPT~\cite{zhu2023vipt} learned the prompts for different modalities, adapting the frozen pre-trained foundation models to diverse multi-modal tracking tasks. HMAD~\cite{hmad} proposed the attention-based shallow feature extraction and feature distribution fusion, fusing both the shallow features and deep features in the feature extraction network. } 
 
Late fusion means tracking the separated modalities independently, then fusing their results lately. We find a limited number of methods in late-fused RGB-D tracking. Precisely, STC~\cite{stc} fused the RGB and depth trackers via combining weighted estimations of final target positions, which was posterior to the feature extraction.
	
We present the performance of different RGB-D trackers on three mainstream benchmark datasets, including PTB~\cite{ptb}, STC~\cite{stc} and DepthTrack~\cite{depthtrack} in Table~\ref{tab:rgbd}. Currently, RGB-D methods always adopt CNN-based feature extractors and heuristic modality fusion strategies. With the success of Vision Transformers, building the backbone to aggregate multi-modal features efficiently will be a hot topic to discuss in the RGB-D tracking family as Transformer architectures show high performance in RGB tracking. Moreover, most fusion modules now are non-parametric, fusing the two modalities via weighted sum or joint distribution. Thus, a robust trainable fusion network will likely improve the tracking performance by performing deep interactions between the RGB and depth features. Lastly, most current RGB-D trackers are far from real-time tracking demands, which is vital to industrial deployments. Therefore, research on tracking efficiency is of practical value and significance.
\subsection{RGB-Thermal Trackers}
\begin{table*}[!ht]
    \centering
    \caption{The performance of different RGB-Thermal trackers. (PR: precision rate; SR: success rate; FPS: mean speed, taking the maximum speed that can be achieved in the corresponding paper)}
\label{tab:thermal+RGB}
    \footnotesize
    \begin{tabular}{c c c|c c|c c|c c|c}
    \hline
        \multirow{2}*{Trackers} & \multirow{2}*{Publication} & \multirow{2}*{Fusion} & \multicolumn{2}{c|}{GTOT} & \multicolumn{2}{c|}{RGBT234} & \multicolumn{2}{c|}{RGBT210}  & \multirow{2}*{FPS}   
        \\
        \cline{4-9}
~ & ~ & ~ & PR(\%) & SR(\%) & PR(\%) & SR(\%) & PR(\%) & SR(\%)  & ~ \\ \hline
SGT\cite{SGT} & ACM2017 & middle & 85.1 & 62.8 & - & - & 67.5 & - & 5   \\ 
SCCF\cite{SCCF} & PRCV2018 & late & 85 & 68.1 & - & - & - & -  & 50   \\ 
LGMG\cite{LGMG} & TCSVT2018  & middle & 82.9 & 65.5 & - & - & 71.1 & 46.8 & 7   \\ 
Cross-modal\cite{cross-modal} & ECCV2018 & middle & 82.7 & 64.3 & - & - & 69.4 & 46.3 & 8   \\ 
FANet\cite{FANet} & TIV2018 & middle & 88.5 & 69.8 & 76.4 & 53.2 & - & -  & 1.3   \\ 
FTSNet\cite{FTSNet} & NC 2018 & middle & 85.2 & 62.6 & - & - & - & -  & 15   \\ 
DenseFuse\cite{Densefuse} & TIP2018 & middle & - & 66.5 & 76.4 & 56.3 & - & -  & 1.5   \\ 
MDNet+RGBT\cite{MDNet+RGBT} & CVPR2018 & middle & - & - & 81.7 & 38.7 & - & - & -   \\ 
CSCF\cite{CSCF} & BICS2018 & late & 83.3 & - & - & - & 66.9 & 37.2  & 50   \\ 
TODA\cite{TODA} & ICIP2018 & middle & 84.3 & 67.7 & - & - & - & -  & 0.349   \\ 
Fast RGBT\cite{Fast} & NC 2019 & middle & 77 & 63.2 & - & - & 52.9 & -  & 227   \\ 
SiamFT\cite{SiamFT} & Access2019 & middle & - & - & 65.9 & 44.8 & 65 & 44.3 & 25+   \\ 
MANet\cite{MANet} & ICCV2019 & middle & 89.4 & 72.4 & 77.7 & 53.9 & - & -  & 1.1   \\ 
DAPNet\cite{DAPNet} & ACM2019 & middle & 88.2 & 70.7 & 76.6 & 53.7 & - & - & -    \\ 
CAT\cite{CAT} & ECCV2020 & middle & 88.9 & 71.7 & 80.4 & 56.1 & 79.2 & 53.3  & 20   \\ 
MaCNet\cite{MaCNet} & Sensors2020 & middle & 88.0  & 71.4 & 63.9 & 42.2 & - & -  & 1.4   \\ 
DMCNet\cite{DMCNet} & TNNLS2020 & middle & 90.9 & 73.3 & 83.9 & 59.3 & - & -  & 2.38   \\ 
CMPP\cite{CMPP} & CVPR2020 & middle & 92.6 & 73.8 & 82.3 & 57.5 & - & -  & 1.3   \\ 
MFGNet\cite{MFGNet} & TMM2021 & middle & 88.9 & 70.7 & 78.3 & 53.5 & 74.9 & 49.4  & 3.37   \\ 
DFNet\cite{DFNet} & None2021 & middle & 81.8 & 71.9 & 77.2 & 53.6 & - & -  & 28.6   \\ 
ADRNet\cite{ADRNet} & IJCV2021 & middle & 90.4 & 73.9 & 79.7 & 53.9 & - & - & 25   \\ 
SiamCDA\cite{SiamCDA} & TCSVT2021 & middle & 87.7 & 73.2 & 76.0  & 56.9 & - & - & 37   \\ 
SiamIVFN\cite{SiamIVFN} & None2021 & middle & 91.5 & 79.3 & 81.1 & 63.2 & - & -  & 147.6   \\ 
SiamFC+RGBT\cite{SiamFC+RGBT} & S\&C2021 & late & - & - & 61.0 & 42.8 & - & -  & -   \\ 
JMMAC\cite{JMMAC} & TIP2021 & late & 90.1 & 73.2 & 79.0  & 57.3 & - & -  & 26.6   \\ 
APFNet\cite{APFNet} & AAAI2022 & middle & 90.5 & 73.7 & 82.7 & 57.9 & - & - & - \\ 
\textcolor{black}{DMCNet}~\cite{lu2022duality} & TNNLS2022 & middle & 90.9 & 73.3 & 83.9&  59.3  & 79.7 & 55.5 & 2.38 \\ 
\textcolor{black}{Protrack}\cite{protrack} & ACM MM 2022 & middle & - & - & 78.6& 58.7  & - & - & -  \\ 
\textcolor{black}{ViPT}~\cite{zhu2023vipt} & CVPR2023 & middle & - & - & 83.5&  61.7  & - & - & 38.5  \\ 
\textcolor{black}{BAT}~\cite{cao2023bi} & None2023 & middle & - & - & 86.8&  64.1  & - & - & -  \\ 
\textcolor{black}{MPLT}~\cite{mplt}& None2023 & middle & - & - & 88.4&  65.7  & 86.2 & 63.0 & 22.8  \\ 
\hline
\end{tabular}
\end{table*}

   	 \begin{figure}[tbp]
		\begin{center}
			\includegraphics[width=1\columnwidth]{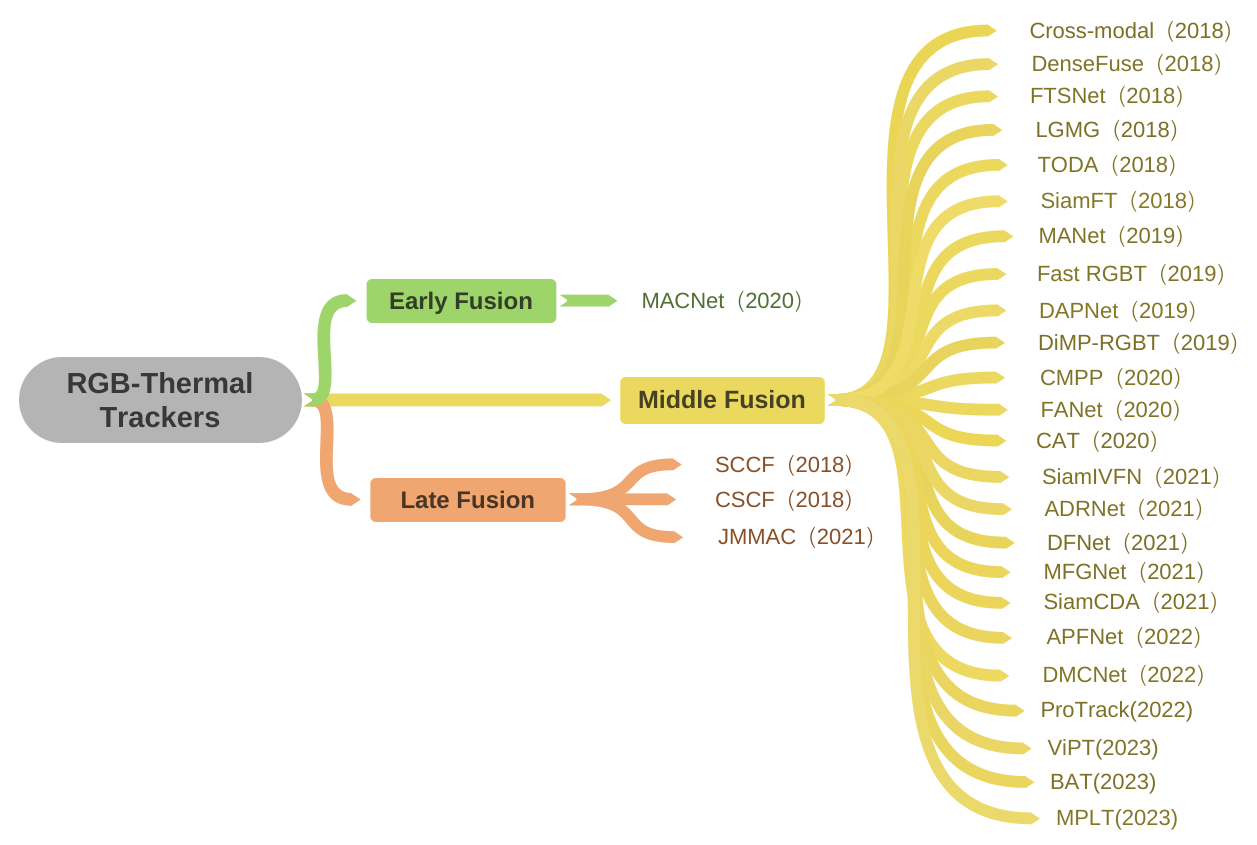}
		\end{center}
  \vspace{-10.5pt}
		\caption{\textcolor{black}{Develop lineage of RGB-Thermal based trackers. We categorize the methods based on the feature fusion manners, which can be summarized into Early Fusion, Middle Fusion, and Late Fusion specifically.}}
  \vspace{-5.5pt}
		\label{figure:thermal+rgb}
	\end{figure}

    RGB and thermal infrared (TIR) images have complementary properties, as shown in Fig.~\ref{figure:proscons}. The TIR images can cope with low illumination, while RGB images have rich appearance information. Hence it is intuitive to combine these two kinds of modalities for VOT, namely, RGB-Thermal (RGB-T) VOT, enabling all-weather and all-day working capabilities. There are many RGB-T methods in VOT literature. Similar to Section~\ref{sec:rgb+depth}, we divided these approaches from the perspective of modality fusion strategies, including early, middle and late fusion as shown in Fig.~\ref{figure:thermal+rgb}.
     
    Early Fusion means fusing the RGB and TIR images in the input level. Therefore, how to extract the features of the early-fused data is significant. For example, MACNet~\cite{MaCNet} adopted a two-stream network to extract features from RGB and thermal images. However, the drawback of these methods is that they require a high degree of alignment between two modalities, which mainly stem from the variations between heterogeneous data. Consequently, there is relatively limited researches related to the early fusion of RGB-T VOT. 
    
    Middle Fusion means that the features of RGB and TIR images are first extracted independently. Then the features are fused in the feature level through extra fusion modules. 
    The most simple combination approach is concatenating the features of the two modalities directly, and then learning a new feature after the interaction of the two modalities through a deep network. The RGB-T trackers that conformed to this paradigm, include SGT~\cite{SGT}, cross-modal~\cite{cross-modal}, LGMG~\cite{LGMG}, DenseFuse\cite{Densefuse},  FTSNet~\cite{FTSNet},  TODA~\cite{TODA}, MANet~\cite{MANet}, Fast RGBT~\cite{Fast}, DAPNet~\cite{DAPNet}, DiMP-RGBT~\cite{DiMP-RGBT}, CMPP~\cite{CMPP}, FANet~\cite{FANet}, CAT~\cite{CAT}, ADRNet~\cite{ADRNet},  DFNet~\cite{DFNet},   MFGNet~\cite{MFGNet}, APFNet~\cite{APFNet}, DMCNet~\cite{DMCNet} and so on. For instance, FTSNet used a two-stream CNN for the feature extraction of these two modalities and designed a FusionNet to fuse the two kinds of features. Based on DiMP\cite{DiMP}, DiMP-RGBT investigated an end-to-end training framework, which explored all three types of fusion strategies (early, middle and late) and verified that the middle fusion achieved the best results. CMPP built an inter-modal pattern-propagation process via affinity correlation to propagate the patterns of one modality to the other. FANet proposed a hierarchical and adaptive aggregation framework to represent cross-modal collaboration information better. SGT jointly fused each modality weight by the sparse representation regularized graph. MFGNet combined the feature maps extracted from thermal and RGB images by feeding them to a dynamic modality-aware filter generation network. APFNet introduced five attribute-specific fusion branches to progressively fuse thermal and RGB features and designed an enhancement fusion transformer to improve modality integration. Moreover, several middle fusion algorithms developed upon Siamese structure, including SiamFT~\cite{SiamFT}, SiamIVFN~\cite{SiamIVFN} and SiamCDA~\cite{SiamCDA}. 
     Specifically, SiamFT was the first to adopt the basic Siamese architecture into RGB-Thermal tracking, which directly used two Siamese networks for two modalities and then concatenated their template and search features, respectively, to do the correlation operation. After that, SiamIVFN used a complementary-feature-fusion network and a contribution-aggregation network to extract the common features and fuse the two modalities. SiamCDA utilized dual-path SiamRPN for unimodal feature extraction and designed a complementarity-aware multi-modal feature fusion module to capture the cross-modal information. In summary, the middle fusion is the mainstream for RGB-Thermal VOT, since it provides learnable parameters to combine the semantic features of the two modalities. \textcolor{black}{DMCNet~\cite{lu2022duality} introduced a duality-gated mutual condition network to exploit the discriminative information of all modalities while suppressing the effects of data noise. Recent advancements in Parameter-Efficient Fine-Tuning (PEFT)~\cite{cocoop,wang2021actionclip,maple,m2clip} have inspired numerous methods to transfer RGB-modal trackers into multimodal tracking~\cite{hou2024sdstrack,cao2023bi,zhu2023vipt}. For instance, Cao et al. introduced BAT~\cite{cao2023bi}, which transferred pre-trained trackers to multi-modal scenarios by learning only a universal bi-directional adapter. Protrack\cite{protrack} presented a prompt learning paradigm to utilize both the large-scale RGB knowledge and the complementary information. ViPT~\cite{zhu2023vipt} proposed a modality-complementary prompter to generate valid visual prompts for the task-oriented multi-modal tracking. MPLT~\cite{mplt} designed a multi-modal visual information prompter to adaptively generate weights from another modality. These PEFT-based methods are very intriguing, as they enable the utilization of pure RGB-modal models with fewer parameters, thus partially addressing the issue of limited multi-modal data availability.}

     Late Fusion means performing fusion upon tracking results of two modalities.
     Methods based on this strategy include SCCF~\cite{SCCF}, CSCF~\cite{CSCF} and JMMAC~\cite{JMMAC}. For example, CSCF generated the final tracking result by fusing the two modalities' response maps. JMMAC used ECO as the appearance model to generate the response maps of the target for the two modalities, respectively. Then it designed a multi-modal feature network to weigh the response maps. Compared with early and middle fusion, late fusion requires fewer computational overheads, whereas having faster tracking speed and fewer limitations. 
     
   We present the performance, fusion strategies and inference speed of the typical RGB-thermal trackers in Table~\ref{tab:thermal+RGB} on four well-known RGB-thermal VOT benchmarks. In conclusion, RGB and TIR modalities have complementary properties. Thus the RGB-T trackers are expected to realize the all-weather complex tracking task, which is significant for various real-world scenarios. The feature representation of both the RGB and TIR, and their interactions and fusions are the key to RGB-T tracking. In the future, stronger feature extractors and cross-modal relation modeling should be explored in depth, such as introducing the Transformer.

\subsection{RGB-LiDAR Trackers}
     \begin{table*}[h]
    \centering
        \caption{The performance of trackers on RGB-LiDAR dataset. The results are evaluated by \textit{Success}/\textit{Precision}.}
    \label{tab:rgb+lidar}
    \footnotesize
    \begin{tabular}{ccc|ccccc}
    \hline
        \multirow{2}*{Trackers} &\multirow{2}*{Publication}   &\multirow{2}*{Fusion}  &\multicolumn{5}{c}{KITTI} \\
        & & &Car &Pedestrian &Van &Cyclist &Mean\\ \hline
        Alireza \textit{et al.}~\cite{alireza} &ITSC2016  &late  &- &- &- &- &-\\
        F-Siamese Tracker~\cite{F-siamese} &IROS2020  &middle  &37.15/50.60 & 16.28/32.28 &- &47.03/77.26 &-\\
        Wang \textit{et al.}~\cite{wang2021facilitating} &PRCV2021 &middle &- &- &- &- &60.8/77.0\\
        \hline
    \end{tabular}
\end{table*}
\textcolor{black}{Research on RGB-LiDAR-based VOT has surfaced in recent years and warrants comprehensive exploration, given the limited number of published studies in this area.} RGB-LiDAR based tracking methods fuse the RGB images with LiDAR point clouds to complement additional information for better performance. As is known, point clouds are always sparse and diverse in density, resulting in substantial computational complexity of generating proposals in the process of tracking-by-detecting. However, RGB information can make up the semantic scarce and improve modeling efficiency. Therefore, the essential problem of RGB-LiDAR tracking is how to fuse the two modalities efficiently and effectively. Alireza \textit{et al.} ~\cite{alireza} utilized two parallel mean-shift localizers to calculate the 2D and 3D locations, respectively, and then the 2D and 3D modalities were fused by a Kalman filter. F-Siamese Tracker~\cite{F-siamese} generated 2D ROI proposals via a Siamese network based on RGB images and then combined the region of interest with point clouds, feeding into another Siamese network for outputting 3D bounding boxes. Moreover, Wang \textit{et al.}~\cite{wang2021facilitating} introduced geometric features of 2D images into 3D point tracking. A double judgment fusion module was proposed for modalities fusion, judging the projected image quality and correlation relations to point clouds adaptively. 
    \begin{table*}[h]
    \centering
     \caption{The performance of trackers with RGB-Language dataset. (NL: results of tracking by natural language only; NL+BBox: results of tracking by joint language and BBox), the results are evaluated by \textit{Precision}/\textit{AUC}.}
    \label{tab:rgb+language}
    \footnotesize
    \begin{tabular}{ccc|cc|cc|cc}
    \hline
        \multirow{2}*{Trackers} &\multirow{2}*{Publication}   &\multirow{2}*{Fusion}  &\multicolumn{2}{c|}{OTB-Lang} &\multicolumn{2}{c|}{LaSOT} &\multicolumn{2}{c}{TNL2K} \\
        & & &NL &NL+BBox &NL &NL+BBox &NL &NL+BBox\\ \hline
        Li \textit{et al.}~\cite{lan-li} &CVPR2017  &middle  &0.29/0.25 &0.72/0.55 &- &- &- &-\\
        Wang \textit{et al.}~\cite{lan-wang1} &None2018  &middle  &- &0.89/0.66 &- &0.30/0.27 &- &-\\
        Feng \textit{et al.}~\cite{lan-feng1} &None2019  &middle  &0.56/0.54 &0.73/0.67 &- &0.56/0.50 &- &0.27/0.34/0.25 \\
        Feng \textit{et al.}~\cite{lan-feng2} &WACV2020  &middle  &0.78/0.54 &0.79/0.61 &0.28/0.28 &0.35/0.35 &- &0.27/0.33/0.25\\
        GTI~\cite{gti} &TCSVT2020  &middle  &- &0.73/0.58 &- &0.47/0.47 &- &- \\
        Wang \textit{et al.}~\cite{lan-wang2} &CVPR2021  &middle  &0.24/0.19 &0.88/0.68 &0.49/0.51 &0.55/0.51 &0.06/0.11/0.11 &0.42/0.50/0.42\\
        SNLT~\cite{snlt} &CVPR2021 &middle  &- &0.85/0.67 &- &0.54/0.57 &- &- \\
        Li \textit{et al.}~\cite{li2022cross} &CVPRW2022 &middle &0.72/0.53  &0.91/0.69 &0.51/0.52 &0.56/0.53 &0.09/0.15/0.14 &0.45/0.52/0.44 \\
        \textcolor{black}{$VLT_{TT}$~\cite{guo2022divert}} &NeurIPS2022 &middle  &- &0.90/0.74 &- &0.72/0.67 &- &0.53/0.53 \\
        \textcolor{black}{All-in-One~\cite{all_in_one}} &ACMMM2023 &middle &- &0.93/0.71 &- &0.79/0.72 &- &0.57/0.55 \\
        \textcolor{black}{OVLM~\cite{ovlm}} &TMM2023 &middle &- &- &- &0.71/0.66 &- &0.67/0.63 \\
        \textcolor{black}{Zhou \textit{et al.}~\cite{jointgroundtrack}} &CVPR2023 &middle &0.78/0.59 &0.86/0.65  &0.59/0.57 &0.64/0.60  &0.55/0.55 &0.58/0.57 \\
        \textcolor{black}{CiteTracker~\cite{li2023citetracker}} &ICCV2023 &middle &- &- &- &0.76/0.70 &- &0.60/0.58\\
        \textcolor{black}{MMTrack~\cite{mmtrack}} &TCSVT2023 &middle &- &0.92/0.71 &- &0.76/0.70 &- &0.59/0.59 \\
        \textcolor{black}{LMTrack~\cite{lu2023naturalmixer}} &ISPDS2023 &middle &- &0.93/0.75 &- &0.78/0.72 &- &0.59/0.57 \\
        \textcolor{black}{UVLTrack}~\cite{ma2024unifying} &None2024 &middle &0.83/0.64 &0.92/0.71 &0.64/0.60 &0.79/0.71 &0.61/0.58 &0.69/0.65 \\
        \textcolor{black}{SATracker~\cite{ge2023beyond}} &None2024 &middle & - &0.94/0.74 &- &0.80/0.72 &- &0.64/0.61\\
        \hline
    \end{tabular}
\end{table*}
	
The performances of RGB-LiDAR trackers are reported in Table~\ref{tab:rgb+lidar}. The number of trackers and datasets is limited as related research is still in its infancy. This new research topic emerged based on the development of LiDAR tracking, combining the geometric and appearance features of LiDAR and RGB, respectively. The research on RGB-LiDAR VOT will follow the research trend of LiDAR-based tracking, integrating the RGB features into the current LiDAR-based frameworks. Moreover, the research on fusion manners of RGB-LiDAR is essential, as discrepancy exists in the 2D and 3D data format. Gaping the discrepancy stimulates the correspondence of the 2D plane and 3D world, empowering the tracking model with more substantial perceptive capability.

	\subsection{RGB-Language Trackers}

    Linguistic information can serve as good hints for visual tasks with the development of vision-language co-training~\cite{clip, align}. Language modality is also integrated with the visual tracking field to supplement extra linguistic information for RGB images. The task can be formulated as given a video frame and a language expression as a query, the goal is to track the target according to the language descriptions. Compared with original RGB tracking that relies merely on bounding boxes, it can more accurately express both the spatial location and semantic information as discussed in~\cite{lan-wang1}. As the pioneering work, Li \textit{et al}.~\cite{lan-li} first introduced linguistic descriptions to visual tracking, extending two tracking datasets with linguistic sentences. Feng \textit{et al.}~\cite{lan-feng2} employed RNN layers to locate the target conditioned on a linguistic network. Moreover, some methods borrow ideas from the visual grounding task, such as~\cite{lan-wang2, gti}. Wang \textit{et al.}~\cite{lan-wang2} leveraged the pretrained BERT~\cite{devlin2018bert} to perform visual grounding to help the visual tracking module. The large pretrained NLP models such as BERT~\cite{devlin2018bert} are beneficial to capturing semantic information from vision. GTI~\cite{gti} also adopted the paradigm of \textit{Grounding-Tracking}, then it integrated grounding and tracking based on the region correctness score and template quality score. 
	SNLT~\cite{snlt} paid more attention to dynamic fusion of RGB and language by introducing a dynamic aggregation module. 
    \textcolor{black}{All-in-One~\cite{all_in_one} proposed a multi-modal alignment module and a unified Transformer network, where the language tokens are projected to the space of vision tokens and concatenated together to feed into the Transformer for feature extraction and interaction. Similar to All-in-One, OVLM~\cite{ovlm} utilized a one-stream Transformer to jointly tackle vision-language tokens, but it proposed a memory mechanism and built a more compact target model. 
    Zhou \textit{et al.}~\cite{jointgroundtrack} aimed to integrate the grounding into the tracking framework, localizing the referred target based on the given visual-language references. CiteTracker~\cite{li2023citetracker} designed a text generation module that generates a text description of the target object and then estimates the target state based on the text features. MMTrack~\cite{mmtrack} took the spatial coordinates of the target and language tokens as input, iteratively generating target sequences in an auto-regressive manner. VLT$_{TT}$~\cite{modamixer} and UVLTrack~\cite{ma2024unifying} all proposed to utilize contrastive learning strategy to minimize the distances between positive vision-language token pairs. LMTrack~\cite{lu2023naturalmixer} introduced a language-guided attention mixer, fusing the vision and language features by the mixer, which comprises cross-attention and FFN. SATracker~\cite{ge2023beyond} synchronized the vision and language feature extracting,  proposing the Target Enhance Module (TEM) and the Semantic Aware Module (SAM) modules for efficient semantic understanding. }

	As shown in Table~\ref{tab:rgb+language}, results on three benchmark datasets are reported to compare the RGB-Language based trackers. The performance based on language only and language with bounding boxes is evaluated for more comprehensive evaluations. In summary, the language modality is more understandable to humans and can be utilized to provide more accurate spatial locations and semantic information. With the advent of vision-language foundation models like CLIP~\cite{clip}, RGB-Language tracking can be boosted by leveraging the knowledge of vision-language pretraining. Therefore, instead of using pretrained NLP model like Wang \textit{et al.}~\cite{lan-wang2} does, the vision-language model can be considered to utilize to minimize the gap between spatial images and semantic textual sentences via contrastive learning. Moreover, with the help of linguistic hints, open-world tracking tasks can be explored to build more general and robust trackers to track objects out of the domain.

%% file: datasets.tex
	\section{Datasets}\label{sec:datasets}
	Datasets are fundamental to VOT research, especially to DNN-based methods. In this section, we introduce the widely-used VOT benchmarks for different modalities. Table~\ref{tab:benchmarks} shows some critical information of datasets, including the proposed year, the number of videos and labeled boxes, object classes, and average duration. Table~\ref{tab:RGB2024},~\ref{tab:RGB2021},~\ref{tab:RGBbefore2018},~\ref{tab:thermal},~\ref{tab:lidar_kitti},~\ref{tab:lidar_nuscenes} and~\ref{tab:lidar_waymo} list the comparison results of single-modality tracking algorithms, including RGB-based, TIR-based and LiDAR-based. Table~\ref{tab:rgbd},~\ref{tab:thermal+RGB}, ~\ref{tab:rgb+lidar} and ~\ref{tab:rgb+language} list the comparison results of multiple modalities, including RGB-Depth, RGB-TIR, RGB-LiDAR and RGB-Language.
  \begin{table}[h]
    \centering
     \caption{Representative benchmarks of VOT with various data modalities (AD: Average Durations, RGB-La: RGB-Language.}
    \label{tab:benchmarks}
    \footnotesize
    \scalebox{0.92}{\begin{tabular}{c|cccccc}
    \hline
        \multicolumn{2}{c}{Dataset} &Year  &Videos &Boxes &Class &AD \\ \hline 
        \multirow{18}{*}{\rotatebox{90}{RGB}}&OTB100~\cite{otb100} & 2015  & 100 & 58.61k &22 & 19.68s \\
        &TC128~\cite{tc128} & 2015  & 128  & 55.65k   &27  & 15.6s \\
        &UAV123~\cite{uav123} &2016  &123  & 113.48k  &9  & 30.48s\\
        &UAV20L~\cite{uav123} &2016  &20   &58.67k   &5   & 97.8s \\
        &NFS~\cite{nfs} &2017  & 100 & 383k &31 & 15.96s\\
        &VOT2013~\cite{vot2013} &2013  & 16  &5.68k & 7 & 11.83s\\
        &VOT2014~\cite{vot2014} &2014 &25   &10.21k    &10  & 13.62s\\
        &VOT2015~\cite{vot2015} &2015  &60   &21.46k &21  &11.92s \\
        &VOT2016~\cite{vot2016} &2016  &60   &21.46k &21  & 11.92s \\
        &VOT2017~\cite{vot2017} &2017  &60 & 21.36k &26  &11.86s \\
        &VOT2018~\cite{vot2018} &2018 &60   &21.36k &26  & 11.86s \\
        &VOT2019~\cite{vot2019} &2019  & 60  &19.94k   &27 &11.08s \\
        &VOT2020~\cite{vot2020} &2020 & 60  &19.94k    &27  &11.08s \\
        &VOT2021~\cite{vot2021} &2021  &60   &19.45k    &28  &10.80s \\
        &VOT2022~\cite{vot2022} &2022  &62   & 19.90k   &29  &10.70s \\
        &TrackingNet~\cite{trackingnet} &2018  &30.64k &14M &21 &16.7s \\
        &LaSOT~\cite{lasot}     &2019  &1.4k &3.52M  &70   &83.57s \\
        &GOT-10k~\cite{got10k}  &2019  &10k  &1.5M   &0.5k &15s \\ \hline
        \multirow{7}{*}{\rotatebox{90}{Thermal}}
        &LTIR~\cite{DSST-TIR} & 2015 & 20 & 6.2k & 6 & 10.3s\\
        &VOT-TIR~\cite{VOT-TIR2016} & 2016 & 25 & 13.8k & 6 & 18.4s\\
        &PTB-TIR~\cite{PTB-TIR} & 2019 & 60 & 30.1k & 1 & 16.7s\\
        &MLSSNet~\cite{MLSSNet} & 2020 & 430 & 200k & 20 & 15.5s\\
        &MMNet~\cite{MMNet} & 2020 & 1.1k & 530k & 30 & 16.0s\\
        &LSOTB-TIR~\cite{LSOTB-TIR} & 2020 & 1.4k & 730k & 12 & 17.3s\\
        &ECO-MM~\cite{ECO-MM} & 2022 & 1.1k & 450k & 30 & 13.6s\\\hline
        \multirow{3}{*}{\rotatebox{90}{LiDAR}}
        &KITTI~\cite{KITTI2012} & 2012 & 21 & - & 8 & -\\
        &nuScenes~\cite{nuscenes} & 2019 &1k  & 1.4M & 23 & 20s\\
        &Waymo Open~\cite{waymo} & 2020 & 1.95k & 12.6M & 4 & 20s\\ \hline
        \multirow{4}{*}{\rotatebox{90}{RGB-D}}
        &PTB~\cite{ptb} & 2013 & 100 & 20.33k & 3 & 6.7s\\
        &STC~\cite{stc} & 2018 &36  & 9.19k & 18 & 10s\\
        &CDTB~\cite{cdtb} & 2019 & 80 & 101.95k & 21 & 42.1s\\ 
        &DepthTrack~\cite{depthtrack} & 2021 & 200 & 218.20k & 46 & 48.0s\\\hline
        \multirow{8}{*}{\rotatebox{90}{RGB-Thermal}}
        &GTOT~\cite{GTOT} & 2016 & 50 & 15.8k & 4 & 10.5s\\
        &RGBT210~\cite{RGBT210} & 2017 & 210 & 210k & 12 & 33.3s \\
        &RGBT234~\cite{RGBT234} & 2019 & 234 & 233.8k & 12 & 33.3s \\
        &VOT-RGBT2019~\cite{VOT-RGBT2019}& 2019 & 60 & 40.2k & - & 22.3s\\
        &VOT-RGBT2020~\cite{VOT-RGBT2020}& 2020 & 60 & 20.0k & - & 33.5s\\
        &LSS~\cite{SiamCDA} & 2021 & 4.8k & 12k & 40 & 2.48s\\
        &LasHeR~\cite{LasHeR}& 2021 & 1.2k & 734.8k & 32 & 20s\\
        &VTUAV~\cite{HMFT} & 2022 & 500 & 2.7M & 18 & 18s \\ \hline
        \multirow{3}{*}{\rotatebox{90}{RGB-La}}
        &OTB-Lang~\cite{li2017tracking} & 2017 & 99 & 58k & 22 & 19.68s\\
        &LaSOT~\cite{lasot} & 2018 &1.4k  & 3.52M &70  & 85.57s\\
        &TNL2K~\cite{tnl2k} & 2021 & 2k & 1.24M &-  & 18.6s\\
        \hline
       
    \end{tabular}}
\end{table}

 \subsection{RGB-based dataset}\label{sec:rgbdataset}
\textbf{OTB.} OTB is a popular VOT benchmark, especially before the deep learning era. It has 100 fully annotated sequences, where the number of frames in each sequence ranges from hundreds to thousands. 
OTB has three subsets, OTB2013~\cite{otb50}, OTB50~\cite{otb50} and OTB100~\cite{otb100} (OTB2015). OTB2013 contains 51 video sequences and OTB50 contains 50 video sequences selected from OTB100, while OTB100 contains all those 100 sequences. OTB has 11 annotated video attributes like illumination variation, scale variation, etc. OTB uses one pass evaluation (OPE) to evaluate trackers with two metrics, precision and area under the curve (AUC) of the success plot.\\
\textbf{TC128.} TC128~\cite{tc128} is designed for the color impact evaluation of VOT, which consists of 128 fully labeled sequences of color images of 27 object categories. Among the 128 videos, 50 sequences are the same as OTB. Besides, TC128 also has the same 11 attributes and evaluation protocols as OTB.\\
\textbf{UAV123.} UAV123~\cite{uav123} contains sequences captured from low-altitude UAVs. Unlike other tracking datasets, the viewpoint of UAV123 is aerial and the targets to be tracked are usually small. This dataset is further divided into two subsets, UAV123 and UAV20L. UAV123 includes 123 short videos  of 9 diverse object categories. UAV20L consists of 20 long videos of 5 object classes generated from a flight simulator. This dataset is evaluated with the same protocols as OTB. \\  
\textbf{NFS.} NFS~\cite{nfs} is the first high frame rate benchmark for VOT. The dataset comprises 100 videos (380K frames) captured with now commonly available higher frame rate (240 FPS) cameras from real-world scenarios. All frames are annotated with axis-aligned bounding boxes and all sequences are manually labeled with nine visual attributes. There are two versions of NFS, namely, NFS240 and NFS30, corresponding to 240 FPS capture, and 30 FPS with synthesized motion blur (generated with Adobe After Effects).
\\
\textbf{VOT.} VOT dataset series~\cite{vot2013,vot2014,vot2015,vot2016,vot2017,vot2018,vot2019,vot2020,vot2021,vot2022} begin from 2013 and is updated every year, which organizes the annual VOT challenge competition to benchmark tracking performance. The VOT series are labeled with rotated bounding boxes with various tracking challenges. In the beginning, VOT2013 is used to evaluate short-term VOT performance. Evolution to the present, the VOT2022 could be used to address short-term, long-term, real-time, RGB and RGB-D trackers. The sequences vary every year. The VOT challenges have their own evaluation metrics, namely Accuracy (A), Robustness(R), and Expected  Average Overlap (EAO) metrics. 
\\
\textbf{TrackingNet.} TrackingNet~\cite{trackingnet} is a large-scale dataset consisting of real-world videos sampled from YouTube. It contains more than 30000 sequences with 14 million dense annotations. The test set consists of 511 sequences and covers diverse object classes and scenes, requiring trackers to have both discriminative and generative capabilities. The trackers are evaluated using an online evaluation server on the test set. Similar to OTB, TrackingNet adopts OPE to evaluate success and precision, with another metric Normalized Precision Plot. \\
\textbf{LaSOT.} LaSOT~\cite{lasot} provides large-scale, high-quality dense annotations with 1,120 videos in the train set and 280 videos in the test set. It has 70 categories of objects containing twenty sequences, and the average video length is 2512 frames, sufficient to evaluate long-term trackers. Besides, LaSOT also has the same evaluation protocols as TrackingNet.
\\
\textbf{GOT-10k.} GOT-10k~\cite{got10k} is a large highly-diverse dataset that contains over 10000 video segments from a semantic hierarchy of WordNet~\cite{wordnet} with more than 1.5 million manually labeled bounding
boxes. It has 9340 sequences with 480 object categories in the training split, and 420 videos with 83 object categories in the test split. Each sequence has an average length of 127  frames. The train and test splits have no overlap in object classes, thus overfitting on particular classes is avoided. In addition, this benchmark requires all trackers to use only the train split for model training, while external datasets are forbidden. Average Overlap (AO), $SR_{0.50}$ and $SR_{0.75}$ are used to evaluate the trackers.

\subsection{Thermal-based dataset}
\textbf{LTIR.}
 LTIR\cite{DSST-TIR} is a short-term single-object tracking dataset, which provides 20 thermal sequences with 6 object classes. The average sequence length is 563 frames. As LTIR contains varying challenging events of different objects,  an evaluation toolkit for tracking methods is performed for visual sequences on both visual and thermal data.\\
\textbf{VOT-TIR.}
 VOT-TIR 2015\cite{VOT-2015} is the first standardized and commonly-used benchmark for TIR sequences, which adapts VOT2013\cite{vot2013} to thermal format. One year later, another dataset called VOT-TIR 2016\cite{VOT-TIR2016} extends from VOT-TIR2015 and becomes more challenging. Specifically, it contains 25 grayscale TIR sequences with eight object classes that can be used to evaluate a tracker on specific attributes.\\
\textbf{PTB-TIR.}
PTB-TIR\cite{PTB-TIR} is a challenging pedestrian tracking dataset of thermal images, containing 60 sequences with manual annotations and nine attributes. The total number of frames is 30128, and the data resolution is the highest till the time of creation, which can reach $1280 \times 720$. Regrettably, the application scenarios are limited due to the lack of division between different scenario attributes.\\
\textbf{MLSSNet.}
MLSSNet~\cite{MLSSNet} is a large-scale TIR dataset with 500 video sequences and 228k frames in total. Different from other datasets, its max resolution can reach $1920 \times 1080$. Besides, MLSSNet-data has great diversity in various scenarios and most of them are shotted at night.\\
\textbf{MMNet.}
MMNet~\cite{MMNet} provides over 1100 sequences with 30 classes for TIR-based VOT. There are 450k video frames and 530k bounding boxes captured from different shooting devices, including hand-held, vehicle-mounted, surveillance, and drone-mounted. The TIR images are stored with a white-hot style and an 8 bits depth.\\
\textbf{LSOTB-TIR.}
 LSOTB-TIR\cite{LSOTB-TIR} is a large-scale and high-diversity TIR-based VOT dataset. It has 1400 TIR sequences, over 600k frames, and 730k bounding boxes with high-quality annotations for training datasets. For evaluation, it defines several scenario attributes, including thermal crossover, distractors, intensity variation, etc., indicating various challenges. \\
 \textbf{ECO-MM.}
ECO-MM~\cite{ECO-MM} consists of 1100 TIR sequences with 450k frames and 30 classes. It is captured from more than 30 shooting devices and 4 kinds of view angles. The TIR videos have a wide range of object classes, shotting devices, shotting scenes, and shotting view angles, which ensure the diversity of the dataset.

\subsection{LiDAR-based dataset}\label{sec:lidardataset}
\textbf{KITTI.}
KITTI~\cite{KITTI2012} is one of the most popular datasets used in mobile robotics and autonomous driving. As for the tracking, the benchmark consists of 21 training sequences. The LiDAR-base trackers always split the training sequences into train/val/test, with scenes 0-16 for training, scenes 7-18 for validation and scenes 19-20 for testing respectively. The dataset has four categories, namely Car, Pedestrian, Van and Bicycle, while the data samples of different classes are imbalanced. \\
\textbf{nuScenes.}
The nuScenes~\cite{nuscenes} dataset is a large-scale public dataset for autonomous driving developed by the team at Motional (formerly nuTonomy). The dataset contains 1000 driving scenes collected from Boston and Singapore, and the scenes of 20-second length are manually selected to show a diverse and interesting set of driving maneuvers, traffic situations and unexpected behaviors. In the configurations of LiDAR-based VOT, the train/val/test sets make up 700/150/150 of the whole 1000 scenes respectively. Though nuScenes samples at 20Hz, the annotations of sampled data are provided at a frequency of 2 Hz. This introduces huge difficulties for the tracking tasks as such a lower frequency for keyframes results in large information inconsistency.\\
\textbf{Waymo Open.}
The Waymo Open dataset~\cite{waymo} is a perception and motion planning video dataset for self-driving cars that comprises high-resolution sensor data. It includes 1150 scenes, with 798 for training, 202 for validation, and 150 for testing. The experiments conducted by the tracking methods always leverage training and validation sets to perform training and testing, respectively. The Waymo Open dataset is much more challenging than the KITTI dataset, with greater data volume and complexity.

\subsection{RGB-Depth dataset}
\textbf{PTB.}
PTB~\cite{ptb} is proposed as the first well-designed dataset for RGBD object tracking. 
It contains 100 video clips with RGB and depth modality captured by a Microsoft Kinect 1.0. The videos are collected indoors, varying the depth values from 0.5 to 1.0 meters. All the video frames are manually annotated by one person for high consistency, and 95 of the 100 videos are not publicly accessible, only allowing online submissions for evaluation. 
The remaining five videos are public for model validation before online submission. The overall tracking performance is measured by the success rate, i.e., the percentage of frames where the overlap between ground truth and the predicted bounding box exceeds 0.5.\\
\textbf{STC.}
STC~\cite{stc} is first proposed in 2018 to complement the PTB dataset in terms of the visual attribute.  
It is recorded by two Asus Xtion RGBD sensors and contains mainly indoor sequences and a small number of low-light outside sequences. The STC is a small dataset that contains only 36 sequences and around 9k frames. However, it includes manual annotations of thirteen attributes, including illumination variation, depth variation, scale variation, color
distribution variation and so on.\\
\textbf{CDTB.}
CDTB~\cite{cdtb} is one of the most challenging datasets in RGB-D families.  
The sequences are captured in the long-term tracking scenario with a Kinect v2, outputting 24-bit $1920\times 1080$ images and 32-bit $512\times 424$ depth images. All sequences are manually annotated per-frame with 13 attributes. 
The CDTB contains 80 video clips and around 102k frames total, and the occlusion and disappearance of target objects are common in the frames, making it quite a challenging benchmark for RGB-D tracking. \\
\textbf{DepthTrack.}
The DepthTrack dataset~\cite{depthtrack} is the most recent and the largest, most diverse of the available datasets in the area of RGB-D tracking. The data collection of DepthTrack is based on a single Intel Realsense 415 sensor and is particularly focused on content diversity. The DepthTrack consists of 200 video sequences and around 218k frames overall, with the most diverse categories and scenarios, up to 46 classes and 15 attributes of both indoors and outdoors. 
Moreover, DepthTrack is substantially larger than any previous benchmarks. 

\subsection{RGB-Thermal dataset}
\textbf{GTOT.}
GTOT\cite{GTOT} contains 50 grayscale-thermal sequences and over 15.8k frames with detailed annotation data, including bounding boxes and challenging attributes. The video pairs are recorded in sixteen scenes, including laboratory rooms, campus roads, playgrounds and water
pools, etc. To guarantee consistency, all annotations are done by a full-time annotator.\\
\textbf{RGBT 210 and RGBT234.}
The two benchmarks are captured from a low-resolution CCD camera and a DLS-H37DM-A infrared imager. RGBT210~\cite{RGBT210} is a large-scale dataset that was proposed in 2017, containing 210 pairs of videos and 210k frames in total and providing attribute annotations as well as occlusion annotations. In 2018, another 24 video pairs were added to RGBT210, and a new dataset, RGBT234~\cite{RGBT234}, was generated. This dataset provides around 234k frames in total, which are sufficient for performance evaluation. Besides, more accurate attributes and evaluation metrics are considered for measuring performance.\\
\textbf{VOT-RGBT.}
VOT-RGBT is a subset of VOT datasets mentioned in Sec.~\ref{sec:rgbdataset}. Specifically, both VOT-RGBT2019\cite{vot2019} and VOT-RGBT2020\cite{vot2020} challenges provide a dataset consisting of 60 visible and infrared video pairs from RGBT234~\cite{RGBT234}, focusing on short-term RGB-T tracking. To achieve a higher significance of results, VOT-RGBT2019 further extends the annotations of the original RGBT234 with rotated bounding box labels. The difference between the VOT-RGBT2019 and VOT-RGBT2020 is that VOT-RGBT2020 extended anchor frames for the new re-initialization challenge. For both datasets, three evaluation measures named accuracy (A), robustness (R), and Expected Average Overlap (EAO) are used to estimate the results.\\
\textbf{LSS.}
 LSS~\cite{SiamCDA} is a synthetic RGB-T dataset that has 12k synthetic thermal images and 3862 synthetic thermal videos produced from COCO and VID datasets. It uses the semantic-aware image-to-image translation method to generate synthetic thermal images from real RGB images.
 Additionally, it includes 969 synthetic RGB videos created by a video colorization method~\cite{lei2019fully}. LSS has around 40 categories, covering animals to cars.\\
\textbf{LasHeR.}
LasHeR~\cite{LasHeR} is a large-scale RGB-T dataset with 1224 video pairs and more than 730k frames in total. This dataset contains 32 object classes captured from real-world scenarios with densely annotated and spatially aligned. Besides, LasHeR is extensive in terms of challenge diversity, including aspect ratio change, frame loss, camera moving, and so on. \\
\textbf{VTUAV.}
VTUAV\cite{HMFT} is the latest RGB-T dataset, which has 500 sequences with 1.7 million high-resolution (1920$\times$ 1080) frame
pairs. It is a large-scale benchmark with high diversity for RGB-T UAV tracking. The tracked target can be broken down into 13 sub-classes and 5 super-classes (pedestrian, vehicle, animal, train, and ship), which can cover the majority of categories for practical applications. Moreover,
VTUAV includes short-term, long-term tracking and segmentation mask prediction to achieve a comprehensive evaluation with wider applications.

\subsection{RGB-LiDAR dataset}
For the RGB-LiDAR VOT, the most common dataset used in current research turns out to be the KITTI~\cite{KITTI2012} dataset as introduced in Sec~\ref{sec:lidardataset}. 

\subsection{RGB-Language dataset}
\textbf{OTB-Lang.}
The OTB-Lang dataset was first proposed by Li \textit{et al.}~\cite{li2017tracking}. The authors augment the videos in OTB100 with natural language descriptions of the target object. The annotators are asked to make the discriminative referring description of the target. The total number of video-language pairs is 99, as one video can not be annotated by the annotators.\\
\textbf{LaSOT.}
LaSOT~\cite{lasot} contains 1,400 videos with language annotation. The dataset is split into training and testing splits, with 1,120 videos for training and
280 videos for testing. This benchmark is more challenging than OTB-Lang. The foreground objects constantly suffer occlusion and the background contains various distractors.\\
\textbf{TNL2K.}
TNL2K~\cite{tnl2k} is the most recent benchmark, containing 2000 long video sequences, most of which are downloaded and clipped from Youtube, surveillance cameras, and mobile phones. Each video has one sentence describing the target and each frame has one bounding box to indicate the localization of the target. The descriptions contain 663 English words and express the attributes and locations of target objects. This benchmark is more challenging than the OTB-Lang~\cite{li2017tracking} and LaSOT~\cite{lasot}.

%% file: discussion.tex
	\section{Discussion}~\label{sec:discussion}
	In the previous sections, we comprehensively review the existing VOT methods from the perspective of various data modalities. Next, we will discuss some promising and potential directions of VOT discovered from the content we summarized.
\subsection{\textcolor{black}{Parameter-Efficient Transfer Learning}}
 \textcolor{black}{Parameter-Efficient Fine-Tuning (PEFT) has become a highly popular direction in the past couple of years, widely adopted across various tasks from NLP~\cite{houlsby2019parameter} to computer vision~\cite{cocoop,maple}. In VOT, there is also a growing trend where methods are employing efficient transfer learning techniques to transfer single RGB modal models to multi-modal tracking tasks, like Protrack\cite{protrack}, ViPT~\cite{zhu2023vipt} and MPLT~\cite{mplt}. 
Previous attempts in multi-modal tracking primarily utilized the "pre-trained RGB baselines + multi-modal data fine-tuned" paradigm. However, this paradigm of overtraining may lead to a knowledge forgetting problem. PEFT has been demonstrated as a superior approach to leverage the knowledge of pre-trained RGB-based models at scale, while only introducing a few trainable parameters~\cite{m2clip,zhu2023vipt}. This approach significantly reduces the risk of overfitting. Therefore, PEFT represents a promising research direction for VOT. Currently, it is primarily applied to modalities such as RGB-Depth and RGB-Thermal. However, other multi-modalities can also explore the application of this technique.
 }
	\subsection{Schema}
	As discussed in Section~\ref{sec:unifiedTrans}, the latest RGB-based VOT begins to employ a unified Transformer schema~\cite{mixformer2022,sbt2022,ostrack2022,simtrack2022}, without Siamese backbones and separated matchers. In other literature like object detection~\cite{chen2021pix2seq}, Transformers have also shown excellent capabilities to simplify frameworks. The combination of feature extraction and matcher into a unified Transformer largely simplifies the conventional DCF and siamese paradigms, while achieving promising results. This implies that the unified paradigm will have great potential to cover the VOT task of other data modalities. In addition, for RGB modality, this schema also remains plenty of room to explore, for example, architectures, efficiency, online learning, etc.
	
\subsection{Backbone Architectures}
	In the deep-learning era, the majority of trackers are dominated by the traditional CNNs like AlexNet~\cite{alexnet}, GoogLeNet~\cite{googlenet} and ResNet~\cite{resnet}. Until recently, the vision Transformer revolutionizes many computer vision tasks that include VOT~\cite{stark2021,transT2021,aiatrack2022,CSWinTT2022}. Significantly, the well-known cross-correlation operation~\cite{siamfc,siamrpn++2019} could be replaced with the cross-attention mechanism of the Transformer. Also, the previous backbone architectures like ResNet could be replaced with more powerful Transformer backbones like ViT~\cite{vit} and Swin-Transformer~\cite{swin}. \textcolor{black}{While Transformers have the potential for modality interaction and fusion~\cite{clip,kim2021vilt,wang2021actionclip} for multi-modality VOT, effectively integrating information from different perceptual modalities remains a challenge. Each perceptual modality possesses unique features and representations, and further research is needed to explore how to effectively fuse them together to enhance tracking performance.}
	Therefore, we believe that the vision Transformer deserves further exploration in VOT.
	
	\subsection{Online Learning}
	Online model updating is a well-worn problem in VOT, which aims to adapt the tracking model to the target appearance changes. This component is essential in the DCF-based and ICD-based schemas, while the siamese-based and OST-based schemas often adopt an offline-trained and online-fixed strategy without the model update to keep high tracking speed. There is a speed-accuracy dilemma for the online learning of VOT. Early handcraft-feature-based DCF methods~\cite{henriques2012exploiting,henriques2014high} rely on the circular matrix and FFT to improve efficiency. The DNN-based model updaters usually choose to control the update frequency~\cite{stark2021}, try optimizers with fast convergence~\cite{DiMP} and employ meta learning~\cite{dai2020high,roam2020,mlt2019}. However, achieving the speed-accuracy trade-off remains an open question, especially in the non-RGB VOT areas. For example, there are no LiDAR-based VOT trackers equipped with model update modules.
	
	\subsection{Unsupervised Learning}
	Unsupervised learning in VOT is an emerging problem, since the DNN-based methods are data hungry while constructing extensive benchmarks requires substantial storage resources, enormous human effort and time. Meanwhile, unsupervised learning could leverage the availability of unlabelled data to train the models, significantly reducing the requirement for large labeled datasets. Hence, some researchers have begun to explore the unsupervised methods, like ResPUL~\cite{respul}, S2Siam~\cite{s2siamfc}, LUDT~\cite{wang2021unsupervised}, UDT~\cite{wang2019unsupervised}, CycleSiam~\cite{yuan2020self} USOT~\cite{zheng2021learning} and ULAST~\cite{shen2022unsupervised}. The unsupervised learning strategies provide more possibilities in practical or specific applications when the training data is not annotated. In addition, the unsupervised learning methods will be useful in multi-modal VOT, which usually lacks well-established benchmarks. 
	
	\subsection{Long-term Tracking}
 \textcolor{black}{Most existing methods focus on short-term tracking, assuming that the target is always in the field of view. In contrast, long-term tracking algorithms are designed for scenarios where the target may disappear from the field of view, remain fully occluded for extended periods, and encounter cuts, such as unpredictable abrupt changes in target pose and appearance~\cite{lukezivc2020performance}.}
This is difficult since the trackers easily drift into extremely hard challenges like occlusions, out-of-the-view and distractors.
There are VOT methods focused on long-term tracking, like SPLT~\cite{splt2019}, Globaltrack~\cite{globaltrack2020}, LTMU~\cite{ltmu2020}, SiamRCNN~\cite{SiamRCNN2020}, DMTrack~\cite{dmtrack2021} and Zhou et  al.~\cite{zhou2022global}. These methods will equip with a re-detection module or ensemble several local trackers to re-locate the targets when encountering difficult target lost situations. We have observed that long-term tracking is mainly studied  in the RGB modality, but is also an important function in many real-world scenes with different modalities.
	
\subsection{Multi-task Learning}
	Multi-task learning means the model is learned with the guidance of VOT labels and other related tasks like segmentation and multiple object tracking (MOT). Some works take advantage of multi-task learning to improve VOT performance and realize multiple functions. There are mainly two kinds of multi-task models. \textcolor{black}{The first category is VOT with object segmentation, like RTS~\cite{rts2022}, which output the pixel-level object masks and the object bounding boxes together.}
	The second has emerged recently, namely, VOT with MOT, like Unicorn~\cite{unicorn} and UTT~\cite{utt2022}, which unify these two tasks into one framework and achieve one tracker for all tracking tasks. What tasks to merge and how to integrate remain to be open questions and are fascinating directions for research.
	
	\subsection{Multi-modal Tracking}
	Multi-modal tracking is a prospective research direction that has become popular recently, since one sensor could not handle all the diverse and dynamic changes in the real world. However, we find that some multi-modal combinations of VOT could be better explored, like RGB-Language and RGB-LiDAR. At the same time, they are helpful in practical applications, like video conferencing equipment, human-machine interactions and autonomous driving. The way of cross-modal interaction and fusion, single-modal representations and the decision-making strategies deserve further explorations.
	We have noticed that recent research of RGB-LiDAR is a hot topic in 3D detection~\cite{liang2022bevfusion,liu2022bevfusion,bai2022transfusion}, which exploit the Bird Eye view (BEV) of feature representation and fusion. Similarly, the RGB-Language fusion technology has been researched a lot in recently, like in image classification~\cite{clip}, grounding~\cite{glip}, referring segmentation~\cite{li2021mail} and so on. This suggested that multi-modal learning will be an essential and promising direction for VOT.

\subsection{\textcolor{black}{Transparent Object Tracking}}
\textcolor{black}{Current research in VOT field mainly focuses on opaque object tracking, while very little attention is paid to transparent objects~\cite{transparent,transparent2023,transparent2024new,transT2021}. However, this presents a highly practical direction, which can be widely applied in autonomous driving systems, industrial automation and the medical field. Tracking transparent objects is more challenging compared to opaque objects due to their unique characteristics.  The inherent challenge lies in the distinctive traits of transparent objects, characterized by subtle appearances influenced by their surroundings, complicating conventional tracking methodologies. Designing tracking systems specifically tailored for transparent objects constitutes a challenging yet valuable research direction worthy of considerable attention.}
	
	\subsection{Datasets}
It is well known that DNN-based models are always data-hungry. This is also true in VOT. The deep trackers flourished when large-scale datasets like LaSOT~\cite{lasot}, TrackingNet~\cite{trackingnet} and GOT-10k~\cite{got10k} were proposed. However, in some modalities, such as LiDAR and the multi-modal VOT, the size of the datasets is still small for training large DNNs, which is one of the most significant limitations that hinders further promotions. However, constructing a large-scale dataset is not easy, requiring a large amount of time, labor and storage resources. Therefore, it is desired and urgent to find fast annotation technologies to reduce the benchmark construction cost. Diverse and large datasets are still required to facilitate and advance the VOT areas.

%% file: final.bbl
\begin{thebibliography}{100}
\providecommand{\url}[1]{#1}
\csname url@samestyle\endcsname
\providecommand{\newblock}{\relax}
\providecommand{\bibinfo}[2]{#2}
\providecommand{\BIBentrySTDinterwordspacing}{\spaceskip=0pt\relax}
\providecommand{\BIBentryALTinterwordstretchfactor}{4}
\providecommand{\BIBentryALTinterwordspacing}{\spaceskip=\fontdimen2\font plus
\BIBentryALTinterwordstretchfactor\fontdimen3\font minus
  \fontdimen4\font\relax}
\providecommand{\BIBforeignlanguage}[2]{{%
\expandafter\ifx\csname l@#1\endcsname\relax
\typeout{** WARNING: IEEEtran.bst: No hyphenation pattern has been}%
\typeout{** loaded for the language `#1'. Using the pattern for}%
\typeout{** the default language instead.}%
\else
\language=\csname l@#1\endcsname
\fi
#2}}
\providecommand{\BIBdecl}{\relax}
\BIBdecl

\bibitem{simtrack2022}
B.~Chen, P.~Li, L.~Bai, L.~Qiao, Q.~Shen, B.~Li, W.~Gan, W.~Wu, and W.~Ouyang,
  ``Backbone is all your need: A simplified architecture for visual object
  tracking,'' \emph{arXiv preprint arXiv:2203.05328}, 2022.

\bibitem{henriques2014high}
J.~F. Henriques, R.~Caseiro, P.~Martins, and J.~Batista, ``High-speed tracking
  with kernelized correlation filters,'' \emph{IEEE transactions on pattern
  analysis and machine intelligence}, vol.~37, no.~3, pp. 583--596, 2014.

\bibitem{siamrpn2018}
B.~Li, J.~Yan, W.~Wu, Z.~Zhu, and X.~Hu, ``High performance visual tracking
  with siamese region proposal network,'' in \emph{Proceedings of the IEEE
  conference on computer vision and pattern recognition}, 2018, pp. 8971--8980.

\bibitem{3dsiameserpn}
Z.~Fang, S.~Zhou, Y.~Cui, and S.~Scherer, ``3d-siamrpn: An end-to-end learning
  method for real-time 3d single object tracking using raw point cloud,''
  \emph{IEEE Sensors Journal}, vol.~21, no.~4, pp. 4995--5011, 2020.

\bibitem{v2b}
L.~Hui, L.~Wang, M.~Cheng, J.~Xie, and J.~Yang, ``3d siamese voxel-to-bev
  tracker for sparse point clouds,'' \emph{Advances in Neural Information
  Processing Systems}, vol.~34, pp. 28\,714--28\,727, 2021.

\bibitem{wang2017}
M.~Wang, D.~Su, L.~Shi, Y.~Liu, and J.~V. Miro, ``Real-time 3d human tracking
  for mobile robots with multisensors,'' in \emph{2017 IEEE International
  Conference on Robotics and Automation (ICRA)}.\hskip 1em plus 0.5em minus
  0.4em\relax IEEE, 2017, pp. 5081--5087.

\bibitem{wang2018}
M.~Wang, Y.~Liu, D.~Su, Y.~Liao, L.~Shi, J.~Xu, and J.~V. Miro, ``Accurate and
  real-time 3-d tracking for the following robots by fusing vision and
  ultrasonar information,'' \emph{IEEE/ASME Transactions On Mechatronics},
  vol.~23, no.~3, pp. 997--1006, 2018.

\bibitem{gti}
Z.~Yang, T.~Kumar, T.~Chen, J.~Su, and J.~Luo,
  ``Grounding-tracking-integration,'' \emph{IEEE Transactions on Circuits and
  Systems for Video Technology}, vol.~31, no.~9, pp. 3433--3443, 2020.

\bibitem{snlt}
Q.~Feng, V.~Ablavsky, Q.~Bai, and S.~Sclaroff, ``Siamese natural language
  tracker: Tracking by natural language descriptions with siamese trackers,''
  in \emph{Proceedings of the IEEE/CVF Conference on Computer Vision and
  Pattern Recognition}, 2021, pp. 5851--5860.

\bibitem{kalal2011tracking}
Z.~Kalal, K.~Mikolajczyk, and J.~Matas, ``Tracking-learning-detection,''
  \emph{IEEE transactions on pattern analysis and machine intelligence},
  vol.~34, no.~7, pp. 1409--1422, 2011.

\bibitem{hare2015struck}
S.~Hare, S.~Golodetz, A.~Saffari, V.~Vineet, M.-M. Cheng, S.~L. Hicks, and
  P.~H. Torr, ``Struck: Structured output tracking with kernels,'' \emph{IEEE
  transactions on pattern analysis and machine intelligence}, vol.~38, no.~10,
  pp. 2096--2109, 2015.

\bibitem{dsst2016}
M.~Danelljan, G.~H{\"a}ger, F.~S. Khan, and M.~Felsberg, ``Discriminative scale
  space tracking,'' \emph{IEEE transactions on pattern analysis and machine
  intelligence}, vol.~39, no.~8, pp. 1561--1575, 2016.

\bibitem{ostrack2022}
B.~Ye, H.~Chang, B.~Ma, S.~Shan, and X.~Chen, ``Joint feature learning and
  relation modeling for tracking: A one-stream framework,'' in \emph{European
  Conference on Computer Vision}.\hskip 1em plus 0.5em minus 0.4em\relax
  Springer, 2022, pp. 341--357.

\bibitem{chen2020siamese}
Z.~Chen, B.~Zhong, G.~Li, S.~Zhang, and R.~Ji, ``Siamese box adaptive network
  for visual tracking,'' in \emph{Proceedings of the IEEE/CVF conference on
  computer vision and pattern recognition}, 2020, pp. 6668--6677.

\bibitem{fear2022}
V.~Borsuk, R.~Vei, O.~Kupyn, T.~Martyniuk, I.~Krashenyi, and J.~Matas, ``Fear:
  Fast, efficient, accurate and robust visual tracker,'' in \emph{European
  Conference on Computer Vision}.\hskip 1em plus 0.5em minus 0.4em\relax
  Springer, 2022, pp. 644--663.

\bibitem{siamfc}
L.~Bertinetto, J.~Valmadre, J.~F. Henriques, A.~Vedaldi, and P.~H. Torr,
  ``Fully-convolutional siamese networks for object tracking,'' in
  \emph{European conference on computer vision}.\hskip 1em plus 0.5em minus
  0.4em\relax Springer, 2016, pp. 850--865.

\bibitem{DiMP}
G.~Bhat, M.~Danelljan, L.~V. Gool, and R.~Timofte, ``Learning discriminative
  model prediction for tracking,'' in \emph{Proceedings of the IEEE/CVF
  international conference on computer vision}, 2019, pp. 6182--6191.

\bibitem{lasot}
H.~Fan, L.~Lin, F.~Yang, P.~Chu, G.~Deng, S.~Yu, H.~Bai, Y.~Xu, C.~Liao, and
  H.~Ling, ``Lasot: A high-quality benchmark for large-scale single object
  tracking,'' in \emph{Proceedings of the IEEE/CVF conference on computer
  vision and pattern recognition}, 2019, pp. 5374--5383.

\bibitem{got10k}
L.~Huang, X.~Zhao, and K.~Huang, ``Got-10k: A large high-diversity benchmark
  for generic object tracking in the wild,'' \emph{IEEE Transactions on Pattern
  Analysis and Machine Intelligence}, vol.~43, no.~5, pp. 1562--1577, 2019.

\bibitem{trackingnet}
M.~Muller, A.~Bibi, S.~Giancola, S.~Alsubaihi, and B.~Ghanem, ``Trackingnet: A
  large-scale dataset and benchmark for object tracking in the wild,'' in
  \emph{Proceedings of the European conference on computer vision (ECCV)},
  2018, pp. 300--317.

\bibitem{mayer2022transforming}
C.~Mayer, M.~Danelljan, G.~Bhat, M.~Paul, D.~P. Paudel, F.~Yu, and L.~Van~Gool,
  ``Transforming model prediction for tracking,'' in \emph{Proceedings of the
  IEEE/CVF Conference on Computer Vision and Pattern Recognition}, 2022, pp.
  8731--8740.

\bibitem{siamrpn++2019}
B.~Li, W.~Wu, Q.~Wang, F.~Zhang, J.~Xing, and J.~Yan, ``Siamrpn++: Evolution of
  siamese visual tracking with very deep networks,'' in \emph{Proceedings of
  the IEEE/CVF Conference on Computer Vision and Pattern Recognition}, 2019,
  pp. 4282--4291.

\bibitem{siamcar2020}
D.~Guo, J.~Wang, Y.~Cui, Z.~Wang, and S.~Chen, ``Siamcar: Siamese fully
  convolutional classification and regression for visual tracking,'' in
  \emph{Proceedings of the IEEE/CVF conference on computer vision and pattern
  recognition}, 2020, pp. 6269--6277.

\bibitem{siamban2022}
Z.~Chen, B.~Zhong, G.~Li, S.~Zhang, R.~Ji, Z.~Tang, and X.~Li, ``Siamban:
  Target-aware tracking with siamese box adaptive network,'' \emph{IEEE
  Transactions on Pattern Analysis and Machine Intelligence}, 2022.

\bibitem{MDNet}
H.~Nam and B.~Han, ``Learning multi-domain convolutional neural networks for
  visual tracking,'' in \emph{Proceedings of the IEEE conference on computer
  vision and pattern recognition}, 2016, pp. 4293--4302.

\bibitem{uct2017}
Z.~Zhu, G.~Huang, W.~Zou, D.~Du, and C.~Huang, ``Uct: Learning unified
  convolutional networks for real-time visual tracking,'' in \emph{Proceedings
  of the IEEE international conference on computer vision workshops}, 2017, pp.
  1973--1982.

\bibitem{han2017branchout}
B.~Han, J.~Sim, and H.~Adam, ``Branchout: Regularization for online ensemble
  tracking with convolutional neural networks,'' in \emph{Proceedings of the
  IEEE conference on computer vision and pattern recognition}, 2017, pp.
  3356--3365.

\bibitem{sbt2022}
F.~Xie, C.~Wang, G.~Wang, Y.~Cao, W.~Yang, and W.~Zeng, ``Correlation-aware
  deep tracking,'' in \emph{Proceedings of the IEEE/CVF Conference on Computer
  Vision and Pattern Recognition}, 2022, pp. 8751--8760.

\bibitem{mixformer2022}
Y.~Cui, C.~Jiang, L.~Wang, and G.~Wu, ``Mixformer: End-to-end tracking with
  iterative mixed attention,'' in \emph{Proceedings of the IEEE/CVF Conference
  on Computer Vision and Pattern Recognition}, 2022, pp. 13\,608--13\,618.

\bibitem{ECO-MM}
Q.~Liu, D.~Yuan, N.~Fan, P.~Gao, X.~Li, and Z.~He, ``Learning dual-level deep
  representation for thermal infrared tracking,'' \emph{IEEE Transactions on
  Multimedia}, 2022.

\bibitem{GFSNet}
R.~Chen, S.~Liu, Z.~Miao, and F.~Li, ``Gfsnet: Generalization-friendly siamese
  network for thermal infrared object tracking,'' \emph{Infrared Physics \&
  Technology}, p. 104190, 2022.

\bibitem{ECO-LS}
H.~Zhang, Z.~Yin, and H.~Zhang, ``Thermal infrared object tracking using
  correlation filters improved by level set,'' \emph{Signal, Image and Video
  Processing}, pp. 1--7, 2022.

\bibitem{p2b}
H.~Qi, C.~Feng, Z.~Cao, F.~Zhao, and Y.~Xiao, ``P2b: Point-to-box network for
  3d object tracking in point clouds,'' in \emph{Proceedings of the IEEE/CVF
  Conference on Computer Vision and Pattern Recognition}, 2020, pp. 6329--6338.

\bibitem{bat}
C.~Zheng, X.~Yan, J.~Gao, W.~Zhao, W.~Zhang, Z.~Li, and S.~Cui, ``Box-aware
  feature enhancement for single object tracking on point clouds,'' in
  \emph{Proceedings of the IEEE/CVF International Conference on Computer
  Vision}, 2021, pp. 13\,199--13\,208.

\bibitem{m2track}
C.~Zheng, X.~Yan, H.~Zhang, B.~Wang, S.~Cheng, S.~Cui, and Z.~Li, ``Beyond 3d
  siamese tracking: A motion-centric paradigm for 3d single object tracking in
  point clouds,'' in \emph{Proceedings of the IEEE/CVF Conference on Computer
  Vision and Pattern Recognition}, 2022, pp. 8111--8120.

\bibitem{SOWP}
H.-U. Kim, D.-Y. Lee, J.-Y. Sim, and C.-S. Kim, ``Sowp: Spatially ordered and
  weighted patch descriptor for visual tracking,'' in \emph{Proceedings of the
  IEEE International Conference on Computer Vision}, 2015, pp. 3011--3019.

\bibitem{SGT}
C.~Li, N.~Zhao, Y.~Lu, C.~Zhu, and J.~Tang, ``Weighted sparse representation
  regularized graph learning for rgb-t object tracking,'' in \emph{Proceedings
  of the 25th ACM international conference on Multimedia}, 2017, pp.
  1856--1864.

\bibitem{ADRNet}
P.~Zhang, D.~Wang, H.~Lu, and X.~Yang, ``Learning adaptive attribute-driven
  representation for real-time rgb-t tracking,'' \emph{International Journal of
  Computer Vision}, vol. 129, no.~9, pp. 2714--2729, 2021.

\bibitem{F-siamese}
H.~Zou, J.~Cui, X.~Kong, C.~Zhang, Y.~Liu, F.~Wen, and W.~Li, ``F-siamese
  tracker: A frustum-based double siamese network for 3d single object
  tracking,'' in \emph{2020 IEEE/RSJ International Conference on Intelligent
  Robots and Systems (IROS)}.\hskip 1em plus 0.5em minus 0.4em\relax IEEE,
  2020, pp. 8133--8139.

\bibitem{alireza}
A.~Asvadi, P.~Girao, P.~Peixoto, and U.~Nunes, ``3d object tracking using rgb
  and lidar data,'' in \emph{2016 IEEE 19th International Conference on
  Intelligent Transportation Systems (ITSC)}.\hskip 1em plus 0.5em minus
  0.4em\relax IEEE, 2016, pp. 1255--1260.

\bibitem{guo2022divert}
M.~Guo, Z.~Zhang, H.~Fan, and L.~Jing, ``Divert more attention to
  vision-language tracking,'' \emph{arXiv preprint arXiv:2207.01076}, 2022.

\bibitem{fiaz2019handcrafted}
M.~Fiaz, A.~Mahmood, S.~Javed, and S.~K. Jung, ``Handcrafted and deep trackers:
  Recent visual object tracking approaches and trends,'' \emph{ACM Computing
  Surveys (CSUR)}, vol.~52, no.~2, pp. 1--44, 2019.

\bibitem{javed2022visual}
S.~Javed, M.~Danelljan, F.~S. Khan, M.~H. Khan, M.~Felsberg, and J.~Matas,
  ``Visual object tracking with discriminative filters and siamese networks: a
  survey and outlook,'' \emph{IEEE Transactions on Pattern Analysis and Machine
  Intelligence}, vol.~45, no.~5, pp. 6552--6574, 2022.

\bibitem{ondravsovivc2021siamese}
M.~Ondra{\v{s}}ovi{\v{c}} and P.~Tar{\'a}bek, ``Siamese visual object tracking:
  A survey,'' \emph{IEEE Access}, vol.~9, pp. 110\,149--110\,172, 2021.

\bibitem{zhang2021recent}
Y.~Zhang, T.~Wang, K.~Liu, B.~Zhang, and L.~Chen, ``Recent advances of
  single-object tracking methods: A brief survey,'' \emph{Neurocomputing}, vol.
  455, pp. 1--11, 2021.

\bibitem{chen2022visual}
F.~Chen, X.~Wang, Y.~Zhao, S.~Lv, and X.~Niu, ``Visual object tracking: A
  survey,'' \emph{Computer Vision and Image Understanding}, vol. 222, p.
  103508, 2022.

\bibitem{marvasti2021deep}
S.~M. Marvasti-Zadeh, L.~Cheng, H.~Ghanei-Yakhdan, and S.~Kasaei, ``Deep
  learning for visual tracking: A comprehensive survey,'' \emph{IEEE
  Transactions on Intelligent Transportation Systems}, 2021.

\bibitem{li2018deep}
P.~Li, D.~Wang, L.~Wang, and H.~Lu, ``Deep visual tracking: Review and
  experimental comparison,'' \emph{Pattern Recognition}, vol.~76, pp. 323--338,
  2018.

\bibitem{soleimanitaleb2022single}
Z.~Soleimanitaleb and M.~A. Keyvanrad, ``Single object tracking: A survey of
  methods, datasets, and evaluation metrics,'' \emph{arXiv preprint
  arXiv:2201.13066}, 2022.

\bibitem{han2022single}
R.~Han, W.~Feng, Q.~Guo, and Q.~Hu, ``Single object tracking research: A
  survey,'' \emph{arXiv preprint arXiv:2204.11410}, 2022.

\bibitem{zhang2020multi}
P.~Zhang, D.~Wang, and H.~Lu, ``Multi-modal visual tracking: Review and
  experimental comparison,'' \emph{arXiv preprint arXiv:2012.04176}, 2020.

\bibitem{walia2016recent}
G.~S. Walia and R.~Kapoor, ``Recent advances on multicue object tracking: a
  survey,'' \emph{Artificial Intelligence Review}, vol.~46, no.~1, pp. 1--39,
  2016.

\bibitem{kumar2020recent}
A.~Kumar, G.~S. Walia, and K.~Sharma, ``Recent trends in multicue based visual
  tracking: A review,'' \emph{Expert Systems with Applications}, vol. 162, p.
  113711, 2020.

\bibitem{yang2011recent}
H.~Yang, L.~Shao, F.~Zheng, L.~Wang, and Z.~Song, ``Recent advances and trends
  in visual tracking: A review,'' \emph{Neurocomputing}, vol.~74, no.~18, pp.
  3823--3831, 2011.

\bibitem{li2013survey}
X.~Li, W.~Hu, C.~Shen, Z.~Zhang, A.~Dick, and A.~V.~D. Hengel, ``A survey of
  appearance models in visual object tracking,'' \emph{ACM transactions on
  Intelligent Systems and Technology (TIST)}, vol.~4, no.~4, pp. 1--48, 2013.

\bibitem{smeulders2013visual}
A.~W. Smeulders, D.~M. Chu, R.~Cucchiara, S.~Calderara, A.~Dehghan, and
  M.~Shah, ``Visual tracking: An experimental survey,'' \emph{IEEE transactions
  on pattern analysis and machine intelligence}, vol.~36, no.~7, pp.
  1442--1468, 2013.

\bibitem{wang2019comparison}
J.~Wang, L.~Zheng, M.~Tang, and J.~Feng, ``A comparison of correlation
  filter-based trackers and struck trackers,'' \emph{IEEE Transactions on
  Circuits and Systems for Video Technology}, vol.~30, no.~9, pp. 3106--3118,
  2019.

\bibitem{zhang2013sparse}
S.~Zhang, H.~Yao, X.~Sun, and X.~Lu, ``Sparse coding based visual tracking:
  Review and experimental comparison,'' \emph{Pattern Recognition}, vol.~46,
  no.~7, pp. 1772--1788, 2013.

\bibitem{pflugfelder2017depth}
R.~Pflugfelder, ``An in-depth analysis of visual tracking with siamese neural
  networks,'' \emph{arXiv preprint arXiv:1707.00569}, 2017.

\bibitem{yilmaz2006object}
A.~Yilmaz, O.~Javed, and M.~Shah, ``Object tracking: A survey,'' \emph{Acm
  computing surveys (CSUR)}, vol.~38, no.~4, pp. 13--es, 2006.

\bibitem{yang2022rgbd}
J.~Yang, Z.~Li, S.~Yan, F.~Zheng, A.~Leonardis, J.-K. K{\"a}m{\"a}r{\"a}inen,
  and L.~Shao, ``Rgbd object tracking: An in-depth review,'' \emph{arXiv
  preprint arXiv:2203.14134}, 2022.

\bibitem{zhang2020object}
X.~Zhang, P.~Ye, H.~Leung, K.~Gong, and G.~Xiao, ``Object fusion tracking based
  on visible and infrared images: A comprehensive review,'' \emph{Information
  Fusion}, vol.~63, pp. 166--187, 2020.

\bibitem{vot2018}
M.~Kristan, A.~Leonardis, J.~Matas, M.~Felsberg, R.~Pflugfelder,
  L.~ˇCehovin~Zajc, T.~Vojir, G.~Bhat, A.~Lukezic, A.~Eldesokey \emph{et~al.},
  ``The sixth visual object tracking vot2018 challenge results,'' in
  \emph{Proceedings of the European Conference on Computer Vision (ECCV)
  Workshops}, 2018, pp. 0--0.

\bibitem{uav123}
M.~Mueller, N.~Smith, and B.~Ghanem, ``A benchmark and simulator for uav
  tracking,'' in \emph{European conference on computer vision}.\hskip 1em plus
  0.5em minus 0.4em\relax Springer, 2016, pp. 445--461.

\bibitem{nfs}
H.~Kiani~Galoogahi, A.~Fagg, C.~Huang, D.~Ramanan, and S.~Lucey, ``Need for
  speed: A benchmark for higher frame rate object tracking,'' in
  \emph{Proceedings of the IEEE International Conference on Computer Vision},
  2017, pp. 1125--1134.

\bibitem{otb100}
Y.~Wu, J.~Lim, and M.-H. Yang, ``Object tracking benchmark,'' \emph{IEEE
  transactions on pattern analysis and machine intelligence}, vol.~37, no.~9,
  pp. 1834–--1848, 2015.

\bibitem{zhu2021robust}
X.-F. Zhu, X.-J. Wu, T.~Xu, Z.-H. Feng, and J.~Kittler, ``Robust visual object
  tracking via adaptive attribute-aware discriminative correlation filters,''
  \emph{IEEE Transactions on Multimedia}, vol.~24, pp. 301--312, 2021.

\bibitem{BWRR2021}
H.~Zhu, H.~Peng, G.~Xu, L.~Deng, Y.~Cheng, and A.~Song, ``Bilateral weighted
  regression ranking model with spatial-temporal correlation filter for visual
  tracking,'' \emph{IEEE Transactions on Multimedia}, vol.~24, pp. 2098--2111,
  2021.

\bibitem{yang2021siamcorners}
K.~Yang, Z.~He, W.~Pei, Z.~Zhou, X.~Li, D.~Yuan, and H.~Zhang, ``Siamcorners:
  Siamese corner networks for visual tracking,'' \emph{IEEE Transactions on
  Multimedia}, vol.~24, pp. 1956--1967, 2021.

\bibitem{jiang2020stgl}
B.~Jiang, Y.~Zhang, B.~Luo, X.~Cao, and J.~Tang, ``Stgl: Spatial-temporal graph
  representation and learning for visual tracking,'' \emph{IEEE Transactions on
  Multimedia}, vol.~23, pp. 2162--2171, 2020.

\bibitem{fan2021discriminative}
B.~Fan, J.~Tian, Y.~Peng, and Y.~Tang, ``Discriminative siamese complementary
  tracker with flexible update,'' \emph{IEEE Transactions on Multimedia}, 2021.

\bibitem{chan2022siamese}
S.~Chan, J.~Tao, X.~Zhou, C.~Bai, and X.~Zhang, ``Siamese implicit region
  proposal network with compound attention for visual tracking,'' \emph{IEEE
  Transactions on Image Processing}, vol.~31, pp. 1882--1894, 2022.

\bibitem{li2022}
Z.~Li, J.~Zhang, Y.~Li, J.~Zhu, S.~Long, D.~Xue, and L.~Fan, ``Learning feature
  channel weighting for real-time visual tracking,'' \emph{IEEE Transactions on
  Image Processing}, vol.~31, pp. 2190--2200, 2022.

\bibitem{zhou2022global}
Z.~Zhou, J.~Chen, W.~Pei, K.~Mao, H.~Wang, and Z.~He, ``Global tracking via
  ensemble of local trackers,'' in \emph{Proceedings of the IEEE/CVF Conference
  on Computer Vision and Pattern Recognition}, 2022, pp. 8761--8770.

\bibitem{utt2022}
F.~Ma, M.~Z. Shou, L.~Zhu, H.~Fan, Y.~Xu, Y.~Yang, and Z.~Yan, ``Unified
  transformer tracker for object tracking,'' in \emph{Proceedings of the
  IEEE/CVF Conference on Computer Vision and Pattern Recognition}, 2022, pp.
  8781--8790.

\bibitem{RBO2022}
F.~Tang and Q.~Ling, ``Ranking-based siamese visual tracking,'' in
  \emph{Proceedings of the IEEE/CVF Conference on Computer Vision and Pattern
  Recognition}, 2022, pp. 8741--8750.

\bibitem{CSWinTT2022}
Z.~Song, J.~Yu, Y.-P.~P. Chen, and W.~Yang, ``Transformer tracking with cyclic
  shifting window attention,'' in \emph{Proceedings of the IEEE/CVF Conference
  on Computer Vision and Pattern Recognition}, 2022, pp. 8791--8800.

\bibitem{InBN2022}
M.~Guo, Z.~Zhang, H.~Fan, L.~Jing, Y.~Lyu, B.~Li, and W.~Hu, ``Learning
  target-aware representation for visual tracking via informative
  interactions,'' \emph{arXiv preprint arXiv:2201.02526}, 2022.

\bibitem{sparsett2022}
Z.~Fu, Z.~Fu, Q.~Liu, W.~Cai, and Y.~Wang, ``Sparsett: Visual tracking with
  sparse transformers,'' \emph{arXiv preprint arXiv:2205.03776}, 2022.

\bibitem{HybTransT2022}
I.~Jung, M.~Kim, E.~Park, and B.~Han, ``Online hybrid lightweight
  representations learning: Its application to visual tracking,'' \emph{arXiv
  preprint arXiv:2205.11179}, 2022.

\bibitem{SLTtrack2022}
M.~Kim, S.~Lee, J.~Ok, B.~Han, and M.~Cho, ``Towards sequence-level training
  for visual tracking,'' in \emph{European Conference on Computer
  Vision}.\hskip 1em plus 0.5em minus 0.4em\relax Springer, 2022, pp. 534--551.

\bibitem{unicorn}
B.~Yan, Y.~Jiang, P.~Sun, D.~Wang, Z.~Yuan, P.~Luo, and H.~Lu, ``Towards grand
  unification of object tracking,'' in \emph{European Conference on Computer
  Vision}.\hskip 1em plus 0.5em minus 0.4em\relax Springer, 2022, pp. 733--751.

\bibitem{rts2022}
M.~Paul, M.~Danelljan, C.~Mayer, and L.~Van~Gool, ``Robust visual tracking by
  segmentation,'' in \emph{European Conference on Computer Vision}.\hskip 1em
  plus 0.5em minus 0.4em\relax Springer, 2022, pp. 571--588.

\bibitem{aiatrack2022}
S.~Gao, C.~Zhou, C.~Ma, X.~Wang, and J.~Yuan, ``Aiatrack: Attention in
  attention for transformer visual tracking,'' in \emph{European Conference on
  Computer Vision}.\hskip 1em plus 0.5em minus 0.4em\relax Springer, 2022, pp.
  146--164.

\bibitem{yang2023foreground}
D.~Yang, J.~He, Y.~Ma, Q.~Yu, and T.~Zhang, ``Foreground-background
  distribution modeling transformer for visual object tracking,'' in
  \emph{Proceedings of the IEEE/CVF International Conference on Computer
  Vision}, 2023, pp. 10\,117--10\,127.

\bibitem{romtrack}
Y.~Cai, J.~Liu, J.~Tang, and G.~Wu, ``Robust object modeling for visual
  tracking,'' in \emph{Proceedings of the IEEE/CVF International Conference on
  Computer Vision}, 2023, pp. 9589--9600.

\bibitem{mixformerv2}
Y.~Cui, T.~Song, G.~Wu, and L.~Wang, ``Mixformerv2: Efficient fully transformer
  tracking,'' \emph{Advances in Neural Information Processing Systems},
  vol.~36, 2024.

\bibitem{rfgm}
X.~Zhou, P.~Guo, L.~Hong, J.~Li, W.~Zhang, W.~Ge, and W.~Zhang, ``Reading
  relevant feature from global representation memory for visual object
  tracking,'' \emph{Advances in Neural Information Processing Systems},
  vol.~36, 2023.

\bibitem{dropmae}
Q.~Wu, T.~Yang, Z.~Liu, B.~Wu, Y.~Shan, and A.~B. Chan, ``Dropmae: Masked
  autoencoders with spatial-attention dropout for tracking tasks,'' in
  \emph{Proceedings of the IEEE/CVF Conference on Computer Vision and Pattern
  Recognition}, 2023, pp. 14\,561--14\,571.

\bibitem{grm}
S.~Gao, C.~Zhou, and J.~Zhang, ``Generalized relation modeling for transformer
  tracking,'' in \emph{Proceedings of the IEEE/CVF Conference on Computer
  Vision and Pattern Recognition}, 2023, pp. 18\,686--18\,695.

\bibitem{artrack}
X.~Wei, Y.~Bai, Y.~Zheng, D.~Shi, and Y.~Gong, ``Autoregressive visual
  tracking,'' in \emph{Proceedings of the IEEE/CVF Conference on Computer
  Vision and Pattern Recognition}, 2023, pp. 9697--9706.

\bibitem{seqtrack}
X.~Chen, H.~Peng, D.~Wang, H.~Lu, and H.~Hu, ``Seqtrack: Sequence to sequence
  learning for visual object tracking,'' in \emph{Proceedings of the IEEE/CVF
  Conference on Computer Vision and Pattern Recognition}, 2023, pp.
  14\,572--14\,581.

\bibitem{videotrack}
F.~Xie, L.~Chu, J.~Li, Y.~Lu, and C.~Ma, ``Videotrack: Learning to track
  objects via video transformer,'' in \emph{Proceedings of the IEEE/CVF
  Conference on Computer Vision and Pattern Recognition}, 2023, pp.
  22\,826--22\,835.

\bibitem{hiptrack}
W.~Cai, Q.~Liu, and Y.~Wang, ``Learning historical status prompt for accurate
  and robust visual tracking,'' \emph{arXiv preprint arXiv:2311.02072}, 2023.

\bibitem{odtrack}
Y.~Zheng, B.~Zhong, Q.~Liang, Z.~Mo, S.~Zhang, and X.~Li, ``Odtrack: Online
  dense temporal token learning for visual tracking,'' \emph{arXiv preprint
  arXiv:2401.01686}, 2024.

\bibitem{evptrack}
L.~Shi, B.~Zhong, Q.~Liang, N.~Li, S.~Zhang, and X.~Li, ``Explicit visual
  prompts for visual object tracking,'' \emph{arXiv preprint arXiv:2401.03142},
  2024.

\bibitem{DSARCF2019}
W.~Feng, R.~Han, Q.~Guo, J.~Zhu, and S.~Wang, ``Dynamic saliency-aware
  regularization for correlation filter-based object tracking,'' \emph{IEEE
  Transactions on Image Processing}, vol.~28, no.~7, pp. 3232--3245, 2019.

\bibitem{LADCF2019}
T.~Xu, Z.-H. Feng, X.-J. Wu, and J.~Kittler, ``Learning adaptive discriminative
  correlation filters via temporal consistency preserving spatial feature
  selection for robust visual object tracking,'' \emph{IEEE Transactions on
  Image Processing}, vol.~28, no.~11, pp. 5596--5609, 2019.

\bibitem{dong2019quadruplet}
X.~Dong, J.~Shen, D.~Wu, K.~Guo, X.~Jin, and F.~Porikli, ``Quadruplet network
  with one-shot learning for fast visual object tracking,'' \emph{IEEE
  Transactions on Image Processing}, vol.~28, no.~7, pp. 3516--3527, 2019.

\bibitem{li2018visual}
C.~Li, L.~Lin, W.~Zuo, J.~Tang, and M.-H. Yang, ``Visual tracking via dynamic
  graph learning,'' \emph{IEEE transactions on pattern analysis and machine
  intelligence}, vol.~41, no.~11, pp. 2770--2782, 2018.

\bibitem{LDES2019}
Y.~Li, J.~Zhu, S.~C. Hoi, W.~Song, Z.~Wang, and H.~Liu, ``Robust estimation of
  similarity transformation for visual object tracking,'' in \emph{Proceedings
  of the AAAI conference on artificial intelligence}, vol.~33, no.~01, 2019,
  pp. 8666--8673.

\bibitem{wang2019spm}
G.~Wang, C.~Luo, Z.~Xiong, and W.~Zeng, ``Spm-tracker: Series-parallel matching
  for real-time visual object tracking,'' in \emph{Proceedings of the IEEE/CVF
  conference on computer vision and pattern recognition}, 2019, pp. 3643--3652.

\bibitem{sun2019roi}
Y.~Sun, C.~Sun, D.~Wang, Y.~He, and H.~Lu, ``Roi pooled correlation filters for
  visual tracking,'' in \emph{Proceedings of the IEEE/CVF conference on
  computer vision and pattern recognition}, 2019, pp. 5783--5791.

\bibitem{siammask2019}
Q.~Wang, L.~Zhang, L.~Bertinetto, W.~Hu, and P.~H. Torr, ``Fast online object
  tracking and segmentation: A unifying approach,'' in \emph{Proceedings of the
  IEEE/CVF conference on Computer Vision and Pattern Recognition}, 2019, pp.
  1328--1338.

\bibitem{TADT}
X.~Li, C.~Ma, B.~Wu, Z.~He, and M.-H. Yang, ``Target-aware deep tracking,'' in
  \emph{Proceedings of the IEEE/CVF conference on computer vision and pattern
  recognition}, 2019, pp. 1369--1378.

\bibitem{C-RPN2019}
H.~Fan and H.~Ling, ``Siamese cascaded region proposal networks for real-time
  visual tracking,'' in \emph{Proceedings of the IEEE/CVF conference on
  computer vision and pattern recognition}, 2019, pp. 7952--7961.

\bibitem{ASRCF2019}
K.~Dai, D.~Wang, H.~Lu, C.~Sun, and J.~Li, ``Visual tracking via adaptive
  spatially-regularized correlation filters,'' in \emph{Proceedings of the
  IEEE/CVF Conference on Computer Vision and Pattern Recognition}, 2019, pp.
  4670--4679.

\bibitem{siamdw2019}
Z.~Zhang and H.~Peng, ``Deeper and wider siamese networks for real-time visual
  tracking,'' in \emph{Proceedings of the IEEE/CVF Conference on Computer
  Vision and Pattern Recognition}, 2019, pp. 4591--4600.

\bibitem{gao2019graph}
J.~Gao, T.~Zhang, and C.~Xu, ``Graph convolutional tracking,'' in
  \emph{Proceedings of the IEEE/CVF Conference on Computer Vision and Pattern
  Recognition}, 2019, pp. 4649--4659.

\bibitem{danelljan2019atom}
M.~Danelljan, G.~Bhat, F.~S. Khan, and M.~Felsberg, ``Atom: Accurate tracking
  by overlap maximization,'' in \emph{Proceedings of the IEEE/CVF Conference on
  Computer Vision and Pattern Recognition}, 2019, pp. 4660--4669.

\bibitem{huang2019bridging}
L.~Huang, X.~Zhao, and K.~Huang, ``Bridging the gap between detection and
  tracking: A unified approach,'' in \emph{Proceedings of the IEEE/CVF
  International Conference on Computer Vision}, 2019, pp. 3999--4009.

\bibitem{mlt2019}
J.~Choi, J.~Kwon, and K.~M. Lee, ``Deep meta learning for real-time
  target-aware visual tracking,'' in \emph{Proceedings of the IEEE/CVF
  international conference on computer vision}, 2019, pp. 911--920.

\bibitem{li2019gradnet}
P.~Li, B.~Chen, W.~Ouyang, D.~Wang, X.~Yang, and H.~Lu, ``Gradnet:
  Gradient-guided network for visual object tracking,'' in \emph{Proceedings of
  the IEEE/CVF International conference on computer vision}, 2019, pp.
  6162--6171.

\bibitem{xu2019joint}
T.~Xu, Z.-H. Feng, X.-J. Wu, and J.~Kittler, ``Joint group feature selection
  and discriminative filter learning for robust visual object tracking,'' in
  \emph{Proceedings of the IEEE/CVF International Conference on Computer
  Vision}, 2019, pp. 7950--7960.

\bibitem{updatenet2019}
L.~Zhang, A.~Gonzalez-Garcia, J.~v.~d. Weijer, M.~Danelljan, and F.~S. Khan,
  ``Learning the model update for siamese trackers,'' in \emph{Proceedings of
  the IEEE/CVF international conference on computer vision}, 2019, pp.
  4010--4019.

\bibitem{splt2019}
B.~Yan, H.~Zhao, D.~Wang, H.~Lu, and X.~Yang, ``'skimming-perusal'tracking: A
  framework for real-time and robust long-term tracking,'' in \emph{Proceedings
  of the IEEE/CVF International Conference on Computer Vision}, 2019, pp.
  2385--2393.

\bibitem{ma2020situp}
H.~Ma, S.~T. Acton, and Z.~Lin, ``Situp: Scale invariant tracking using average
  peak-to-correlation energy,'' \emph{IEEE Transactions on Image Processing},
  vol.~29, pp. 3546--3557, 2020.

\bibitem{CMKCF2020}
B.~Huang, T.~Xu, S.~Jiang, Y.~Chen, and Y.~Bai, ``Robust visual tracking via
  constrained multi-kernel correlation filters,'' \emph{IEEE Transactions on
  Multimedia}, vol.~22, no.~11, pp. 2820--2832, 2020.

\bibitem{fDeepSTRCF2020}
N.~Wang, W.~Zhou, Y.~Song, C.~Ma, and H.~Li, ``Real-time correlation tracking
  via joint model compression and transfer,'' \emph{IEEE Transactions on Image
  Processing}, vol.~29, pp. 6123--6135, 2020.

\bibitem{zhang2020mining}
Y.~Zhang, X.~Gao, Z.~Chen, H.~Zhong, H.~Xie, and C.~Yan, ``Mining
  spatial-temporal similarity for visual tracking,'' \emph{IEEE Transactions on
  Image Processing}, vol.~29, pp. 8107--8119, 2020.

\bibitem{WSCF2020}
R.~Han, W.~Feng, and S.~Wang, ``Fast learning of spatially regularized and
  content aware correlation filter for visual tracking,'' \emph{IEEE
  Transactions on Image Processing}, vol.~29, pp. 7128--7140, 2020.

\bibitem{han2019ensemble}
Y.~Han, P.~Zhang, T.~Zhuo, W.~Huang, Y.~Zha, and Y.~Zhang, ``Ensemble tracking
  based on diverse collaborative framework with multi-cue dynamic fusion,''
  \emph{IEEE Transactions on Multimedia}, vol.~22, no.~10, pp. 2698--2710,
  2019.

\bibitem{lu2020deep}
X.~Lu, C.~Ma, J.~Shen, X.~Yang, I.~Reid, and M.-H. Yang, ``Deep object tracking
  with shrinkage loss,'' \emph{IEEE transactions on pattern analysis and
  machine intelligence}, 2020.

\bibitem{drol2020}
J.~Zhou, P.~Wang, and H.~Sun, ``Discriminative and robust online learning for
  siamese visual tracking,'' in \emph{Proceedings of the AAAI Conference on
  Artificial Intelligence}, vol.~34, no.~07, 2020, pp. 13\,017--13\,024.

\bibitem{siamfc++2020}
Y.~Xu, Z.~Wang, Z.~Li, Y.~Yuan, and G.~Yu, ``Siamfc++: Towards robust and
  accurate visual tracking with target estimation guidelines,'' in
  \emph{Proceedings of the AAAI Conference on Artificial Intelligence},
  vol.~34, no.~07, 2020, pp. 12\,549--12\,556.

\bibitem{globaltrack2020}
L.~Huang, X.~Zhao, and K.~Huang, ``Globaltrack: A simple and strong baseline
  for long-term tracking,'' in \emph{Proceedings of the AAAI Conference on
  Artificial Intelligence}, vol.~34, no.~07, 2020, pp. 11\,037--11\,044.

\bibitem{zhang2020online}
W.~Zhang, R.~Song, Y.~Li \emph{et~al.}, ``Online decision based visual tracking
  via reinforcement learning,'' \emph{Advances in Neural Information Processing
  Systems}, vol.~33, pp. 11\,778--11\,788, 2020.

\bibitem{li2021tlpg}
S.~Li, Z.~Zhang, Z.~Liu, A.~Wang, L.~Qiu, and F.~Du, ``Tlpg-tracker: Joint
  learning of target localization and proposal generation for visual
  tracking,'' in \emph{Proceedings of the Twenty-Ninth International Conference
  on International Joint Conferences on Artificial Intelligence}, 2021, pp.
  708--715.

\bibitem{bhat2020know}
G.~Bhat, M.~Danelljan, L.~V. Gool, and R.~Timofte, ``Know your surroundings:
  Exploiting scene information for object tracking,'' in \emph{European
  Conference on Computer Vision}.\hskip 1em plus 0.5em minus 0.4em\relax
  Springer, 2020, pp. 205--221.

\bibitem{liao2020pg}
B.~Liao, C.~Wang, Y.~Wang, Y.~Wang, and J.~Yin, ``Pg-net: Pixel to global
  matching network for visual tracking,'' in \emph{European Conference on
  Computer Vision}.\hskip 1em plus 0.5em minus 0.4em\relax Springer, 2020, pp.
  429--444.

\bibitem{liu2020object}
Y.~Liu, R.~Li, Y.~Cheng, R.~T. Tan, and X.~Sui, ``Object tracking using
  spatio-temporal networks for future prediction location,'' in \emph{European
  Conference on Computer Vision}.\hskip 1em plus 0.5em minus 0.4em\relax
  Springer, 2020, pp. 1--17.

\bibitem{ocean2020}
Z.~Zhang, H.~Peng, J.~Fu, B.~Li, and W.~Hu, ``Ocean: Object-aware anchor-free
  tracking,'' in \emph{European Conference on Computer Vision}.\hskip 1em plus
  0.5em minus 0.4em\relax Springer, 2020, pp. 771--787.

\bibitem{wang2020tracking}
G.~Wang, C.~Luo, X.~Sun, Z.~Xiong, and W.~Zeng, ``Tracking by instance
  detection: A meta-learning approach,'' in \emph{Proceedings of the IEEE/CVF
  conference on computer vision and pattern recognition}, 2020, pp. 6288--6297.

\bibitem{SiamRCNN2020}
P.~Voigtlaender, J.~Luiten, P.~H. Torr, and B.~Leibe, ``Siam r-cnn: Visual
  tracking by re-detection,'' in \emph{Proceedings of the IEEE/CVF conference
  on computer vision and pattern recognition}, 2020, pp. 6578--6588.

\bibitem{roam2020}
T.~Yang, P.~Xu, R.~Hu, H.~Chai, and A.~B. Chan, ``Roam: Recurrently optimizing
  tracking model,'' in \emph{Proceedings of the IEEE/CVF conference on computer
  vision and pattern recognition}, 2020, pp. 6718--6727.

\bibitem{gao2020recursive}
J.~Gao, W.~Hu, and Y.~Lu, ``Recursive least-squares estimator-aided online
  learning for visual tracking,'' in \emph{Proceedings of the IEEE/CVF
  Conference on Computer Vision and Pattern Recognition}, 2020, pp. 7386--7395.

\bibitem{danelljan2020probabilistic}
M.~Danelljan, L.~V. Gool, and R.~Timofte, ``Probabilistic regression for visual
  tracking,'' in \emph{Proceedings of the IEEE/CVF conference on computer
  vision and pattern recognition}, 2020, pp. 7183--7192.

\bibitem{ltmu2020}
K.~Dai, Y.~Zhang, D.~Wang, J.~Li, H.~Lu, and X.~Yang, ``High-performance
  long-term tracking with meta-updater,'' in \emph{Proceedings of the IEEE/CVF
  Conference on Computer Vision and Pattern Recognition}, 2020, pp. 6298--6307.

\bibitem{SiamAttn2020}
Y.~Yu, Y.~Xiong, W.~Huang, and M.~R. Scott, ``Deformable siamese attention
  networks for visual object tracking,'' in \emph{Proceedings of the IEEE/CVF
  conference on computer vision and pattern recognition}, 2020, pp. 6728--6737.

\bibitem{CGACD2020}
F.~Du, P.~Liu, W.~Zhao, and X.~Tang, ``Correlation-guided attention for corner
  detection based visual tracking,'' in \emph{Proceedings of the IEEE/CVF
  Conference on Computer Vision and Pattern Recognition}, 2020, pp. 6836--6845.

\bibitem{mayer2021learning}
C.~Mayer, M.~Danelljan, D.~P. Paudel, and L.~Van~Gool, ``Learning target
  candidate association to keep track of what not to track,'' in
  \emph{Proceedings of the IEEE/CVF International Conference on Computer
  Vision}, 2021, pp. 13\,444--13\,454.

\bibitem{yu2021high}
B.~Yu, M.~Tang, L.~Zheng, G.~Zhu, J.~Wang, H.~Feng, X.~Feng, and H.~Lu,
  ``High-performance discriminative tracking with transformers,'' in
  \emph{Proceedings of the IEEE/CVF International Conference on Computer
  Vision}, 2021, pp. 9856--9865.

\bibitem{automatch2021}
Z.~Zhang, Y.~Liu, X.~Wang, B.~Li, and W.~Hu, ``Learn to match: Automatic
  matching network design for visual tracking,'' in \emph{Proceedings of the
  IEEE/CVF International Conference on Computer Vision}, 2021, pp.
  13\,339--13\,348.

\bibitem{saot2021}
Z.~Zhou, W.~Pei, X.~Li, H.~Wang, F.~Zheng, and Z.~He, ``Saliency-associated
  object tracking,'' in \emph{Proceedings of the IEEE/CVF International
  Conference on Computer Vision}, 2021, pp. 9866--9875.

\bibitem{stark2021}
B.~Yan, H.~Peng, J.~Fu, D.~Wang, and H.~Lu, ``Learning spatio-temporal
  transformer for visual tracking,'' in \emph{Proceedings of the IEEE/CVF
  International Conference on Computer Vision}, 2021, pp. 10\,448--10\,457.

\bibitem{RRCF2020}
M.~Guan and C.~Wen, ``Adaptive multi-feature reliability re-determinative
  correlation filter for visual tracking,'' \emph{IEEE Transactions on
  Multimedia}, vol.~23, pp. 3841--3852, 2020.

\bibitem{tian2020siamese}
S.~Tian, X.~Liu, M.~Liu, S.~Li, and B.~Yin, ``Siamese tracking network with
  informative enhanced loss,'' \emph{IEEE Transactions on Multimedia}, vol.~23,
  pp. 120--132, 2020.

\bibitem{zhang2020cat}
S.~Zhang, X.~Zhao, and L.~Fang, ``Cat: Corner aided tracking with deep
  regression network,'' \emph{IEEE Transactions on Multimedia}, vol.~23, pp.
  859--870, 2020.

\bibitem{pu2020learning}
S.~Pu, Y.~Song, C.~Ma, H.~Zhang, and M.-H. Yang, ``Learning recurrent memory
  activation networks for visual tracking,'' \emph{IEEE Transactions on Image
  Processing}, vol.~30, pp. 725--738, 2020.

\bibitem{yang2019visual}
T.~Yang and A.~B. Chan, ``Visual tracking via dynamic memory networks,''
  \emph{IEEE transactions on pattern analysis and machine intelligence},
  vol.~43, no.~1, pp. 360--374, 2019.

\bibitem{tan2021nocal}
H.~Tan, X.~Zhang, Z.~Zhang, L.~Lan, W.~Zhang, and Z.~Luo, ``Nocal-siam:
  Refining visual features and response with advanced non-local blocks for
  real-time siamese tracking,'' \emph{IEEE Transactions on Image Processing},
  vol.~30, pp. 2656--2668, 2021.

\bibitem{zhou2021siamcan}
W.~Zhou, L.~Wen, L.~Zhang, D.~Du, T.~Luo, and Y.~Wu, ``Siamcan: Real-time
  visual tracking based on siamese center-aware network,'' \emph{IEEE
  Transactions on Image Processing}, vol.~30, pp. 3597--3609, 2021.

\bibitem{dong2019dynamical}
X.~Dong, J.~Shen, W.~Wang, L.~Shao, H.~Ling, and F.~Porikli, ``Dynamical
  hyperparameter optimization via deep reinforcement learning in tracking,''
  \emph{IEEE transactions on pattern analysis and machine intelligence},
  vol.~43, no.~5, pp. 1515--1529, 2019.

\bibitem{zhang2021toward}
Z.~Zhang, Y.~Liu, B.~Li, W.~Hu, and H.~Peng, ``Toward accurate pixelwise object
  tracking via attention retrieval,'' \emph{IEEE Transactions on Image
  Processing}, vol.~30, pp. 8553--8566, 2021.

\bibitem{wang2021transformer}
N.~Wang, W.~Zhou, J.~Wang, and H.~Li, ``Transformer meets tracker: Exploiting
  temporal context for robust visual tracking,'' in \emph{Proceedings of the
  IEEE/CVF Conference on Computer Vision and Pattern Recognition}, 2021, pp.
  1571--1580.

\bibitem{transT2021}
X.~Chen, B.~Yan, J.~Zhu, D.~Wang, X.~Yang, and H.~Lu, ``Transformer tracking,''
  in \emph{Proceedings of the IEEE/CVF Conference on Computer Vision and
  Pattern Recognition}, 2021, pp. 8126--8135.

\bibitem{dmtrack2021}
Z.~Zhang, B.~Zhong, S.~Zhang, Z.~Tang, X.~Liu, and Z.~Zhang, ``Distractor-aware
  fast tracking via dynamic convolutions and mot philosophy,'' in
  \emph{Proceedings of the IEEE/CVF Conference on Computer Vision and Pattern
  Recognition}, 2021, pp. 1024--1033.

\bibitem{RESiamNets2021}
D.~K. Gupta, D.~Arya, and E.~Gavves, ``Rotation equivariant siamese networks
  for tracking,'' in \emph{Proceedings of the IEEE/CVF Conference on Computer
  Vision and Pattern Recognition}, 2021, pp. 12\,362--12\,371.

\bibitem{ACM2021}
W.~Han, X.~Dong, F.~S. Khan, L.~Shao, and J.~Shen, ``Learning to fuse
  asymmetric feature maps in siamese trackers,'' in \emph{Proceedings of the
  IEEE/CVF Conference on Computer Vision and Pattern Recognition}, 2021, pp.
  16\,570--16\,580.

\bibitem{siamrn2021}
S.~Cheng, B.~Zhong, G.~Li, X.~Liu, Z.~Tang, X.~Li, and J.~Wang, ``Learning to
  filter: Siamese relation network for robust tracking,'' in \emph{Proceedings
  of the IEEE/CVF Conference on Computer Vision and Pattern Recognition}, 2021,
  pp. 4421--4431.

\bibitem{guo2021graph}
D.~Guo, Y.~Shao, Y.~Cui, Z.~Wang, L.~Zhang, and C.~Shen, ``Graph attention
  tracking,'' in \emph{Proceedings of the IEEE/CVF conference on computer
  vision and pattern recognition}, 2021, pp. 9543--9552.

\bibitem{ma2021capsulerrt}
D.~Ma and X.~Wu, ``Capsulerrt: Relationships-aware regression tracking via
  capsules,'' in \emph{Proceedings of the IEEE/CVF Conference on Computer
  Vision and Pattern Recognition}, 2021, pp. 10\,948--10\,957.

\bibitem{yan2021alpha}
B.~Yan, X.~Zhang, D.~Wang, H.~Lu, and X.~Yang, ``Alpha-refine: Boosting
  tracking performance by precise bounding box estimation,'' in
  \emph{Proceedings of the IEEE/CVF Conference on Computer Vision and Pattern
  Recognition}, 2021, pp. 5289--5298.

\bibitem{fan2021cract}
H.~Fan and H.~Ling, ``Cract: Cascaded regression-align-classification for
  robust tracking,'' in \emph{2021 IEEE/RSJ International Conference on
  Intelligent Robots and Systems (IROS)}.\hskip 1em plus 0.5em minus
  0.4em\relax IEEE, 2021, pp. 7013--7020.

\bibitem{zhou2021model}
L.~Zhou, A.~Ledent, Q.~Hu, T.~Liu, J.~Zhang, and M.~Kloft, ``Model uncertainty
  guides visual object tracking,'' in \emph{Proceedings of the AAAI Conference
  on Artificial Intelligence}, vol.~35, no.~4, 2021, pp. 3581--3589.

\bibitem{zhang2021visual}
D.~Zhang, Z.~Zheng, R.~Jia, and M.~Li, ``Visual tracking via hierarchical deep
  reinforcement learning,'' in \emph{Proceedings of the AAAI Conference on
  Artificial Intelligence}, vol.~35, no.~4, 2021, pp. 3315--3323.

\bibitem{mkcf2015}
M.~Tang and J.~Feng, ``Multi-kernel correlation filter for visual tracking,''
  in \emph{Proceedings of the IEEE international conference on computer
  vision}, 2015, pp. 3038--3046.

\bibitem{RAJSSC2015}
M.~Zhang, J.~Xing, J.~Gao, X.~Shi, Q.~Wang, and W.~Hu, ``Joint scale-spatial
  correlation tracking with adaptive rotation estimation,'' in
  \emph{Proceedings of the IEEE international conference on computer vision
  workshops}, 2015, pp. 32--40.

\bibitem{danelljan2015learning}
M.~Danelljan, G.~Hager, F.~Shahbaz~Khan, and M.~Felsberg, ``Learning spatially
  regularized correlation filters for visual tracking,'' in \emph{Proceedings
  of the IEEE international conference on computer vision}, 2015, pp.
  4310--4318.

\bibitem{ma2015hierarchical}
C.~Ma, J.-B. Huang, X.~Yang, and M.-H. Yang, ``Hierarchical convolutional
  features for visual tracking,'' in \emph{Proceedings of the IEEE
  international conference on computer vision}, 2015, pp. 3074--3082.

\bibitem{ma2015long}
C.~Ma, X.~Yang, C.~Zhang, and M.-H. Yang, ``Long-term correlation tracking,''
  in \emph{Proceedings of the IEEE conference on computer vision and pattern
  recognition}, 2015, pp. 5388--5396.

\bibitem{hong2015multi}
Z.~Hong, Z.~Chen, C.~Wang, X.~Mei, D.~Prokhorov, and D.~Tao, ``Multi-store
  tracker (muster): A cognitive psychology inspired approach to object
  tracking,'' in \emph{Proceedings of the IEEE conference on computer vision
  and pattern recognition}, 2015, pp. 749--758.

\bibitem{danelljan2015convolutional}
M.~Danelljan, G.~Hager, F.~Shahbaz~Khan, and M.~Felsberg, ``Convolutional
  features for correlation filter based visual tracking,'' in \emph{Proceedings
  of the IEEE international conference on computer vision workshops}, 2015, pp.
  58--66.

\bibitem{hong2015online}
S.~Hong, T.~You, S.~Kwak, and B.~Han, ``Online tracking by learning
  discriminative saliency map with convolutional neural network,'' in
  \emph{International conference on machine learning}.\hskip 1em plus 0.5em
  minus 0.4em\relax PMLR, 2015, pp. 597--606.

\bibitem{wang2015visual}
L.~Wang, W.~Ouyang, X.~Wang, and H.~Lu, ``Visual tracking with fully
  convolutional networks,'' in \emph{Proceedings of the IEEE international
  conference on computer vision}, 2015, pp. 3119--3127.

\bibitem{chen2017ycnn}
K.~Chen and W.~Tao, ``Once for all: a two-flow convolutional neural network for
  visual tracking,'' \emph{IEEE Transactions on Circuits and Systems for Video
  Technology}, vol.~28, no.~12, pp. 3377--3386, 2017.

\bibitem{bertinetto2016learning}
L.~Bertinetto, J.~F. Henriques, J.~Valmadre, P.~Torr, and A.~Vedaldi,
  ``Learning feed-forward one-shot learners,'' \emph{Advances in neural
  information processing systems}, vol.~29, 2016.

\bibitem{tao2016siamese}
R.~Tao, E.~Gavves, and A.~W. Smeulders, ``Siamese instance search for
  tracking,'' in \emph{Proceedings of the IEEE conference on computer vision
  and pattern recognition}, 2016, pp. 1420--1429.

\bibitem{wang2016stct}
L.~Wang, W.~Ouyang, X.~Wang, and H.~Lu, ``Stct: Sequentially training
  convolutional networks for visual tracking,'' in \emph{Proceedings of the
  IEEE conference on computer vision and pattern recognition}, 2016, pp.
  1373--1381.

\bibitem{held2016learning}
D.~Held, S.~Thrun, and S.~Savarese, ``Learning to track at 100 fps with deep
  regression networks,'' in \emph{European conference on computer
  vision}.\hskip 1em plus 0.5em minus 0.4em\relax Springer, 2016, pp. 749--765.

\bibitem{zhang2016robust}
K.~Zhang, Q.~Liu, Y.~Wu, and M.-H. Yang, ``Robust visual tracking via
  convolutional networks without training,'' \emph{IEEE Transactions on Image
  Processing}, vol.~25, no.~4, pp. 1779--1792, 2016.

\bibitem{bertinetto2016staple}
L.~Bertinetto, J.~Valmadre, S.~Golodetz, O.~Miksik, and P.~H. Torr, ``Staple:
  Complementary learners for real-time tracking,'' in \emph{Proceedings of the
  IEEE conference on computer vision and pattern recognition}, 2016, pp.
  1401--1409.

\bibitem{qi2016hedged}
Y.~Qi, S.~Zhang, L.~Qin, H.~Yao, Q.~Huang, J.~Lim, and M.-H. Yang, ``Hedged
  deep tracking,'' in \emph{Proceedings of the IEEE conference on computer
  vision and pattern recognition}, 2016, pp. 4303--4311.

\bibitem{DMSRDCF2016}
S.~Gladh, M.~Danelljan, F.~S. Khan, and M.~Felsberg, ``Deep motion features for
  visual tracking,'' in \emph{2016 23rd international conference on pattern
  recognition (ICPR)}.\hskip 1em plus 0.5em minus 0.4em\relax IEEE, 2016, pp.
  1243--1248.

\bibitem{SRDCFde2016}
M.~Danelljan, G.~Hager, F.~Shahbaz~Khan, and M.~Felsberg, ``Adaptive
  decontamination of the training set: A unified formulation for discriminative
  visual tracking,'' in \emph{Proceedings of the IEEE conference on computer
  vision and pattern recognition}, 2016, pp. 1430--1438.

\bibitem{C-COT}
M.~Danelljan, A.~Robinson, F.~Shahbaz~Khan, and M.~Felsberg, ``Beyond
  correlation filters: Learning continuous convolution operators for visual
  tracking,'' in \emph{European conference on computer vision}.\hskip 1em plus
  0.5em minus 0.4em\relax Springer, 2016, pp. 472--488.

\bibitem{fan2017sanet}
H.~Fan and H.~Ling, ``Sanet: Structure-aware network for visual tracking,'' in
  \emph{Proceedings of the IEEE conference on computer vision and pattern
  recognition workshops}, 2017, pp. 42--49.

\bibitem{yun2017action}
S.~Yun, J.~Choi, Y.~Yoo, K.~Yun, and J.~Young~Choi, ``Action-decision networks
  for visual tracking with deep reinforcement learning,'' in \emph{Proceedings
  of the IEEE conference on computer vision and pattern recognition}, 2017, pp.
  2711--2720.

\bibitem{mueller2017context}
M.~Mueller, N.~Smith, and B.~Ghanem, ``Context-aware correlation filter
  tracking,'' in \emph{Proceedings of the IEEE conference on computer vision
  and pattern recognition}, 2017, pp. 1396--1404.

\bibitem{johnander2017dcco}
J.~Johnander, M.~Danelljan, F.~S. Khan, and M.~Felsberg, ``Dcco: Towards
  deformable continuous convolution operators for visual tracking,'' in
  \emph{International Conference on Computer Analysis of Images and
  Patterns}.\hskip 1em plus 0.5em minus 0.4em\relax Springer, 2017, pp. 55--67.

\bibitem{kiani2017learning}
H.~Kiani~Galoogahi, A.~Fagg, and S.~Lucey, ``Learning background-aware
  correlation filters for visual tracking,'' in \emph{Proceedings of the IEEE
  international conference on computer vision}, 2017, pp. 1135--1143.

\bibitem{DCF}
A.~Lukezic, T.~Vojir, L.~ˇCehovin~Zajc, J.~Matas, and M.~Kristan,
  ``Discriminative correlation filter with channel and spatial reliability,''
  in \emph{Proceedings of the IEEE conference on computer vision and pattern
  recognition}, 2017, pp. 6309--6318.

\bibitem{MCPF2017}
T.~Zhang, C.~Xu, and M.-H. Yang, ``Multi-task correlation particle filter for
  robust object tracking,'' in \emph{Proceedings of the IEEE conference on
  computer vision and pattern recognition}, 2017, pp. 4335--4343.

\bibitem{CFNet2017}
J.~Valmadre, L.~Bertinetto, J.~Henriques, A.~Vedaldi, and P.~H. Torr,
  ``End-to-end representation learning for correlation filter based tracking,''
  in \emph{Proceedings of the IEEE conference on computer vision and pattern
  recognition}, 2017, pp. 2805--2813.

\bibitem{wang2017large}
M.~Wang, Y.~Liu, and Z.~Huang, ``Large margin object tracking with circulant
  feature maps,'' in \emph{Proceedings of the IEEE conference on computer
  vision and pattern recognition}, 2017, pp. 4021--4029.

\bibitem{zhang2017robust}
L.~Zhang, J.~Varadarajan, P.~Nagaratnam~Suganthan, N.~Ahuja, and P.~Moulin,
  ``Robust visual tracking using oblique random forests,'' in \emph{Proceedings
  of the IEEE conference on computer vision and pattern recognition}, 2017, pp.
  5589--5598.

\bibitem{danelljan2017eco}
M.~Danelljan, G.~Bhat, F.~Shahbaz~Khan, and M.~Felsberg, ``Eco: Efficient
  convolution operators for tracking,'' in \emph{Proceedings of the IEEE
  conference on computer vision and pattern recognition}, 2017, pp. 6638--6646.

\bibitem{teng2017robust}
Z.~Teng, J.~Xing, Q.~Wang, C.~Lang, S.~Feng, and Y.~Jin, ``Robust object
  tracking based on temporal and spatial deep networks,'' in \emph{Proceedings
  of the IEEE International Conference on Computer Vision}, 2017, pp.
  1144--1153.

\bibitem{CFCF2018}
E.~Gundogdu and A.~A. Alatan, ``Good features to correlate for visual
  tracking,'' \emph{IEEE Transactions on Image Processing}, vol.~27, no.~5, pp.
  2526--2540, 2018.

\bibitem{yang2017recurrent}
T.~Yang and A.~B. Chan, ``Recurrent filter learning for visual tracking,'' in
  \emph{Proceedings of the IEEE International Conference on Computer Vision
  Workshops}, 2017, pp. 2010--2019.

\bibitem{chi2017dual}
Z.~Chi, H.~Li, H.~Lu, and M.-H. Yang, ``Dual deep network for visual
  tracking,'' \emph{IEEE Transactions on Image Processing}, vol.~26, no.~4, pp.
  2005--2015, 2017.

\bibitem{zhang2018robust}
T.~Zhang, C.~Xu, and M.-H. Yang, ``Robust structural sparse tracking,''
  \emph{IEEE transactions on pattern analysis and machine intelligence},
  vol.~41, no.~2, pp. 473--486, 2018.

\bibitem{lu2014online}
Y.~Lu, T.~Wu, and S.~Chun~Zhu, ``Online object tracking, learning and parsing
  with and-or graphs,'' in \emph{Proceedings of the IEEE Conference on Computer
  Vision and Pattern Recognition}, 2014, pp. 3462--3469.

\bibitem{li2017integrating}
F.~Li, Y.~Yao, P.~Li, D.~Zhang, W.~Zuo, and M.-H. Yang, ``Integrating boundary
  and center correlation filters for visual tracking with aspect ratio
  variation,'' in \emph{Proceedings of the IEEE International Conference on
  Computer Vision Workshops}, 2017, pp. 2001--2009.

\bibitem{he2017correlation}
Z.~He, Y.~Fan, J.~Zhuang, Y.~Dong, and H.~Bai, ``Correlation filters with
  weighted convolution responses,'' in \emph{Proceedings of the IEEE
  International Conference on Computer Vision Workshops}, 2017, pp. 1992--2000.

\bibitem{Dsiam2017}
Q.~Guo, W.~Feng, C.~Zhou, R.~Huang, L.~Wan, and S.~Wang, ``Learning dynamic
  siamese network for visual object tracking,'' in \emph{Proceedings of the
  IEEE international conference on computer vision}, 2017, pp. 1763--1771.

\bibitem{huang2017learning}
C.~Huang, S.~Lucey, and D.~Ramanan, ``Learning policies for adaptive tracking
  with deep feature cascades,'' in \emph{Proceedings of the IEEE international
  conference on computer vision}, 2017, pp. 105--114.

\bibitem{CREST2017}
Y.~Song, C.~Ma, L.~Gong, J.~Zhang, R.~W. Lau, and M.-H. Yang, ``Crest:
  Convolutional residual learning for visual tracking,'' in \emph{Proceedings
  of the IEEE international conference on computer vision}, 2017, pp.
  2555--2564.

\bibitem{FlowTrack}
Z.~Zhu, W.~Wu, W.~Zou, and J.~Yan, ``End-to-end flow correlation tracking with
  spatial-temporal attention,'' in \emph{Proceedings of the IEEE conference on
  computer vision and pattern recognition}, 2018, pp. 548--557.

\bibitem{song2018vital}
Y.~Song, C.~Ma, X.~Wu, L.~Gong, L.~Bao, W.~Zuo, C.~Shen, R.~W. Lau, and M.-H.
  Yang, ``Vital: Visual tracking via adversarial learning,'' in
  \emph{Proceedings of the IEEE conference on computer vision and pattern
  recognition}, 2018, pp. 8990--8999.

\bibitem{gao2018p2t}
J.~Gao, T.~Zhang, X.~Yang, and C.~Xu, ``P2t: Part-to-target tracking via deep
  regression learning,'' \emph{IEEE Transactions on Image Processing}, vol.~27,
  no.~6, pp. 3074--3086, 2018.

\bibitem{du2017iterative}
D.~Du, L.~Wen, H.~Qi, Q.~Huang, Q.~Tian, and S.~Lyu, ``Iterative graph seeking
  for object tracking,'' \emph{IEEE Transactions on Image Processing}, vol.~27,
  no.~4, pp. 1809--1821, 2017.

\bibitem{li2019learning}
B.~Li, W.~Xie, W.~Zeng, and W.~Liu, ``Learning to update for object tracking
  with recurrent meta-learner,'' \emph{IEEE Transactions on Image Processing},
  vol.~28, no.~7, pp. 3624--3635, 2019.

\bibitem{li2018visual-tracking}
Z.~Li, J.~Zhang, K.~Zhang, and Z.~Li, ``Visual tracking with weighted adaptive
  local sparse appearance model via spatio-temporal context learning,''
  \emph{IEEE Transactions on Image Processing}, vol.~27, no.~9, pp. 4478--4489,
  2018.

\bibitem{liu2019deformable}
W.~Liu, Y.~Song, D.~Chen, S.~He, Y.~Yu, T.~Yan, G.~P. Hancke, and R.~W. Lau,
  ``Deformable object tracking with gated fusion,'' \emph{IEEE Transactions on
  Image Processing}, vol.~28, no.~8, pp. 3766--3777, 2019.

\bibitem{RASNet}
Q.~Wang, Z.~Teng, J.~Xing, J.~Gao, W.~Hu, and S.~Maybank, ``Learning
  attentions: residual attentional siamese network for high performance online
  visual tracking,'' in \emph{Proceedings of the IEEE conference on computer
  vision and pattern recognition}, 2018, pp. 4854--4863.

\bibitem{2018tm3}
F.~Liu, C.~Gong, X.~Huang, T.~Zhou, J.~Yang, and D.~Tao, ``Robust visual
  tracking revisited: From correlation filter to template matching,''
  \emph{IEEE Transactions on Image Processing}, vol.~27, no.~6, pp. 2777--2790,
  2018.

\bibitem{CSRDCF2018}
Y.~Zhou, J.~Han, F.~Yang, K.~Zhang, and R.~Hong, ``Efficient correlation
  tracking via center-biased spatial regularization,'' \emph{IEEE Transactions
  on Image Processing}, vol.~27, no.~12, pp. 6159--6173, 2018.

\bibitem{sun2018correlation}
C.~Sun, D.~Wang, H.~Lu, and M.-H. Yang, ``Correlation tracking via joint
  discrimination and reliability learning,'' in \emph{Proceedings of the IEEE
  conference on computer vision and pattern recognition}, 2018, pp. 489--497.

\bibitem{dong2018hyperparameter}
X.~Dong, J.~Shen, W.~Wang, Y.~Liu, L.~Shao, and F.~Porikli, ``Hyperparameter
  optimization for tracking with continuous deep q-learning,'' in
  \emph{Proceedings of the IEEE conference on computer vision and pattern
  recognition}, 2018, pp. 518--527.

\bibitem{SA-Siam}
A.~He, C.~Luo, X.~Tian, and W.~Zeng, ``A twofold siamese network for real-time
  object tracking,'' in \emph{Proceedings of the IEEE conference on computer
  vision and pattern recognition}, 2018, pp. 4834--4843.

\bibitem{LSART2018}
C.~Sun, D.~Wang, H.~Lu, and M.-H. Yang, ``Learning spatial-aware regressions
  for visual tracking,'' in \emph{Proceedings of the IEEE Conference on
  Computer Vision and Pattern Recognition}, 2018, pp. 8962--8970.

\bibitem{bhat2018unveiling}
G.~Bhat, J.~Johnander, M.~Danelljan, F.~S. Khan, and M.~Felsberg, ``Unveiling
  the power of deep tracking,'' in \emph{Proceedings of the European Conference
  on Computer Vision (ECCV)}, 2018, pp. 483--498.

\bibitem{li2018learning}
F.~Li, C.~Tian, W.~Zuo, L.~Zhang, and M.-H. Yang, ``Learning spatial-temporal
  regularized correlation filters for visual tracking,'' in \emph{Proceedings
  of the IEEE conference on computer vision and pattern recognition}, 2018, pp.
  4904--4913.

\bibitem{wang2018multi}
N.~Wang, W.~Zhou, Q.~Tian, R.~Hong, M.~Wang, and H.~Li, ``Multi-cue correlation
  filters for robust visual tracking,'' in \emph{Proceedings of the IEEE
  conference on computer vision and pattern recognition}, 2018, pp. 4844--4853.

\bibitem{wang2018sint++}
X.~Wang, C.~Li, B.~Luo, and J.~Tang, ``Sint++: Robust visual tracking via
  adversarial positive instance generation,'' in \emph{Proceedings of the IEEE
  conference on computer vision and pattern recognition}, 2018, pp. 4864--4873.

\bibitem{pu2018deep}
S.~Pu, Y.~Song, C.~Ma, H.~Zhang, and M.-H. Yang, ``Deep attentive tracking via
  reciprocative learning,'' \emph{Advances in neural information processing
  systems}, vol.~31, 2018.

\bibitem{yang2018learning}
T.~Yang and A.~B. Chan, ``Learning dynamic memory networks for object
  tracking,'' in \emph{Proceedings of the European conference on computer
  vision (ECCV)}, 2018, pp. 152--167.

\bibitem{DaSiamRPN2018}
Z.~Zhu, Q.~Wang, B.~Li, W.~Wu, J.~Yan, and W.~Hu, ``Distractor-aware siamese
  networks for visual object tracking,'' in \emph{Proceedings of the European
  conference on computer vision (ECCV)}, 2018, pp. 101--117.

\bibitem{ren2018deep}
L.~Ren, X.~Yuan, J.~Lu, M.~Yang, and J.~Zhou, ``Deep reinforcement learning
  with iterative shift for visual tracking,'' in \emph{Proceedings of the
  European conference on computer vision (ECCV)}, 2018, pp. 684--700.

\bibitem{lu2018deep}
X.~Lu, C.~Ma, B.~Ni, X.~Yang, I.~Reid, and M.-H. Yang, ``Deep regression
  tracking with shrinkage loss,'' in \emph{Proceedings of the European
  conference on computer vision (ECCV)}, 2018, pp. 353--369.

\bibitem{chen2018real}
B.~Chen, D.~Wang, P.~Li, S.~Wang, and H.~Lu,
  ``Real-time'actor-critic'tracking,'' in \emph{Proceedings of the European
  conference on computer vision (ECCV)}, 2018, pp. 318--334.

\bibitem{jung2018real}
I.~Jung, J.~Son, M.~Baek, and B.~Han, ``Real-time mdnet,'' in \emph{Proceedings
  of the European conference on computer vision (ECCV)}, 2018, pp. 83--98.

\bibitem{StructSiam}
Y.~Zhang, L.~Wang, J.~Qi, D.~Wang, M.~Feng, and H.~Lu, ``Structured siamese
  network for real-time visual tracking,'' in \emph{Proceedings of the European
  conference on computer vision (ECCV)}, 2018, pp. 351--366.

\bibitem{zhang2018visual}
M.~Zhang, Q.~Wang, J.~Xing, J.~Gao, P.~Peng, W.~Hu, and S.~Maybank, ``Visual
  tracking via spatially aligned correlation filters network,'' in
  \emph{Proceedings of the European conference on computer vision (ECCV)},
  2018, pp. 469--485.

\bibitem{SiamFC-tri}
X.~Dong and J.~Shen, ``Triplet loss in siamese network for object tracking,''
  in \emph{Proceedings of the European conference on computer vision (ECCV)},
  2018, pp. 459--474.

\bibitem{zhang2012real}
K.~Zhang, L.~Zhang, and M.-H. Yang, ``Real-time compressive tracking,'' in
  \emph{European conference on computer vision}.\hskip 1em plus 0.5em minus
  0.4em\relax Springer, 2012, pp. 864--877.

\bibitem{wang2016robust}
M.~Wang, Y.~Liu, and R.~Xiong, ``Robust object tracking with a hierarchical
  ensemble framework,'' in \emph{2016 IEEE/RSJ International Conference on
  Intelligent Robots and Systems (IROS)}.\hskip 1em plus 0.5em minus
  0.4em\relax IEEE, 2016, pp. 438--445.

\bibitem{danelljan2014adaptive}
M.~Danelljan, F.~Shahbaz~Khan, M.~Felsberg, and J.~Van~de Weijer, ``Adaptive
  color attributes for real-time visual tracking,'' in \emph{Proceedings of the
  IEEE conference on computer vision and pattern recognition}, 2014, pp.
  1090--1097.

\bibitem{cn2014}
------, ``Adaptive color attributes for real-time visual tracking,'' in
  \emph{Proceedings of the IEEE conference on computer vision and pattern
  recognition}, 2014, pp. 1090--1097.

\bibitem{babenko2010robust}
B.~Babenko, M.-H. Yang, and S.~Belongie, ``Robust object tracking with online
  multiple instance learning,'' \emph{IEEE transactions on pattern analysis and
  machine intelligence}, vol.~33, no.~8, pp. 1619--1632, 2010.

\bibitem{danelljan2014accurate}
M.~Danelljan, G.~H{\"a}ger, F.~Khan, and M.~Felsberg, ``Accurate scale
  estimation for robust visual tracking,'' in \emph{British Machine Vision
  Conference, Nottingham, September 1-5, 2014}.\hskip 1em plus 0.5em minus
  0.4em\relax Bmva Press, 2014.

\bibitem{wang2013learning}
N.~Wang and D.-Y. Yeung, ``Learning a deep compact image representation for
  visual tracking,'' \emph{Advances in neural information processing systems},
  vol.~26, 2013.

\bibitem{bolme2010visual}
D.~S. Bolme, J.~R. Beveridge, B.~A. Draper, and Y.~M. Lui, ``Visual object
  tracking using adaptive correlation filters,'' in \emph{2010 IEEE computer
  society conference on computer vision and pattern recognition}.\hskip 1em
  plus 0.5em minus 0.4em\relax IEEE, 2010, pp. 2544--2550.

\bibitem{henriques2012exploiting}
J.~F. Henriques, R.~Caseiro, P.~Martins, and J.~Batista, ``Exploiting the
  circulant structure of tracking-by-detection with kernels,'' in
  \emph{European conference on computer vision}.\hskip 1em plus 0.5em minus
  0.4em\relax Springer, 2012, pp. 702--715.

\bibitem{zhang2014fast}
K.~Zhang, L.~Zhang, Q.~Liu, D.~Zhang, and M.-H. Yang, ``Fast visual tracking
  via dense spatio-temporal context learning,'' in \emph{European conference on
  computer vision}.\hskip 1em plus 0.5em minus 0.4em\relax Springer, 2014, pp.
  127--141.

\bibitem{samf2014}
Y.~Li and J.~Zhu, ``A scale adaptive kernel correlation filter tracker with
  feature integration,'' in \emph{European conference on computer
  vision}.\hskip 1em plus 0.5em minus 0.4em\relax Springer, 2014, pp. 254--265.

\bibitem{li2015reliable}
Y.~Li, J.~Zhu, and S.~C. Hoi, ``Reliable patch trackers: Robust visual tracking
  by exploiting reliable patches,'' in \emph{Proceedings of the IEEE Conference
  on Computer Vision and Pattern Recognition}, 2015, pp. 353--361.

\bibitem{kiani2015correlation}
H.~Kiani~Galoogahi, T.~Sim, and S.~Lucey, ``Correlation filters with limited
  boundaries,'' in \emph{Proceedings of the IEEE conference on computer vision
  and pattern recognition}, 2015, pp. 4630--4638.

\bibitem{bibi2016target}
A.~Bibi, M.~Mueller, and B.~Ghanem, ``Target response adaptation for
  correlation filter tracking,'' in \emph{European conference on computer
  vision}.\hskip 1em plus 0.5em minus 0.4em\relax Springer, 2016, pp. 419--433.

\bibitem{sui2016real}
Y.~Sui, Z.~Zhang, G.~Wang, Y.~Tang, and L.~Zhang, ``Real-time visual tracking:
  Promoting the robustness of correlation filter learning,'' in \emph{European
  conference on computer vision}.\hskip 1em plus 0.5em minus 0.4em\relax
  Springer, 2016, pp. 662--678.

\bibitem{liu2016structural}
S.~Liu, T.~Zhang, X.~Cao, and C.~Xu, ``Structural correlation filter for robust
  visual tracking,'' in \emph{Proceedings of the IEEE Conference on Computer
  Vision and Pattern Recognition}, 2016, pp. 4312--4320.

\bibitem{fan2017parallel}
H.~Fan and H.~Ling, ``Parallel tracking and verifying: A framework for
  real-time and high accuracy visual tracking,'' in \emph{Proceedings of the
  IEEE international conference on computer vision}, 2017, pp. 5486--5494.

\bibitem{sun2018learning}
C.~Sun, D.~Wang, H.~Lu, and M.-H. Yang, ``Learning spatial-aware regressions
  for visual tracking,'' in \emph{Proceedings of the IEEE Conference on
  Computer Vision and Pattern Recognition}, 2018, pp. 8962--8970.

\bibitem{danelljan2016beyond}
M.~Danelljan, A.~Robinson, F.~Shahbaz~Khan, and M.~Felsberg, ``Beyond
  correlation filters: Learning continuous convolution operators for visual
  tracking,'' in \emph{European conference on computer vision}.\hskip 1em plus
  0.5em minus 0.4em\relax Springer, 2016, pp. 472--488.

\bibitem{ECO}
M.~Danelljan, G.~Bhat, F.~Shahbaz~Khan, and M.~Felsberg, ``Eco: Efficient
  convolution operators for tracking,'' in \emph{Proceedings of the IEEE
  conference on computer vision and pattern recognition}, 2017, pp. 6638--6646.

\bibitem{dai2020high}
K.~Dai, Y.~Zhang, D.~Wang, J.~Li, H.~Lu, and X.~Yang, ``High-performance
  long-term tracking with meta-updater,'' in \emph{Proceedings of the IEEE/CVF
  Conference on Computer Vision and Pattern Recognition}, 2020, pp. 6298--6307.

\bibitem{bromley1993signature}
J.~Bromley, I.~Guyon, Y.~LeCun, E.~S{\"a}ckinger, and R.~Shah, ``Signature
  verification using a" siamese" time delay neural network,'' \emph{Advances in
  neural information processing systems}, vol.~6, 1993.

\bibitem{schroff2015facenet}
F.~Schroff, D.~Kalenichenko, and J.~Philbin, ``Facenet: A unified embedding for
  face recognition and clustering,'' in \emph{Proceedings of the IEEE
  conference on computer vision and pattern recognition}, 2015, pp. 815--823.

\bibitem{parkhi2015deep}
O.~M. Parkhi, A.~Vedaldi, and A.~Zisserman, ``Deep face recognition,'' 2015.

\bibitem{chen2021exploring}
X.~Chen and K.~He, ``Exploring simple siamese representation learning,'' in
  \emph{Proceedings of the IEEE/CVF Conference on Computer Vision and Pattern
  Recognition}, 2021, pp. 15\,750--15\,758.

\bibitem{oh2019video}
S.~W. Oh, J.-Y. Lee, N.~Xu, and S.~J. Kim, ``Video object segmentation using
  space-time memory networks,'' in \emph{Proceedings of the IEEE/CVF
  International Conference on Computer Vision}, 2019, pp. 9226--9235.

\bibitem{wang2022delving}
M.~Wang, J.~Mei, L.~Liu, G.~Tian, Y.~Liu, and Z.~Pan, ``Delving deeper into
  mask utilization in video object segmentation,'' \emph{IEEE Transactions on
  Image Processing}, 2022.

\bibitem{ren2015faster}
S.~Ren, K.~He, R.~Girshick, and J.~Sun, ``Faster r-cnn: Towards real-time
  object detection with region proposal networks,'' \emph{Advances in neural
  information processing systems}, vol.~28, 2015.

\bibitem{cract2021}
H.~Fan and H.~Ling, ``Cract: Cascaded regression-align-classification for
  robust tracking,'' in \emph{2021 IEEE/RSJ International Conference on
  Intelligent Robots and Systems (IROS)}.\hskip 1em plus 0.5em minus
  0.4em\relax IEEE, 2021, pp. 7013--7020.

\bibitem{DTT2021}
B.~Yu, M.~Tang, L.~Zheng, G.~Zhu, J.~Wang, H.~Feng, X.~Feng, and H.~Lu,
  ``High-performance discriminative tracking with transformers,'' in
  \emph{Proceedings of the IEEE/CVF International Conference on Computer
  Vision}, 2021, pp. 9856--9865.

\bibitem{fan2010human}
J.~Fan, W.~Xu, Y.~Wu, and Y.~Gong, ``Human tracking using convolutional neural
  networks,'' \emph{IEEE transactions on Neural Networks}, vol.~21, no.~10, pp.
  1610--1623, 2010.

\bibitem{li2014robust}
H.~Li, Y.~Li, and F.~Porikli, ``Robust online visual tracking with a single
  convolutional neural network,'' in \emph{Asian Conference on Computer
  Vision}.\hskip 1em plus 0.5em minus 0.4em\relax Springer, 2014, pp. 194--209.

\bibitem{wang2015transferring}
N.~Wang, S.~Li, A.~Gupta, and D.-Y. Yeung, ``Transferring rich feature
  hierarchies for robust visual tracking,'' \emph{arXiv preprint
  arXiv:1501.04587}, 2015.

\bibitem{tian2019fcos}
Z.~Tian, C.~Shen, H.~Chen, and T.~He, ``Fcos: Fully convolutional one-stage
  object detection,'' in \emph{Proceedings of the IEEE/CVF international
  conference on computer vision}, 2019, pp. 9627--9636.

\bibitem{lin2017focal}
T.-Y. Lin, P.~Goyal, R.~Girshick, K.~He, and P.~Doll{\'a}r, ``Focal loss for
  dense object detection,'' in \emph{Proceedings of the IEEE international
  conference on computer vision}, 2017, pp. 2980--2988.

\bibitem{nam2016learning}
H.~Nam and B.~Han, ``Learning multi-domain convolutional neural networks for
  visual tracking,'' in \emph{Proceedings of the IEEE conference on computer
  vision and pattern recognition}, 2016, pp. 4293--4302.

\bibitem{park2018meta}
E.~Park and A.~C. Berg, ``Meta-tracker: Fast and robust online adaptation for
  visual object trackers,'' in \emph{Proceedings of the European Conference on
  Computer Vision (ECCV)}, 2018, pp. 569--585.

\bibitem{zheng2021rethinking}
S.~Zheng, J.~Lu, H.~Zhao, X.~Zhu, Z.~Luo, Y.~Wang, Y.~Fu, J.~Feng, T.~Xiang,
  P.~H. Torr \emph{et~al.}, ``Rethinking semantic segmentation from a
  sequence-to-sequence perspective with transformers,'' in \emph{Proceedings of
  the IEEE/CVF conference on computer vision and pattern recognition}, 2021,
  pp. 6881--6890.

\bibitem{li2021mail}
Z.~Li, M.~Wang, J.~Mei, and Y.~Liu, ``Mail: A unified mask-image-language
  trimodal network for referring image segmentation,'' \emph{arXiv preprint
  arXiv:2111.10747}, 2021.

\bibitem{chen2021pix2seq}
T.~Chen, S.~Saxena, L.~Li, D.~J. Fleet, and G.~Hinton, ``Pix2seq: A language
  modeling framework for object detection,'' \emph{arXiv preprint
  arXiv:2109.10852}, 2021.

\bibitem{wang2021actionclip}
M.~Wang, J.~Xing, J.~Mei, Y.~Liu, and Y.~Jiang, ``Actionclip: Adapting
  language-image pretrained models for video action recognition,'' \emph{IEEE
  Transactions on Neural Networks and Learning Systems}, 2023.

\bibitem{fan2021multiscale}
H.~Fan, B.~Xiong, K.~Mangalam, Y.~Li, Z.~Yan, J.~Malik, and C.~Feichtenhofer,
  ``Multiscale vision transformers,'' in \emph{Proceedings of the IEEE/CVF
  International Conference on Computer Vision}, 2021, pp. 6824--6835.

\bibitem{kim2021vilt}
W.~Kim, B.~Son, and I.~Kim, ``Vilt: Vision-and-language transformer without
  convolution or region supervision,'' in \emph{International Conference on
  Machine Learning}.\hskip 1em plus 0.5em minus 0.4em\relax PMLR, 2021, pp.
  5583--5594.

\bibitem{MILTracker}
X.~Shi, W.~Hu, Y.~Cheng, G.~Chen, J.~Ji, and H.~Ling, ``Infrared target
  tracking using multiple instance learning with adaptive motion prediction and
  spatially template weighting,'' in \emph{Sensors and Systems for Space
  Applications VI}, vol. 8739.\hskip 1em plus 0.5em minus 0.4em\relax SPIE,
  2013, pp. 351--357.

\bibitem{RLRST}
R.~Liu and Y.~Lu, ``Infrared target tracking in multiple feature pseudo-color
  image with kernel density estimation,'' \emph{Infrared Physics \&
  Technology}, vol.~55, no.~6, pp. 505--512, 2012.

\bibitem{ABCD}
A.~Berg, J.~Ahlberg, and M.~Felsberg, ``Channel coded distribution field
  tracking for thermal infrared imagery,'' in \emph{Proceedings of the IEEE
  Conference on Computer Vision and Pattern Recognition Workshops}, 2016, pp.
  9--17.

\bibitem{MFPCIS}
Y.~He, M.~Li, J.~Zhang, and J.~Yao, ``Infrared target tracking based on robust
  low-rank sparse learning,'' \emph{IEEE Geoscience and Remote Sensing
  Letters}, vol.~13, no.~2, pp. 232--236, 2015.

\bibitem{DSLT}
X.~Yu, Q.~Yu, Y.~Shang, and H.~Zhang, ``Dense structural learning for infrared
  object tracking at 200+ frames per second,'' \emph{Pattern Recognition
  Letters}, vol. 100, pp. 152--159, 2017.

\bibitem{P-CODIFF}
H.~S. Demir and O.~F. Adil, ``Part-based co-difference object tracking
  algorithm for infrared videos,'' in \emph{2018 25th IEEE International
  Conference on Image Processing (ICIP)}.\hskip 1em plus 0.5em minus
  0.4em\relax IEEE, 2018, pp. 3723--3727.

\bibitem{MaskSR}
M.~Li, L.~Peng, Y.~Chen, S.~Huang, F.~Qin, and Z.~Peng, ``Mask sparse
  representation based on semantic features for thermal infrared target
  tracking,'' \emph{Remote Sensing}, vol.~11, no.~17, p. 1967, 2019.

\bibitem{MCFTS}
Q.~Liu, X.~Lu, Z.~He, C.~Zhang, and W.-S. Chen, ``Deep convolutional neural
  networks for thermal infrared object tracking,'' \emph{Knowledge-Based
  Systems}, vol. 134, pp. 189--198, 2017.

\bibitem{ECO-stir}
L.~Zhang, A.~Gonzalez-Garcia, J.~Van De~Weijer, M.~Danelljan, and F.~S. Khan,
  ``Synthetic data generation for end-to-end thermal infrared tracking,''
  \emph{IEEE Transactions on Image Processing}, vol.~28, no.~4, pp. 1837--1850,
  2018.

\bibitem{DGM}
L.~Jin, J.~Cheng, and C.~Zhang, ``Infrared pedestrian tracking with graph
  memory features,'' \emph{IEEE Signal Processing Letters}, vol.~28, pp.
  1933--1937, 2021.

\bibitem{qi2024exploring}
M.~Qi, Q.~Wang, S.~Zhuang, K.~Zhang, K.~Li, Y.~Liu, and Y.~Yang, ``Exploring
  reliable infrared object tracking with spatio-temporal fusion transformer,''
  \emph{Knowledge-Based Systems}, vol. 284, p. 111234, 2024.

\bibitem{li2024two}
S.~Li, G.~Fu, X.~Yang, X.~Cao, S.~Niu, and Z.~Meng, ``Two-stage spatio-temporal
  feature correlation network for infrared ground target tracking,'' \emph{IEEE
  Transactions on Geoscience and Remote Sensing}, 2024.

\bibitem{HSSNet}
X.~Li, Q.~Liu, N.~Fan, Z.~He, and H.~Wang, ``Hierarchical spatial-aware siamese
  network for thermal infrared object tracking,'' \emph{Knowledge-Based
  Systems}, vol. 166, pp. 71--81, 2019.

\bibitem{MLSSNet}
Q.~Liu, X.~Li, Z.~He, N.~Fan, D.~Yuan, and H.~Wang, ``Learning deep multi-level
  similarity for thermal infrared object tracking,'' \emph{IEEE Transactions on
  Multimedia}, vol.~23, pp. 2114--2126, 2020.

\bibitem{MMNet}
Q.~Liu, X.~Li, Z.~He, N.~Fan, D.~Yuan, W.~Liu, and Y.~Liang, ``Multi-task
  driven feature models for thermal infrared tracking,'' in \emph{Proceedings
  of the AAAI Conference on Artificial Intelligence}, vol.~34, no.~07, 2020,
  pp. 11\,604--11\,611.

\bibitem{AFF}
Y.~Wang, J.~Ma, J.~Lv, and Z.~Zhao, ``Thermal infrared object tracking based on
  adaptive feature fusion,'' in \emph{2021 11th International Conference on
  Information Technology in Medicine and Education (ITME)}.\hskip 1em plus
  0.5em minus 0.4em\relax IEEE, 2021, pp. 71--75.

\bibitem{HCFA-Siam}
Y.~Xu, M.~Wan, Q.~Chen, W.~Qian, K.~Ren, and G.~Gu, ``Hierarchical convolution
  fusion-based adaptive siamese network for infrared target tracking,''
  \emph{IEEE Transactions on Instrumentation and Measurement}, vol.~70, pp.
  1--12, 2021.

\bibitem{S&A}
T.~Yao, J.~Hu, B.~Zhang, Y.~Gao, P.~Li, and Q.~Hu, ``Scale and appearance
  variation enhanced siamese network for thermal infrared target tracking,''
  \emph{Infrared Physics \& Technology}, vol. 117, p. 103825, 2021.

\bibitem{STAMT}
D.~Yuan, X.~Shu, Q.~Liu, and Z.~He, ``Structural target-aware model for thermal
  infrared tracking,'' \emph{Neurocomputing}, vol. 491, pp. 44--56, 2022.

\bibitem{WCF}
Y.-J. He, M.~Li, J.~Zhang, and J.-P. Yao, ``Infrared target tracking via
  weighted correlation filter,'' \emph{Infrared Physics \& Technology},
  vol.~73, pp. 103--114, 2015.

\bibitem{TBOOST}
E.~Gundogdu, H.~Ozkan, H.~Seckin~Demir, H.~Ergezer, E.~Akagunduz, and
  S.~Kubilay~Pakin, ``Comparison of infrared and visible imagery for object
  tracking: Toward trackers with superior ir performance,'' in
  \emph{Proceedings of the IEEE Conference on Computer Vision and Pattern
  Recognition Workshops}, 2015, pp. 1--9.

\bibitem{DSST-TIR}
E.~Gundogdu, A.~Koc, B.~Solmaz, R.~I. Hammoud, and A.~Aydin~Alatan,
  ``Evaluation of feature channels for correlation-filter-based visual object
  tracking in infrared spectrum,'' in \emph{Proceedings of the IEEE Conference
  on Computer Vision and Pattern recognition Workshops}, 2016, pp. 24--32.

\bibitem{LMSCO}
P.~Gao, Y.~Ma, K.~Song, C.~Li, F.~Wang, and L.~Xiao, ``Large margin structured
  convolution operator for thermal infrared object tracking,'' in \emph{2018
  24th International Conference on Pattern Recognition (ICPR)}.\hskip 1em plus
  0.5em minus 0.4em\relax IEEE, 2018, pp. 2380--2385.

\bibitem{RCCF-TIR}
T.~Yu, B.~Mo, F.~Liu, H.~Qi, and Y.~Liu, ``Robust thermal infrared object
  tracking with continuous correlation filters and adaptive feature fusion,''
  \emph{Infrared Physics \& Technology}, vol.~98, pp. 69--81, 2019.

\bibitem{CMD}
J.~Sun, L.~Zhang, Y.~Zha, A.~Gonzalez-Garcia, P.~Zhang, W.~Huang, and Y.~Zhang,
  ``Unsupervised cross-modal distillation for thermal infrared tracking,'' in
  \emph{Proceedings of the 29th ACM International Conference on Multimedia},
  2021, pp. 2262--2270.

\bibitem{VOT-2015}
M.~Felsberg, A.~Berg, G.~Hager, J.~Ahlberg, M.~Kristan, J.~Matas, A.~Leonardis,
  L.~Cehovin, G.~Fernandez, T.~Vojir \emph{et~al.}, ``The thermal infrared
  visual object tracking vot-tir2015 challenge results,'' in \emph{Proceedings
  of the ieee international conference on computer vision workshops}, 2015, pp.
  76--88.

\bibitem{PTB-TIR}
Q.~Liu, Z.~He, X.~Li, and Y.~Zheng, ``Ptb-tir: A thermal infrared pedestrian
  tracking benchmark,'' \emph{IEEE Transactions on Multimedia}, vol.~22, no.~3,
  pp. 666--675, 2019.

\bibitem{P+SiamRPN}
L.~Zheng, S.~Zhao, Y.~Zhang, and L.~Yu, ``Thermal infrared pedestrian tracking
  using joint siamese network and exemplar prediction model,'' \emph{Pattern
  Recognition Letters}, vol. 140, pp. 66--72, 2020.

\bibitem{li2022infrared}
Y.~C. Li and S.~Yang, ``Infrared small object tracking based on att-siam
  network,'' \emph{IEEE Access}, 2022.

\bibitem{SC3D}
S.~Giancola, J.~Zarzar, and B.~Ghanem, ``Leveraging shape completion for 3d
  siamese tracking,'' in \emph{Proceedings of the IEEE/CVF Conference on
  Computer Vision and Pattern Recognition}, 2019, pp. 1359--1368.

\bibitem{bev}
J.~Zarzar, S.~Giancola, and B.~Ghanem, ``Efficient bird eye view proposals for
  3d siamese tracking,'' \emph{arXiv preprint arXiv:1903.10168}, 2019.

\bibitem{mlvsnet}
Z.~Wang, Q.~Xie, Y.-K. Lai, J.~Wu, K.~Long, and J.~Wang, ``Mlvsnet: Multi-level
  voting siamese network for 3d visual tracking,'' in \emph{Proceedings of the
  IEEE/CVF International Conference on Computer Vision}, 2021, pp. 3101--3110.

\bibitem{lttr}
Y.~Cui, Z.~Fang, J.~Shan, Z.~Gu, and S.~Zhou, ``3d object tracking with
  transformer,'' \emph{arXiv preprint arXiv:2110.14921}, 2021.

\bibitem{ptt}
J.~Shan, S.~Zhou, Z.~Fang, and Y.~Cui, ``Ptt: Point-track-transformer module
  for 3d single object tracking in point clouds,'' in \emph{2021 IEEE/RSJ
  International Conference on Intelligent Robots and Systems (IROS)}.\hskip 1em
  plus 0.5em minus 0.4em\relax IEEE, 2021, pp. 1310--1316.

\bibitem{pointsiamrcnn}
H.~Zou, C.~Zhang, Y.~Liu, W.~Li, F.~Wen, and H.~Zhang, ``Pointsiamrcnn:
  Target-aware voxel-based siamese tracker for point clouds,'' in \emph{2021
  IEEE/RSJ International Conference on Intelligent Robots and Systems
  (IROS)}.\hskip 1em plus 0.5em minus 0.4em\relax IEEE, 2021, pp. 7029--7035.

\bibitem{oleksiienko2022vpit}
I.~Oleksiienko, P.~Nousi, N.~Passalis, A.~Tefas, and A.~Iosifidis, ``Vpit:
  Real-time embedded single object 3d tracking using voxel pseudo images,''
  \emph{arXiv preprint arXiv:2206.02619}, 2022.

\bibitem{park2022gpt}
M.~Park, H.~Seong, W.~Jang, and E.~Kim, ``Graph-based point tracker for 3d
  object tracking in point clouds,'' 2022.

\bibitem{zhou2022pttr}
C.~Zhou, Z.~Luo, Y.~Luo, T.~Liu, L.~Pan, Z.~Cai, H.~Zhao, and S.~Lu, ``Pttr:
  Relational 3d point cloud object tracking with transformer,'' in
  \emph{Proceedings of the IEEE/CVF Conference on Computer Vision and Pattern
  Recognition}, 2022, pp. 8531--8540.

\bibitem{pttr++}
Z.~Luo, C.~Zhou, L.~Pan, G.~Zhang, T.~Liu, Y.~Luo, H.~Zhao, Z.~Liu, and S.~Lu,
  ``Exploring point-bev fusion for 3d point cloud object tracking with
  transformer,'' \emph{arXiv preprint arXiv:2208.05216}, 2022.

\bibitem{stnet}
L.~Hui, L.~Wang, L.~Tang, K.~Lan, J.~Xie, and J.~Yang, ``3d siamese transformer
  network for single object tracking on point clouds,'' \emph{arXiv preprint
  arXiv:2207.11995}, 2022.

\bibitem{xia2023lightweight}
Y.~Xia, Q.~Wu, W.~Li, A.~B. Chan, and U.~Stilla, ``A lightweight and
  detector-free 3d single object tracker on point clouds,'' \emph{IEEE
  Transactions on Intelligent Transportation Systems}, 2023.

\bibitem{gltt}
J.~Nie, Z.~He, Y.~Yang, M.~Gao, and J.~Zhang, ``Glt-t: Global-local transformer
  voting for 3d single object tracking in point clouds,'' in \emph{Proceedings
  of the AAAI Conference on Artificial Intelligence}, vol.~37, no.~2, 2023, pp.
  1957--1965.

\bibitem{nie2023osp2b}
J.~Nie, Z.~He, Y.~Yang, Z.~Bao, M.~Gao, and J.~Zhang, ``Osp2b: One-stage
  point-to-box network for 3d siamese tracking,'' \emph{arXiv preprint
  arXiv:2304.11584}, 2023.

\bibitem{xu2023cxtrack}
T.-X. Xu, Y.-C. Guo, Y.-K. Lai, and S.-H. Zhang, ``Cxtrack: Improving 3d point
  cloud tracking with contextual information,'' in \emph{Proceedings of the
  IEEE/CVF Conference on Computer Vision and Pattern Recognition}, 2023, pp.
  1084--1093.

\bibitem{cropnet}
M.~Wang, T.~Ma, X.~Zuo, J.~Lv, and Y.~Liu, ``Correlation pyramid network for 3d
  single object tracking,'' in \emph{Proceedings of the IEEE/CVF Conference on
  Computer Vision and Pattern Recognition}, 2023, pp. 3215--3224.

\bibitem{ma2023synchronize}
T.~Ma, M.~Wang, J.~Xiao, H.~Wu, and Y.~Liu, ``Synchronize feature extracting
  and matching: A single branch framework for 3d object tracking,'' in
  \emph{Proceedings of the IEEE/CVF International Conference on Computer
  Vision}, 2023, pp. 9953--9963.

\bibitem{scvtrack}
J.~Zhang, Z.~Zhou, G.~Lu, J.~Tian, and W.~Pei, ``Robust 3d tracking with
  quality-aware shape completion,'' \emph{arXiv preprint arXiv:2312.10608},
  2023.

\bibitem{yang2023bevtrack}
Y.~Yang, Y.~Deng, J.~Zhang, J.~Nie, and Z.-J. Zha, ``Bevtrack: A simple and
  strong baseline for 3d single object tracking in bird's-eye view,''
  \emph{arXiv e-prints}, pp. arXiv--2309, 2023.

\bibitem{liu2023m3sot}
J.~Liu, Y.~Wu, M.~Gong, Q.~Miao, W.~Ma, and C.~Qin, ``M3sot: Multi-frame,
  multi-field, multi-space 3d single object tracking,'' \emph{arXiv preprint
  arXiv:2312.06117}, 2023.

\bibitem{streamtrack}
Z.~Luo, G.~Zhang, C.~Zhou, Z.~Wu, Q.~Tao, L.~Lu, and S.~Lu, ``Modeling
  continuous motion for 3d point cloud object tracking,'' \emph{arXiv preprint
  arXiv:2303.07605}, 2023.

\bibitem{mtmtrack}
Z.~Li, Y.~Lin, Y.~Cui, S.~Li, and Z.~Fang, ``Motion-to-matching: A mixed
  paradigm for 3d single object tracking,'' \emph{IEEE Robotics and Automation
  Letters}, 2023.

\bibitem{cutrack}
J.~Nie, Z.~He, X.~Lv, X.~Zhou, D.-K. Chae, and F.~Xie, ``Towards category
  unification of 3d single object tracking on point clouds,'' \emph{arXiv
  preprint arXiv:2401.11204}, 2024.

\bibitem{votenet}
C.~R. Qi, O.~Litany, K.~He, and L.~J. Guibas, ``Deep hough voting for 3d object
  detection in point clouds,'' in \emph{proceedings of the IEEE/CVF
  International Conference on Computer Vision}, 2019, pp. 9277--9286.

\bibitem{woo2018cbam}
S.~Woo, J.~Park, J.-Y. Lee, and I.~S. Kweon, ``Cbam: Convolutional block
  attention module,'' in \emph{Proceedings of the European conference on
  computer vision (ECCV)}, 2018, pp. 3--19.

\bibitem{dosovitskiy2020image}
A.~Dosovitskiy, L.~Beyer, A.~Kolesnikov, D.~Weissenborn, X.~Zhai,
  T.~Unterthiner, M.~Dehghani, M.~Minderer, G.~Heigold, S.~Gelly \emph{et~al.},
  ``An image is worth 16x16 words: Transformers for image recognition at
  scale,'' \emph{arXiv preprint arXiv:2010.11929}, 2020.

\bibitem{deit}
H.~Touvron, M.~Cord, M.~Douze, F.~Massa, A.~Sablayrolles, and H.~J{\'e}gou,
  ``Training data-efficient image transformers \& distillation through
  attention,'' in \emph{International Conference on Machine Learning}.\hskip
  1em plus 0.5em minus 0.4em\relax PMLR, 2021, pp. 10\,347--10\,357.

\bibitem{zhao2021point}
H.~Zhao, L.~Jiang, J.~Jia, P.~H. Torr, and V.~Koltun, ``Point transformer,'' in
  \emph{Proceedings of the IEEE/CVF International Conference on Computer
  Vision}, 2021, pp. 16\,259--16\,268.

\bibitem{guo2021pct}
M.-H. Guo, J.-X. Cai, Z.-N. Liu, T.-J. Mu, R.~R. Martin, and S.-M. Hu, ``Pct:
  Point cloud transformer,'' \emph{Computational Visual Media}, vol.~7, no.~2,
  pp. 187--199, 2021.

\bibitem{feng2023multi}
S.~Feng, P.~Liang, J.~Gao, and E.~Cheng, ``Multi-correlation siamese
  transformer network with dense connection for 3d single object tracking,''
  \emph{IEEE Robotics and Automation Letters}, 2023.

\bibitem{nuscenes}
H.~Caesar, V.~Bankiti, A.~H. Lang, S.~Vora, V.~E. Liong, Q.~Xu, A.~Krishnan,
  Y.~Pan, G.~Baldan, and O.~Beijbom, ``nuscenes: A multimodal dataset for
  autonomous driving,'' in \emph{Proceedings of the IEEE/CVF conference on
  computer vision and pattern recognition}, 2020, pp. 11\,621--11\,631.

\bibitem{KITTI2012}
A.~Geiger, P.~Lenz, and R.~Urtasun, ``Are we ready for autonomous driving? the
  kitti vision benchmark suite,'' in \emph{2012 IEEE conference on computer
  vision and pattern recognition}.\hskip 1em plus 0.5em minus 0.4em\relax IEEE,
  2012, pp. 3354--3361.

\bibitem{amct}
G.~M. Garc{\'\i}a, D.~A. Klein, J.~St{\"u}ckler, S.~Frintrop, and A.~B.
  Cremers, ``Adaptive multi-cue 3d tracking of arbitrary objects,'' in
  \emph{Joint DAGM (German Association for Pattern Recognition) and OAGM
  Symposium}.\hskip 1em plus 0.5em minus 0.4em\relax Springer, 2012, pp.
  357--366.

\bibitem{ptb}
S.~Song and J.~Xiao, ``Tracking revisited using rgbd camera: Unified benchmark
  and baselines,'' in \emph{Proceedings of the IEEE international conference on
  computer vision}, 2013, pp. 233--240.

\bibitem{mcbt}
Q.~Wang, J.~Fang, and Y.~Yuan, ``Multi-cue based tracking,''
  \emph{Neurocomputing}, vol. 131, pp. 227--236, 2014.

\bibitem{dskcf}
M.~Camplani, S.~L. Hannuna, M.~Mirmehdi, D.~Damen, A.~Paiement, L.~Tao, and
  T.~Burghardt, ``Real-time rgb-d tracking with depth scaling kernelised
  correlation filters and occlusion handling.'' in \emph{BMVC}, vol.~3, 2015.

\bibitem{ol3dc}
B.~Zhong, Y.~Shen, Y.~Chen, W.~Xie, Z.~Cui, H.~Zhang, D.~Chen, T.~Wang, X.~Liu,
  S.~Peng \emph{et~al.}, ``Online learning 3d context for robust visual
  tracking,'' \emph{Neurocomputing}, vol. 151, pp. 710--718, 2015.

\bibitem{dskcf-shape}
S.~Hannuna, M.~Camplani, J.~Hall, M.~Mirmehdi, D.~Damen, T.~Burghardt,
  A.~Paiement, and L.~Tao, ``Ds-kcf: a real-time tracker for rgb-d data,''
  \emph{Journal of Real-Time Image Processing}, vol.~16, no.~5, pp. 1439--1458,
  2019.

\bibitem{3dt}
A.~Bibi, T.~Zhang, and B.~Ghanem, ``3d part-based sparse tracker with automatic
  synchronization and registration,'' in \emph{Proceedings of the IEEE
  Conference on Computer Vision and Pattern Recognition}, 2016, pp. 1439--1448.

\bibitem{oapf}
K.~Meshgi, S.-i. Maeda, S.~Oba, H.~Skibbe, Y.-z. Li, and S.~Ishii, ``An
  occlusion-aware particle filter tracker to handle complex and persistent
  occlusions,'' \emph{Computer Vision and Image Understanding}, vol. 150, pp.
  81--94, 2016.

\bibitem{dls}
N.~An, X.-G. Zhao, and Z.-G. Hou, ``Online rgb-d tracking via
  detection-learning-segmentation,'' in \emph{2016 23rd International
  Conference on Pattern Recognition (ICPR)}.\hskip 1em plus 0.5em minus
  0.4em\relax IEEE, 2016, pp. 1231--1236.

\bibitem{stc}
J.~Xiao, R.~Stolkin, Y.~Gao, and A.~Leonardis, ``Robust fusion of color and
  depth data for rgb-d target tracking using adaptive range-invariant depth
  models and spatio-temporal consistency constraints,'' \emph{IEEE transactions
  on cybernetics}, vol.~48, no.~8, pp. 2485--2499, 2017.

\bibitem{ca3dms}
Y.~Liu, X.-Y. Jing, J.~Nie, H.~Gao, J.~Liu, and G.-P. Jiang, ``Context-aware
  three-dimensional mean-shift with occlusion handling for robust object
  tracking in rgb-d videos,'' \emph{IEEE Transactions on Multimedia}, vol.~21,
  no.~3, pp. 664--677, 2018.

\bibitem{otr}
U.~Kart, A.~Lukezic, M.~Kristan, J.-K. Kamarainen, and J.~Matas, ``Object
  tracking by reconstruction with view-specific discriminative correlation
  filters,'' in \emph{Proceedings of the IEEE/CVF Conference on Computer Vision
  and Pattern Recognition}, 2019, pp. 1339--1348.

\bibitem{dal}
Y.~Qian, S.~Yan, A.~Luke{\v{z}}i{\v{c}}, M.~Kristan, J.-K.
  K{\"a}m{\"a}r{\"a}inen, and J.~Matas, ``Dal: A deep depth-aware long-term
  tracker,'' in \emph{2020 25th International Conference on Pattern Recognition
  (ICPR)}.\hskip 1em plus 0.5em minus 0.4em\relax IEEE, 2021, pp. 7825--7832.

\bibitem{tsdm}
P.~Zhao, Q.~Liu, W.~Wang, and Q.~Guo, ``Tsdm: Tracking by siamrpn++ with a
  depth-refiner and a mask-generator,'' in \emph{2020 25th International
  Conference on Pattern Recognition (ICPR)}.\hskip 1em plus 0.5em minus
  0.4em\relax IEEE, 2021, pp. 670--676.

\bibitem{depthtrack}
S.~Yan, J.~Yang, J.~K{\"a}pyl{\"a}, F.~Zheng, A.~Leonardis, and J.-K.
  K{\"a}m{\"a}r{\"a}inen, ``Depthtrack: Unveiling the power of rgbd tracking,''
  in \emph{Proceedings of the IEEE/CVF International Conference on Computer
  Vision}, 2021, pp. 10\,725--10\,733.

\bibitem{zhu2023vipt}
J.~Zhu, S.~Lai, X.~Chen, D.~Wang, and H.~Lu, ``Visual prompt multi-modal
  tracking,'' in \emph{Proceedings of the IEEE/CVF Conference on Computer
  Vision and Pattern Recognition}, 2023, pp. 9516--9526.

\bibitem{hmad}
B.~Xu, Y.~Xu, R.~Hou, J.~Bei, T.~Ren, and G.~Wu, ``Rgb-d tracking via
  hierarchical modality aggregation and distribution network,'' in
  \emph{Proceedings of the 5th ACM International Conference on Multimedia in
  Asia}, 2023, pp. 1--7.

\bibitem{vot}
M.~Kristan, J.~Matas, A.~Leonardis, M.~Felsberg, R.~Pflugfelder, J.-K.
  Kamarainen, L.~ˇCehovin~Zajc, O.~Drbohlav, A.~Lukezic, A.~Berg
  \emph{et~al.}, ``The seventh visual object tracking vot2019 challenge
  results,'' in \emph{Proceedings of the IEEE/CVF International Conference on
  Computer Vision Workshops}, 2019, pp. 0--0.

\bibitem{svm}
O.~Chapelle, ``Training a support vector machine in the primal,'' \emph{Neural
  computation}, vol.~19, no.~5, pp. 1155--1178, 2007.

\bibitem{SCCF}
Y.~Wang, C.~Li, and J.~Tang, ``Learning soft-consistent correlation filters for
  rgb-t object tracking,'' in \emph{Chinese Conference on Pattern Recognition
  and Computer Vision (PRCV)}.\hskip 1em plus 0.5em minus 0.4em\relax Springer,
  2018, pp. 295--306.

\bibitem{LGMG}
C.~Li, C.~Zhu, J.~Zhang, B.~Luo, X.~Wu, and J.~Tang, ``Learning local-global
  multi-graph descriptors for rgb-t object tracking,'' \emph{IEEE Transactions
  on Circuits and Systems for Video Technology}, vol.~29, no.~10, pp.
  2913--2926, 2018.

\bibitem{cross-modal}
C.~Li, C.~Zhu, Y.~Huang, J.~Tang, and L.~Wang, ``Cross-modal ranking with soft
  consistency and noisy labels for robust rgb-t tracking,'' in
  \emph{Proceedings of the European Conference on Computer Vision (ECCV)},
  2018, pp. 808--823.

\bibitem{FANet}
Y.~Zhu, C.~Li, J.~Tang, and B.~Luo, ``Quality-aware feature aggregation network
  for robust rgbt tracking,'' \emph{IEEE Transactions on Intelligent Vehicles},
  vol.~6, no.~1, pp. 121--130, 2020.

\bibitem{FTSNet}
C.~Li, X.~Wu, N.~Zhao, X.~Cao, and J.~Tang, ``Fusing two-stream convolutional
  neural networks for rgb-t object tracking,'' \emph{Neurocomputing}, vol. 281,
  pp. 78--85, 2018.

\bibitem{Densefuse}
H.~Li and X.-J. Wu, ``Densefuse: A fusion approach to infrared and visible
  images,'' \emph{IEEE Transactions on Image Processing}, vol.~28, no.~5, pp.
  2614--2623, 2018.

\bibitem{MDNet+RGBT}
A.~Lu, C.~Li, Y.~Yan, J.~Tang, and B.~Luo, ``Rgbt tracking via multi-adapter
  network with hierarchical divergence loss,'' \emph{IEEE Transactions on Image
  Processing}, vol.~30, pp. 5613--5625, 2021.

\bibitem{CSCF}
Y.~Wang, C.~Li, J.~Tang, and D.~Sun, ``Learning collaborative sparse
  correlation filter for real-time multispectral object tracking,'' in
  \emph{International Conference on Brain Inspired Cognitive Systems}.\hskip
  1em plus 0.5em minus 0.4em\relax Springer, 2018, pp. 462--472.

\bibitem{TODA}
------, ``Learning collaborative sparse correlation filter for real-time
  multispectral object tracking,'' in \emph{International Conference on Brain
  Inspired Cognitive Systems}.\hskip 1em plus 0.5em minus 0.4em\relax Springer,
  2018, pp. 462--472.

\bibitem{Fast}
S.~Zhai, P.~Shao, X.~Liang, and X.~Wang, ``Fast rgb-t tracking via cross-modal
  correlation filters,'' \emph{Neurocomputing}, vol. 334, pp. 172--181, 2019.

\bibitem{SiamFT}
X.~Zhang, P.~Ye, S.~Peng, J.~Liu, K.~Gong, and G.~Xiao, ``Siamft: An
  rgb-infrared fusion tracking method via fully convolutional siamese
  networks,'' \emph{IEEE Access}, vol.~7, pp. 122\,122--122\,133, 2019.

\bibitem{MANet}
C.~Long~Li, A.~Lu, A.~Hua~Zheng, Z.~Tu, and J.~Tang, ``Multi-adapter rgbt
  tracking,'' in \emph{Proceedings of the IEEE/CVF International Conference on
  Computer Vision Workshops}, 2019, pp. 0--0.

\bibitem{DAPNet}
Y.~Zhu, C.~Li, B.~Luo, J.~Tang, and X.~Wang, ``Dense feature aggregation and
  pruning for rgbt tracking,'' in \emph{Proceedings of the 27th ACM
  International Conference on Multimedia}, 2019, pp. 465--472.

\bibitem{CAT}
C.~Li, L.~Liu, A.~Lu, Q.~Ji, and J.~Tang, ``Challenge-aware rgbt tracking,'' in
  \emph{European Conference on Computer Vision}.\hskip 1em plus 0.5em minus
  0.4em\relax Springer, 2020, pp. 222--237.

\bibitem{MaCNet}
H.~Zhang, L.~Zhang, L.~Zhuo, and J.~Zhang, ``Object tracking in rgb-t videos
  using modal-aware attention network and competitive learning,''
  \emph{Sensors}, vol.~20, no.~2, p. 393, 2020.

\bibitem{DMCNet}
A.~Lu, C.~Qian, C.~Li, J.~Tang, and L.~Wang, ``Duality-gated mutual condition
  network for rgbt tracking,'' \emph{IEEE Transactions on Neural Networks and
  Learning Systems}, 2022.

\bibitem{CMPP}
C.~Wang, C.~Xu, Z.~Cui, L.~Zhou, T.~Zhang, X.~Zhang, and J.~Yang, ``Cross-modal
  pattern-propagation for rgb-t tracking,'' in \emph{Proceedings of the
  IEEE/CVF Conference on Computer Vision and Pattern Recognition}, 2020, pp.
  7064--7073.

\bibitem{MFGNet}
X.~Wang, X.~Shu, S.~Zhang, B.~Jiang, Y.~Wang, Y.~Tian, and F.~Wu, ``Mfgnet:
  Dynamic modality-aware filter generation for rgb-t tracking,'' \emph{arXiv
  preprint arXiv:2107.10433}, 2021.

\bibitem{DFNet}
J.~Peng, H.~Zhao, and Z.~Hu, ``Dynamic fusion network for rgbt tracking,''
  \emph{arXiv preprint arXiv:2109.07662}, 2021.

\bibitem{SiamCDA}
T.~Zhang, X.~Liu, Q.~Zhang, and J.~Han, ``Siamcda: Complementarity-and
  distractor-aware rgb-t tracking based on siamese network,'' \emph{IEEE
  Transactions on Circuits and Systems for Video Technology}, vol.~32, no.~3,
  pp. 1403--1417, 2021.

\bibitem{SiamIVFN}
P.~Jingchao, Z.~Haitao, H.~Zhengwei, Z.~Yi, and W.~Bofan, ``Siamese infrared
  and visible light fusion network for rgb-t tracking,'' \emph{arXiv preprint
  arXiv:2103.07302}, 2021.

\bibitem{SiamFC+RGBT}
Y.~Kuai, D.~Li, and Q.~Qian, ``Learning a twofold siamese network for rgb-t
  object tracking,'' \emph{Journal of Circuits, Systems and Computers},
  vol.~30, no.~05, p. 2150089, 2021.

\bibitem{JMMAC}
P.~Zhang, J.~Zhao, C.~Bo, D.~Wang, H.~Lu, and X.~Yang, ``Jointly modeling
  motion and appearance cues for robust rgb-t tracking,'' \emph{IEEE
  Transactions on Image Processing}, vol.~30, pp. 3335--3347, 2021.

\bibitem{APFNet}
Y.~Xiao, M.~Yang, C.~Li, L.~Liu, and J.~Tang, ``Attribute-based progressive
  fusion network for rgbt tracking,'' 2022.

\bibitem{lu2022duality}
A.~Lu, C.~Qian, C.~Li, J.~Tang, and L.~Wang, ``Duality-gated mutual condition
  network for rgbt tracking,'' \emph{IEEE Transactions on Neural Networks and
  Learning Systems}, 2022.

\bibitem{protrack}
J.~Yang, Z.~Li, F.~Zheng, A.~Leonardis, and J.~Song, ``Prompting for
  multi-modal tracking,'' in \emph{Proceedings of the 30th ACM International
  Conference on Multimedia}, 2022, pp. 3492--3500.

\bibitem{cao2023bi}
B.~Cao, J.~Guo, P.~Zhu, and Q.~Hu, ``Bi-directional adapter for multi-modal
  tracking,'' \emph{arXiv preprint arXiv:2312.10611}, 2023.

\bibitem{mplt}
Y.~Luo, X.~Guo, H.~Feng, and L.~Ao, ``Rgb-t tracking via multi-modal mutual
  prompt learning,'' \emph{arXiv preprint arXiv:2308.16386}, 2023.

\bibitem{DiMP-RGBT}
L.~Zhang, M.~Danelljan, A.~Gonzalez-Garcia, J.~van~de Weijer, and
  F.~Shahbaz~Khan, ``Multi-modal fusion for end-to-end rgb-t tracking,'' in
  \emph{Proceedings of the IEEE/CVF International Conference on Computer Vision
  Workshops}, 2019, pp. 0--0.

\bibitem{cocoop}
K.~Zhou, J.~Yang, C.~C. Loy, and Z.~Liu, ``Conditional prompt learning for
  vision-language models,'' in \emph{CVPR}, 2022.

\bibitem{maple}
M.~U. Khattak, H.~Rasheed, M.~Maaz, S.~Khan, and F.~S. Khan, ``Maple:
  Multi-modal prompt learning,'' in \emph{Proceedings of the IEEE/CVF
  Conference on Computer Vision and Pattern Recognition}, 2023, pp.
  19\,113--19\,122.

\bibitem{m2clip}
M.~Wang, J.~Xing, B.~Jiang, J.~Chen, J.~Mei, X.~Zuo, G.~Dai, J.~Wang, and
  Y.~Liu, ``M2-clip: A multimodal, multi-task adapting framework for video
  action recognition,'' \emph{arXiv preprint arXiv:2401.11649}, 2024.

\bibitem{hou2024sdstrack}
X.~Hou, J.~Xing, Y.~Qian, Y.~Guo, S.~Xin, J.~Chen, K.~Tang, M.~Wang, Z.~Jiang,
  L.~Liu \emph{et~al.}, ``Sdstrack: Self-distillation symmetric adapter
  learning for multi-modal visual object tracking,'' in \emph{Proceedings of
  the IEEE/CVF Conference on Computer Vision and Pattern Recognition}, 2024,
  pp. 26\,551--26\,561.

\bibitem{wang2021facilitating}
L.~Wang, L.~Hui, and J.~Xie, ``Facilitating 3d object tracking in point clouds
  with image semantics and geometry,'' in \emph{Chinese Conference on Pattern
  Recognition and Computer Vision (PRCV)}.\hskip 1em plus 0.5em minus
  0.4em\relax Springer, 2021, pp. 589--601.

\bibitem{lan-li}
Z.~Li, R.~Tao, E.~Gavves, C.~G. Snoek, and A.~W. Smeulders, ``Tracking by
  natural language specification,'' in \emph{Proceedings of the IEEE Conference
  on Computer Vision and Pattern Recognition}, 2017, pp. 6495--6503.

\bibitem{lan-wang1}
X.~Wang, C.~Li, R.~Yang, T.~Zhang, J.~Tang, and B.~Luo, ``Describe and attend
  to track: Learning natural language guided structural representation and
  visual attention for object tracking,'' \emph{arXiv preprint
  arXiv:1811.10014}, 2018.

\bibitem{lan-feng1}
Q.~Feng, V.~Ablavsky, Q.~Bai, and S.~Sclaroff, ``Robust visual object tracking
  with natural language region proposal network,'' 2019.

\bibitem{lan-feng2}
Q.~Feng, V.~Ablavsky, Q.~Bai, G.~Li, and S.~Sclaroff, ``Real-time visual object
  tracking with natural language description,'' in \emph{Proceedings of the
  IEEE/CVF Winter Conference on Applications of Computer Vision}, 2020, pp.
  700--709.

\bibitem{lan-wang2}
X.~Wang, X.~Shu, Z.~Zhang, B.~Jiang, Y.~Wang, Y.~Tian, and F.~Wu, ``Towards
  more flexible and accurate object tracking with natural language: Algorithms
  and benchmark,'' in \emph{Proceedings of the IEEE/CVF Conference on Computer
  Vision and Pattern Recognition}, 2021, pp. 13\,763--13\,773.

\bibitem{li2022cross}
Y.~Li, J.~Yu, Z.~Cai, and Y.~Pan, ``Cross-modal target retrieval for tracking
  by natural language,'' in \emph{Proceedings of the IEEE/CVF Conference on
  Computer Vision and Pattern Recognition}, 2022, pp. 4931--4940.

\bibitem{all_in_one}
C.~Zhang, X.~Sun, Y.~Yang, L.~Liu, Q.~Liu, X.~Zhou, and Y.~Wang, ``All in one:
  Exploring unified vision-language tracking with multi-modal alignment,'' in
  \emph{Proceedings of the 31st ACM International Conference on Multimedia},
  2023, pp. 5552--5561.

\bibitem{ovlm}
H.~Zhang, J.~Wang, J.~Zhang, T.~Zhang, and B.~Zhong, ``One-stream
  vision-language memory network for object tracking,'' \emph{IEEE Transactions
  on Multimedia}, 2023.

\bibitem{jointgroundtrack}
L.~Zhou, Z.~Zhou, K.~Mao, and Z.~He, ``Joint visual grounding and tracking with
  natural language specification,'' in \emph{Proceedings of the IEEE/CVF
  Conference on Computer Vision and Pattern Recognition}, 2023, pp.
  23\,151--23\,160.

\bibitem{li2023citetracker}
X.~Li, Y.~Huang, Z.~He, Y.~Wang, H.~Lu, and M.-H. Yang, ``Citetracker:
  Correlating image and text for visual tracking,'' in \emph{Proceedings of the
  IEEE/CVF International Conference on Computer Vision}, 2023, pp. 9974--9983.

\bibitem{mmtrack}
Y.~Zheng, B.~Zhong, Q.~Liang, G.~Li, R.~Ji, and X.~Li, ``Towards unified token
  learning for vision-language tracking,'' \emph{IEEE Transactions on Circuits
  and Systems for Video Technology}, 2023.

\bibitem{lu2023naturalmixer}
Q.~Lu, G.~Yuan, C.~Li, H.~Zhu, and X.~Qin, ``Natural language guided attention
  mixer for object tracking,'' in \emph{2023 4th International Conference on
  Information Science, Parallel and Distributed Systems (ISPDS)}.\hskip 1em
  plus 0.5em minus 0.4em\relax IEEE, 2023, pp. 160--164.

\bibitem{ma2024unifying}
Y.~Ma, Y.~Tang, W.~Yang, T.~Zhang, J.~Zhang, and M.~Kang, ``Unifying visual and
  vision-language tracking via contrastive learning,'' \emph{arXiv preprint
  arXiv:2401.11228}, 2024.

\bibitem{ge2023beyond}
J.~Ge, X.~Chen, J.~Cao, X.~Zhu, W.~Liu, and B.~Liu, ``Beyond visual cues:
  Synchronously exploring target-centric semantics for vision-language
  tracking,'' \emph{arXiv preprint arXiv:2311.17085}, 2023.

\bibitem{clip}
A.~Radford, J.~W. Kim, C.~Hallacy, A.~Ramesh, G.~Goh, S.~Agarwal, G.~Sastry,
  A.~Askell, P.~Mishkin, J.~Clark \emph{et~al.}, ``Learning transferable visual
  models from natural language supervision,'' in \emph{International Conference
  on Machine Learning}.\hskip 1em plus 0.5em minus 0.4em\relax PMLR, 2021, pp.
  8748--8763.

\bibitem{align}
C.~Jia, Y.~Yang, Y.~Xia, Y.-T. Chen, Z.~Parekh, H.~Pham, Q.~Le, Y.-H. Sung,
  Z.~Li, and T.~Duerig, ``Scaling up visual and vision-language representation
  learning with noisy text supervision,'' in \emph{International Conference on
  Machine Learning}.\hskip 1em plus 0.5em minus 0.4em\relax PMLR, 2021, pp.
  4904--4916.

\bibitem{devlin2018bert}
J.~Devlin, M.-W. Chang, K.~Lee, and K.~Toutanova, ``Bert: Pre-training of deep
  bidirectional transformers for language understanding,'' \emph{arXiv preprint
  arXiv:1810.04805}, 2018.

\bibitem{modamixer}
M.~Guo, Z.~Zhang, L.~Jing, H.~Ling, and H.~Fan, ``Divert more attention to
  vision-language object tracking,'' \emph{arXiv preprint arXiv:2307.10046},
  2023.

\bibitem{tc128}
P.~Liang, E.~Blasch, and H.~Ling, ``Encoding color information for visual
  tracking: Algorithms and benchmark,'' \emph{IEEE transactions on image
  processing}, vol.~24, no.~12, pp. 5630--5644, 2015.

\bibitem{vot2013}
M.~Kristan, R.~Pflugfelder, A.~Leonardis, J.~Matas, F.~Porikli, L.~\v{C}ehovin
  Zajc, G.~Nebehay, G.~Fernandez, T.~Vojir, A.~Gatt, A.~Khajenezhad,
  A.~Salahledin, A.~Soltani-Farani, A.~Zarezade, A.~Petrosino, A.~Milton,
  B.~Bozorgtabar, B.~Li, C.~S. Chan, C.~Heng, D.~Ward, D.~Kearney,
  D.~Monekosso, H.~C. Karaimer, H.~R. Rabiee, J.~Zhu, J.~Gao, J.~Xiao,
  J.~Zhang, J.~Xing, K.~Huang, K.~Lebeda, L.~Cao, M.~E. Maresca, M.~K. Lim,
  M.~E. Helw, M.~Felsberg, P.~Remagnino, R.~Bowden, R.~Goecke, R.~Stolkin,
  S.~Y. Lim, S.~Maher, S.~Poullot, S.~Wong, S.~Satoh, W.~Chen, W.~Hu, X.~Zhang,
  Y.~Li, and Z.~Niu, ``The visual object tracking vot2013 challenge results,''
  Dec 2013.

\bibitem{vot2014}
S.~Hadfield, K.~Lebeda, and R.~Bowden, ``The visual object tracking vot2014
  challenge results,'' in \emph{European Conference on Computer Vision (ECCV)
  Visual Object Tracking Challenge Workshop}, 2014.

\bibitem{vot2015}
M.~Kristan, J.~Matas, A.~Leonardis, M.~Felsberg, L.~Cehovin, G.~Fernandez,
  T.~Vojir, G.~Hager, G.~Nebehay, and R.~Pflugfelder, ``The visual object
  tracking vot2015 challenge results,'' in \emph{Proceedings of the IEEE
  international conference on computer vision workshops}, 2015, pp. 1--23.

\bibitem{vot2016}
\BIBentryALTinterwordspacing
M.~Kristan, A.~Leonardis, J.~Matas, M.~Felsberg, R.~Pflugfelder, L.~\v{C}ehovin
  Zajc, T.~Vojir, G.~H\"{a}ger, A.~Luke\v{z}i\v{c}, and G.~Fernandez, ``The
  visual object tracking vot2016 challenge results,'' Springer, Oct 2016.
  [Online]. Available: \url{http://www.springer.com/gp/book/9783319488806}
\BIBentrySTDinterwordspacing

\bibitem{vot2017}
\BIBentryALTinterwordspacing
M.~Kristan, A.~Leonardis, J.~Matas, M.~Felsberg, R.~Pflugfelder, L.~\v{C}ehovin
  Zajc, T.~Vojir, G.~H\"{a}ger, A.~Luke\v{z}i\v{c}, A.~Eldesokey, and
  G.~Fernandez, ``The visual object tracking vot2017 challenge results,'' 2017.
  [Online]. Available:
  \url{http://openaccess.thecvf.com/content_ICCV_2017_workshops/papers/w28/Kristan_The_Visual_Object_ICCV_2017_paper.pdf}
\BIBentrySTDinterwordspacing

\bibitem{vot2019}
M.~Kristan, J.~Matas, A.~Leonardis, M.~Felsberg, R.~Pflugfelder, J.-K.
  Kamarainen, L.~ˇCehovin~Zajc, O.~Drbohlav, A.~Lukezic, A.~Berg
  \emph{et~al.}, ``The seventh visual object tracking vot2019 challenge
  results,'' in \emph{Proceedings of the IEEE/CVF International Conference on
  Computer Vision Workshops}, 2019, pp. 0--0.

\bibitem{vot2020}
M.~Kristan, A.~Leonardis, J.~Matas, M.~Felsberg, R.~Pflugfelder, J.-K.
  K{\"a}m{\"a}r{\"a}inen, M.~Danelljan, L.~{\v{C}}. Zajc,
  A.~Luke{\v{z}}i{\v{c}}, O.~Drbohlav \emph{et~al.}, ``The eighth visual object
  tracking vot2020 challenge results,'' in \emph{European Conference on
  Computer Vision}.\hskip 1em plus 0.5em minus 0.4em\relax Springer, 2020, pp.
  547--601.

\bibitem{vot2021}
M.~Kristan, J.~Matas, A.~Leonardis, M.~Felsberg, R.~Pflugfelder, J.-K.
  K{\"a}m{\"a}r{\"a}inen, H.~J. Chang, M.~Danelljan, L.~Cehovin,
  A.~Luke{\v{z}}i{\v{c}} \emph{et~al.}, ``The ninth visual object tracking
  vot2021 challenge results,'' in \emph{Proceedings of the IEEE/CVF
  International Conference on Computer Vision}, 2021, pp. 2711--2738.

\bibitem{vot2022}
M.~Kristan, A.~Leonardis, J.~Matas, M.~Felsberg, R.~Pflugfelder, J.-K.
  Kamarainen, H.~J. Chang, M.~Danelljan, L.~\v{C}ehovin Zajc,
  A.~Luke\v{z}i\v{c}, O.~Drbohlav, J.~Bjorklund, Y.~Zhang, Z.~Zhang, S.~Yan,
  W.~Yang, D.~Cai, C.~Mayer, and G.~Fernandez, ``The tenth visual object
  tracking vot2022 challenge results,'' 2022.

\bibitem{VOT-TIR2016}
K.~Lebeda, S.~Hadfield, R.~Bowden \emph{et~al.}, ``The thermal infrared visual
  object tracking vot-tir2016 challenge result,'' in \emph{Proceedings,
  European Conference on Computer Vision (ECCV) workshops}, 2016.

\bibitem{LSOTB-TIR}
Q.~Liu, X.~Li, Z.~He, C.~Li, J.~Li, Z.~Zhou, D.~Yuan, J.~Li, K.~Yang, N.~Fan
  \emph{et~al.}, ``Lsotb-tir: A large-scale high-diversity thermal infrared
  object tracking benchmark,'' in \emph{Proceedings of the 28th ACM
  International Conference on Multimedia}, 2020, pp. 3847--3856.

\bibitem{waymo}
P.~Sun, H.~Kretzschmar, X.~Dotiwalla, A.~Chouard, V.~Patnaik, P.~Tsui, J.~Guo,
  Y.~Zhou, Y.~Chai, B.~Caine \emph{et~al.}, ``Scalability in perception for
  autonomous driving: Waymo open dataset,'' in \emph{Proceedings of the
  IEEE/CVF conference on computer vision and pattern recognition}, 2020, pp.
  2446--2454.

\bibitem{cdtb}
A.~Lukezic, U.~Kart, J.~Kapyla, A.~Durmush, J.-K. Kamarainen, J.~Matas, and
  M.~Kristan, ``Cdtb: A color and depth visual object tracking dataset and
  benchmark,'' in \emph{Proceedings of the IEEE/CVF International Conference on
  Computer Vision}, 2019, pp. 10\,013--10\,022.

\bibitem{GTOT}
C.~Li, H.~Cheng, S.~Hu, X.~Liu, J.~Tang, and L.~Lin, ``Learning collaborative
  sparse representation for grayscale-thermal tracking,'' \emph{IEEE
  Transactions on Image Processing}, vol.~25, no.~12, pp. 5743--5756, 2016.

\bibitem{RGBT210}
C.~Li, N.~Zhao, Y.~Lu, C.~Zhu, and J.~Tang, ``Weighted sparse representation
  regularized graph learning for rgb-t object tracking,'' in \emph{Proceedings
  of the 25th ACM international conference on Multimedia}, 2017, pp.
  1856--1864.

\bibitem{RGBT234}
C.~Li, X.~Liang, Y.~Lu, N.~Zhao, and J.~Tang, ``Rgb-t object tracking:
  Benchmark and baseline,'' \emph{Pattern Recognition}, vol.~96, p. 106977,
  2019.

\bibitem{VOT-RGBT2019}
M.~Kristan, J.~Matas, A.~Leonardis, M.~Felsberg, R.~Pflugfelder, J.-K.
  Kamarainen, L.~ˇCehovin~Zajc, O.~Drbohlav, A.~Lukezic, A.~Berg
  \emph{et~al.}, ``The seventh visual object tracking vot2019 challenge
  results,'' in \emph{Proceedings of the IEEE/CVF International Conference on
  Computer Vision Workshops}, 2019, pp. 0--0.

\bibitem{VOT-RGBT2020}
M.~Kristan, A.~Leonardis, J.~Matas, M.~Felsberg, R.~Pflugfelder, J.-K.
  K{\"a}m{\"a}r{\"a}inen, M.~Danelljan, L.~{\v{C}}. Zajc,
  A.~Luke{\v{z}}i{\v{c}}, O.~Drbohlav \emph{et~al.}, ``The eighth visual object
  tracking vot2020 challenge results,'' in \emph{European Conference on
  Computer Vision}.\hskip 1em plus 0.5em minus 0.4em\relax Springer, 2020, pp.
  547--601.

\bibitem{LasHeR}
C.~Li, W.~Xue, Y.~Jia, Z.~Qu, B.~Luo, J.~Tang, and D.~Sun, ``Lasher: A
  large-scale high-diversity benchmark for rgbt tracking,'' \emph{IEEE
  Transactions on Image Processing}, vol.~31, pp. 392--404, 2021.

\bibitem{HMFT}
P.~Zhang, J.~Zhao, D.~Wang, H.~Lu, and X.~Ruan, ``Visible-thermal uav tracking:
  A large-scale benchmark and new baseline,'' in \emph{Proceedings of the
  IEEE/CVF Conference on Computer Vision and Pattern Recognition}, 2022, pp.
  8886--8895.

\bibitem{li2017tracking}
Z.~Li, R.~Tao, E.~Gavves, C.~G. Snoek, and A.~W. Smeulders, ``Tracking by
  natural language specification,'' in \emph{Proceedings of the IEEE Conference
  on Computer Vision and Pattern Recognition}, 2017, pp. 6495--6503.

\bibitem{tnl2k}
X.~Wang, X.~Shu, Z.~Zhang, B.~Jiang, Y.~Wang, Y.~Tian, and F.~Wu, ``Towards
  more flexible and accurate object tracking with natural language: Algorithms
  and benchmark,'' in \emph{Proceedings of the IEEE/CVF Conference on Computer
  Vision and Pattern Recognition}, 2021, pp. 13\,763--13\,773.

\bibitem{otb50}
Y.~Wu, J.~Lim, and M.-H. Yang, ``Online object tracking: A benchmark,'' in
  \emph{Proceedings of the IEEE conference on computer vision and pattern
  recognition}, 2013, pp. 2411--2418.

\bibitem{wordnet}
G.~A. Miller, ``Wordnet: a lexical database for english,'' \emph{Communications
  of the ACM}, vol.~38, no.~11, pp. 39--41, 1995.

\bibitem{lei2019fully}
C.~Lei and Q.~Chen, ``Fully automatic video colorization with
  self-regularization and diversity,'' in \emph{Proceedings of the IEEE/CVF
  Conference on Computer Vision and Pattern Recognition}, 2019, pp. 3753--3761.

\bibitem{houlsby2019parameter}
N.~Houlsby, A.~Giurgiu, S.~Jastrzebski, B.~Morrone, Q.~De~Laroussilhe,
  A.~Gesmundo, M.~Attariyan, and S.~Gelly, ``Parameter-efficient transfer
  learning for nlp,'' in \emph{International Conference on Machine
  Learning}.\hskip 1em plus 0.5em minus 0.4em\relax PMLR, 2019, pp. 2790--2799.

\bibitem{alexnet}
A.~Krizhevsky, I.~Sutskever, and G.~E. Hinton, ``Imagenet classification with
  deep convolutional neural networks,'' \emph{Communications of the ACM},
  vol.~60, no.~6, pp. 84--90, 2017.

\bibitem{googlenet}
C.~Szegedy, W.~Liu, Y.~Jia, P.~Sermanet, S.~Reed, D.~Anguelov, D.~Erhan,
  V.~Vanhoucke, and A.~Rabinovich, ``Going deeper with convolutions,'' in
  \emph{Proceedings of the IEEE conference on computer vision and pattern
  recognition}, 2015, pp. 1--9.

\bibitem{resnet}
K.~He, X.~Zhang, S.~Ren, and J.~Sun, ``Deep residual learning for image
  recognition,'' in \emph{Proceedings of the IEEE conference on computer vision
  and pattern recognition}, 2016, pp. 770--778.

\bibitem{vit}
A.~Dosovitskiy, L.~Beyer, A.~Kolesnikov, D.~Weissenborn, X.~Zhai,
  T.~Unterthiner, M.~Dehghani, M.~Minderer, G.~Heigold, S.~Gelly \emph{et~al.},
  ``An image is worth 16x16 words: Transformers for image recognition at
  scale,'' \emph{arXiv preprint arXiv:2010.11929}, 2020.

\bibitem{swin}
Z.~Liu, Y.~Lin, Y.~Cao, H.~Hu, Y.~Wei, Z.~Zhang, S.~Lin, and B.~Guo, ``Swin
  transformer: Hierarchical vision transformer using shifted windows,'' in
  \emph{Proceedings of the IEEE/CVF International Conference on Computer
  Vision}, 2021, pp. 10\,012--10\,022.

\bibitem{respul}
Q.~Wu, J.~Wan, and A.~B. Chan, ``Progressive unsupervised learning for visual
  object tracking,'' in \emph{Proceedings of the IEEE/CVF Conference on
  Computer Vision and Pattern Recognition}, 2021, pp. 2993--3002.

\bibitem{s2siamfc}
C.~H. Sio, Y.-J. Ma, H.-H. Shuai, J.-C. Chen, and W.-H. Cheng, ``S2siamfc:
  Self-supervised fully convolutional siamese network for visual tracking,'' in
  \emph{Proceedings of the 28th ACM International Conference on Multimedia},
  2020, pp. 1948--1957.

\bibitem{wang2021unsupervised}
N.~Wang, W.~Zhou, Y.~Song, C.~Ma, W.~Liu, and H.~Li, ``Unsupervised deep
  representation learning for real-time tracking,'' \emph{International Journal
  of Computer Vision}, vol. 129, no.~2, pp. 400--418, 2021.

\bibitem{wang2019unsupervised}
N.~Wang, Y.~Song, C.~Ma, W.~Zhou, W.~Liu, and H.~Li, ``Unsupervised deep
  tracking,'' in \emph{Proceedings of the IEEE/CVF Conference on Computer
  Vision and Pattern Recognition}, 2019, pp. 1308--1317.

\bibitem{yuan2020self}
W.~Yuan, M.~Y. Wang, and Q.~Chen, ``Self-supervised object tracking with
  cycle-consistent siamese networks,'' in \emph{2020 IEEE/RSJ International
  Conference on Intelligent Robots and Systems (IROS)}.\hskip 1em plus 0.5em
  minus 0.4em\relax IEEE, 2020, pp. 10\,351--10\,358.

\bibitem{zheng2021learning}
J.~Zheng, C.~Ma, H.~Peng, and X.~Yang, ``Learning to track objects from
  unlabeled videos,'' in \emph{Proceedings of the IEEE/CVF International
  Conference on Computer Vision}, 2021, pp. 13\,546--13\,555.

\bibitem{shen2022unsupervised}
Q.~Shen, L.~Qiao, J.~Guo, P.~Li, X.~Li, B.~Li, W.~Feng, W.~Gan, W.~Wu, and
  W.~Ouyang, ``Unsupervised learning of accurate siamese tracking,'' in
  \emph{Proceedings of the IEEE/CVF Conference on Computer Vision and Pattern
  Recognition}, 2022, pp. 8101--8110.

\bibitem{lukezivc2020performance}
A.~Luke{\'z}i{\v{c}}, L.~{\v{C}}. Zajc, T.~Voj{\'\i}{\v{r}}, J.~Matas, and
  M.~Kristan, ``Performance evaluation methodology for long-term single-object
  tracking,'' \emph{IEEE transactions on cybernetics}, vol.~51, no.~12, pp.
  6305--6318, 2020.

\bibitem{liang2022bevfusion}
T.~Liang, H.~Xie, K.~Yu, Z.~Xia, Z.~Lin, Y.~Wang, T.~Tang, B.~Wang, and
  Z.~Tang, ``Bevfusion: A simple and robust lidar-camera fusion framework,''
  \emph{arXiv preprint arXiv:2205.13790}, 2022.

\bibitem{liu2022bevfusion}
Z.~Liu, H.~Tang, A.~Amini, X.~Yang, H.~Mao, D.~Rus, and S.~Han, ``Bevfusion:
  Multi-task multi-sensor fusion with unified bird's-eye view representation,''
  \emph{arXiv preprint arXiv:2205.13542}, 2022.

\bibitem{bai2022transfusion}
X.~Bai, Z.~Hu, X.~Zhu, Q.~Huang, Y.~Chen, H.~Fu, and C.-L. Tai, ``Transfusion:
  Robust lidar-camera fusion for 3d object detection with transformers,'' in
  \emph{Proceedings of the IEEE/CVF Conference on Computer Vision and Pattern
  Recognition}, 2022, pp. 1090--1099.

\bibitem{glip}
L.~H. Li, P.~Zhang, H.~Zhang, J.~Yang, C.~Li, Y.~Zhong, L.~Wang, L.~Yuan,
  L.~Zhang, J.-N. Hwang \emph{et~al.}, ``Grounded language-image
  pre-training,'' in \emph{Proceedings of the IEEE/CVF Conference on Computer
  Vision and Pattern Recognition}, 2022, pp. 10\,965--10\,975.

\bibitem{transparent}
H.~Fan, H.~A. Miththanthaya, S.~R. Rajan, X.~Liu, Z.~Zou, Y.~Lin, H.~Ling
  \emph{et~al.}, ``Transparent object tracking benchmark,'' in
  \emph{Proceedings of the IEEE/CVF International Conference on Computer
  Vision}, 2021, pp. 10\,734--10\,743.

\bibitem{transparent2023}
K.~Garigapati, E.~Blasch, J.~Wei, and H.~Ling, ``Transparent object tracking
  with enhanced fusion module,'' in \emph{2023 IEEE/RSJ International
  Conference on Intelligent Robots and Systems (IROS)}.\hskip 1em plus 0.5em
  minus 0.4em\relax IEEE, 2023, pp. 7696--7703.

\bibitem{transparent2024new}
A.~Lukezic, Z.~Trojer, J.~Matas, and M.~Kristan, ``A new dataset and a
  distractor-aware architecture for transparent object tracking,'' \emph{arXiv
  preprint arXiv:2401.03872}, 2024.

\end{thebibliography}
